\documentclass[preprint,12pt, sort&compress]{elsarticle}
\usepackage[utf8]{inputenc}
\usepackage{xcolor}
\usepackage{algpseudocode}
\usepackage{algorithm}
\usepackage{graphicx}
\usepackage{color}
\usepackage{amsmath,amssymb,amsfonts,bm}
\usepackage{lineno}
\usepackage{caption}
\usepackage{subcaption}
\usepackage{float}
\usepackage{tabularx,booktabs,siunitx}
\usepackage{multirow}
\usepackage{comment}

\begin{document}

\title{Randomized Physics-Informed Machine Learning for Uncertainty Quantification in High-Dimensional Inverse Problems}

\author[UIUC]{Yifei Zong}
\author[PNNL]{David Barajas-Solano}
\author[UIUC,PNNL]{Alexandre M. Tartakovsky\corref{mycorrespondingauthor}}
\cortext[mycorrespondingauthor]{Corresponding author}
\ead{amt1998@illinois.edu}

\address[UIUC]{Department of Civil and Environmental Engineering, University of Illinois Urbana-Champaign, Urbana, IL 61801}
\address[PNNL]{Pacific Northwest National Laboratory, Richland, WA 99352}

\begin{abstract}

We propose a physics-informed machine learning method for uncertainty quantification in high-dimensional inverse problems. In this method, the states and parameters of partial differential equations (PDEs) are approximated with truncated conditional Karhunen-Loève expansions (CKLEs), which, by construction, match the measurements of the respective variables. The maximum a posteriori (MAP) solution of the inverse problem is formulated as a minimization problem over CKLE coefficients where the loss function is the sum of the norm of PDE residuals and the $\ell_2$ regularization term. This MAP formulation is known as the physics-informed CKLE (PICKLE) method. Uncertainty in the inverse solution is quantified in terms of the posterior distribution of CKLE coefficients, and we sample the posterior by solving a randomized PICKLE minimization problem, formulated by adding zero-mean Gaussian perturbations in the PICKLE loss function. We call the proposed approach the randomized PICKLE (rPICKLE) method. 

For linear and low-dimensional nonlinear problems (15 CKLE parameters), we show analytically and through comparison with Hamiltonian Monte Carlo (HMC) that the rPICKLE posterior converges to the true posterior given by the Bayes rule. For high-dimensional non-linear problems with 2000 CKLE parameters, we numerically demonstrate that rPICKLE posteriors are highly informative–they provide mean estimates with an accuracy comparable to the estimates given by the MAP solution and the confidence interval that mostly covers the reference solution. We are not able to obtain the HMC posterior to validate rPICKLE's convergence to the true posterior due to the HMC's prohibitive computational cost for the considered high-dimensional problems.  Our results demonstrate the advantages of rPICKLE over HMC for approximately sampling high-dimensional posterior distributions subject to physics constraints.

\end{abstract}

\maketitle

\section{Introduction}\label{sec:intro}
Inverse uncertainty quantification (IUQ) problems are ubiquitous in the modeling of natural and engineering systems governed by partial differential equations (PDEs) (e.g., subsurface flow systems, geothermal systems, CO$_2$ sequestration, climate modeling). IUQ has the same challenges as forward UQ, including the curse of dimensionality (COD), and additional challenges associated with the non-uniqueness of inverse PDE problems~\cite{zhou2014inverse,linde2017uncertainty}. 

In this work, we are interested in IUQ for the PDE model
\begin{equation}\label{eq:PDE_general}
  \mathcal{L}(u(\mathbf{x}),y(\mathbf{x}))=0, \quad \mathbf{x} \in \Omega
\end{equation}
subject to the appropriate boundary conditions, where $\mathcal{L}$ is the differential operator acting on the state variable $u(\mathbf{x})$ and the space-dependent parameter $y(\mathbf{x})$. 
Usually, the domain $\Omega$ is discretized using a grid, $\mathcal{L}$ is discretized using a numerical method of choice, and $y(\mathbf{x})$ and $u(\mathbf{x})$ are given as the corresponding sets of discrete values according to the numerical method of choice. Alternatively, deep learning models such as the physics-informed neural network (PINN) method can be employed \cite{tartakovsky2020pinn,He2021WRR,Zong2023CMAME}, where $y(\mathbf{x})$ and $u(\mathbf{x})$ are represented with deep neural networks (DNNs) and  $\mathcal{L}$ is approximated using automatic differentiation. Then, the forward solution is found by minimizing the PDE residuals over the trainable parameters of the $u$ DNN. 

In the inverse setting, given the measurements of $y$ and $u$, deterministic inverse solutions give the point estimates of $y$ and $u$. The standard method for finding the point estimate is to solve the PDE-constrained optimization problem where the objective function is the sum of the square differences between predicted and observed $y$ and $u$ values. These problems are ill-posed, and a unique solution cannot be found without adding regularity constraints.  Adding regularization, which can be interpreted as prior knowledge, leads to the maximum a posteriori  (MAP) formulation of the inverse solution. However, the PDE-constrained MAP formulations may suffer from the COD. For example, for the diffusion equation model with the unknown space-dependent diffusion coefficient considered in this work, MAP's computational cost increases as $O(n^3)$, where $n$ is the number of unknown parameters (discrete values of the diffusion coefficient) \cite{Yeung2021PICKLE}. 

Models of natural and engineering systems can have thousands of (unknown) parameters, and many dimension reduction methods have been proposed for dealing with model complexity \cite{anderson2015applied}. For unknown parameter fields, the dimension reduction methods usually exploit the spatial correlation of the parameters. For example, in the pilot point method \cite{ramarao1995pilot,doherty2010ppm}, parameters are estimated in a few preselected locations (pilot points), and the rest of the field is reconstructed using the Gaussian process regression (GPR) or ``Kriging'' method.  Another approach is to linearly project parameters onto a low-dimensional subspace using singular value decomposition (SVD) \cite{tonkin2005hybrid} and truncated Karhunen-Lo{\'e}ve expansion  (KLE) \cite{marzouk2009dimensionality} and conditional KLEs (CKLE) \cite{LI2020JCP}. In this approach, the parameter estimation problem is reduced to estimating coefficients in the low-dimensional representation using MAP or Bayesian sampling.

In \cite{tartakovsky2021PICKLE,Yeung2021PICKLE},  both the state variable $u(\mathbf{x})$ and space-dependent parameter $y(\mathbf{x})$ were projected on the CKLE space and the inverse problem was formulated as a PDE residual least-square minimization problem. This method was termed the physics-informed CKLE or PICKLE method.  In \cite{Yeung2021PICKLE}, for a problem with 1000 intrinsic dimensions (the parameter field is represented with a 1000-dimensional truncated CKLE), it was shown that the PICKLE scales as $O(n^{1.15})$ (versus the cubic scaling of MAP).
Nonetheless, (deterministic) MAP and PICKLE only provide point estimates of parameters corresponding to the mode of the posterior distribution (least-square optimum).

IUQ problems are commonly formulated in the Bayesian framework \cite{marzouk2009dimensionality,stuart2010inverse}, where the distributions of the unknown parameters are sought \cite{Langevin2011NSMC, Yoon2013nullspaceMC, LI2020JCP}. When the Bayesian framework is combined with dimension reduction methods, the distribution of parameters in the reduced space is estimated, e.g., the distribution of the coefficients in KLE expansions of parameters fields  \cite{marzouk2009dimensionality,LI2020JCP}.
In the following discussion, we will denote parameters in the reduced space by a vector $\boldsymbol\zeta$, which in the Bayesian framework is treated as a random vector.  The task of IUQ is to estimate the (posterior) distribution of $\boldsymbol\zeta$ given its prior distribution $P(\boldsymbol\zeta)$ and the set of observations  $\mathcal{D}$ of the variables parameterized by $\boldsymbol\zeta$. According to the Bayes' rule, the posterior distribution $P(\boldsymbol\zeta | \mathcal{D})$ is given as 
\begin{eqnarray}\label{eq:Bayes_theorem}
P(\boldsymbol\zeta | \mathcal{D}) = \frac{P(\mathcal{D}| \boldsymbol\zeta )P(\boldsymbol\zeta )}
{P(\mathcal{D})  },
\end{eqnarray}
where $P(\mathcal{D}| \boldsymbol\zeta )$ is the likelihood density and  $P(\mathcal{D}) = \int P(\mathcal{D}|\boldsymbol\zeta )P(\boldsymbol\zeta ) d\boldsymbol\zeta $ is the evidence. 
The prior distribution encompasses prior beliefs or knowledge about the unknowns, e.g., the distribution of parameters and states that can be learned from data only. The likelihood is a probabilistic model that quantifies the distance between observations and predictions from the forward model (i.e., an error model). For a linear forward model with Gaussian priors and a likelihood model, the calculation of $P(\mathcal{D})$ can be analytically performed. In a more general scenario, the computation of $P(\mathcal{D})$ is intractable \cite{brooks1998markov}.

Approximate inference methods avoid directly computing $P(\mathcal{D})$ by either approximating $P(\boldsymbol\zeta | \mathcal{D})$ with a parameterized distribution (variational inference (VI) methods) or generating a sequence of samples from the posterior (Markov chain Monte Carlo (MCMC) methods) \cite{abdar2021UQDLreview}. VI methods optimize parameters in the assumed distribution to minimize the discrepancy between the parameterized and true posterior distributions, based on a specific probabilistic distance metric \cite{sun2020physics, gou2022bayesian}. For example, mean-field VI assumes a fully factorized structure of the posterior distribution \cite{blei2017vi}. However, such specific structures can yield biased approximations, particularly for the posterior exhibiting strongly correlated structure \cite{yao2019quality}, which is common in PDE-constrained inversions. Furthermore, mean-field VI commonly underestimates the posterior variance, which is undesirable for reliability analysis and risk assessment. 

MCMC methods construct multiple chains of $\boldsymbol\zeta$ samples with stationary distributions equal to the target posterior distribution $P(\boldsymbol\zeta | \mathcal{D})$. However, MCMC suffers from the COD, especially the random-walk MCMC methods. One way to accelerate MCMC is to reduce autocorrelation between consecutive samples in the Markov chains. HMC reduces the autocorrelation by making larger jumps in the parameter space using properties of the simulated Hamiltonian dynamics and is a common choice for Bayesian training of statistical models \cite{neal2011mcmc,fichtner2019hmctopo}.  However, there are challenges in applying HMC for high-dimensional IUQ problems. HMC is also not immune to COD because the increasing number of unknown parameters generally reduces the sampling efficiency of HMC and increases the autocorrelations of the consecutive samples. Another significant challenge with using HMC for IUQ problems is that the posterior covariance structure could be poorly conditioned \cite{langmore2023hmcillcond} because the posterior distribution has vastly different correlations between different pairs of parameters.  This can lead to biased HMC estimates \cite{neal2011mcmc, betancourt2017conceptual}. Nonlinearity in the model generally exacerbates the complexity of the posterior distribution and increases the HMC bias.

Randomized MAP \cite{wang2018rmap} and similar randomization methods such as randomized maximum likelihood \cite{chen2012ensemble} and randomize-then-optimize \cite{bardsley2014randomize} were proposed as alternatives to VI and MCMC methods for posterior sampling. For example, in randomized MAP, a (deterministic) PDE-constrained optimization problem is replaced with a stochastic problem where a stochastic objective function is minimized subject to the same deterministic PDE constraint as in the standard MAP. Then, the samples of the posterior distribution are obtained by solving the minimization problem for different realizations of the random terms in the objective function. While avoiding the MCMC challenges of sampling a complex posterior, the randomized MAP has the same cubic dependence on the number of unknown parameters as the standard MAP and is therefore limited to relatively low-dimensional problems.    

Finally, we should mention the iterative ensemble smoother (IES) and ensemble Kalman filter methods \cite{white_ies_2018}, which can be used to approximate the posterior distribution of parameters. 
Despite their reliance on linear regression models to approximate maps from input to output variables, these methods have demonstrated robustness even for nonlinear PDE problems.

In this work, we propose an efficient sampling method for high-dimensional IUQ based on the residual least-square formulation of the inverse problem found, among others, in the PICKLE and PINN methods.  The main ideas of this approach include adding independent Gaussian noise to each term of the objective function in a residual least-square method and solving the resulting problem for different realizations of the noise terms. The mean and variance of the noise distributions are chosen to make the ensemble of optimization problem solutions converge to the target posterior distribution. We test this approach for the PICKLE residual least-square formulation.
We apply the ``randomized PICKLE'' (rPICKLE) method to a 2000-dimensional IUQ problem and demonstrate that it avoids the challenges of sampling complex posterior distributions  (i.e., distributions with ill-conditioned covariance matrices).  Our analysis shows that the distribution of the rPICKLE samples converges to the exact posterior in the linear case with the Gaussian priors. For nonlinear problems, the possible deviations from the exact posterior can be adjusted by the Metropolis algorithm. In numerical experiments, we find the deviations of the sampled distribution from the posterior distribution are minor and can be disregarded, which eliminates the need for Metropolization.

This paper is organized as follows. In Section \ref{sec:PDE_general_rpickle}, we review the PICKLE formulation and present the randomized PICKLE method for approximate Bayesian parameter estimation in the  PDE~\eqref{eq:PDE_general}.  In Section \ref{sec:Hanford_problem_formulation}, we formulate rPICKLE for the IUQ problem arising in groundwater flow modeling. In Section \ref{sec:numerical_results}, we test rPICKLE for low- (15) and high-dimensional (2000) parameter estimation problems. A comparison with HMC is provided for the low-dimensional case.  Discussion and conclusions are provided in Section \ref{discussion}.

\section{Randomized PICKLE Formulation}\label{sec:PDE_general_rpickle}

\subsection{Conditional KLE}\label{sec:CKLE}
Here, we formulate rPICKLE for approximate Bayesian parameter estimation in the PDE model \eqref{eq:PDE_general}. The starting point of rPICKLE is to model the PDE parameter $y(\mathbf{x})$ and state $u(\mathbf{x})$ in Eq.~\eqref{eq:PDE_general} as random processes
 $y(\mathbf{x}, \tilde{\omega})$ and $u(\mathbf{x}, \tilde{\omega})$, respectively, where $\tilde{\omega}$ is a coordinate in the outcome space. Given observations $\mathbf{u}^{\text{obs}} = \{u^{\text{obs}}_{i}\}_{i = 1}^{N_u^{\text{obs}}}$ of $u$ and $\mathbf{y}^{\text{obs}} = \{y_i^{\text{obs}}\}_{i = 1}^{N_y^{\text{obs}}}$ of  $y$, $y(\mathbf{x}, \tilde{\omega})$ and $u(\mathbf{x}, \tilde{\omega})$ can be approximated with truncated CKLEs. For $y(\mathbf{x}, \tilde{\omega})$, the CKLE takes the form
\begin{eqnarray}\label{eq:CKLE_k}
       y(\mathbf{x},\tilde{\omega}) \approx \hat{y}(\mathbf{x}, \boldsymbol\xi) &=& \overline{y^c}(\mathbf{x}) + \sum_{i = 1}^{N_\xi} \sqrt{\lambda^y_i} \psi_i^y(\mathbf{x}){\xi}_i,
\end{eqnarray}
where $\overline{y^c}$ is the prior mean of $y$ conditioned on $y$ observations and $\{ \lambda^y_i \}_{i = 1}^{N_\xi}$ and $\{ \psi_i^y(\mathbf{x}) \}_{i = 1}^{N_\xi}$ are the leading $N_\xi$ eigenvalues and eigenfunctions of the prior covariance of $y$ conditioned on $y$ observations, $C_y^c(\mathbf{x}, \mathbf{x^\prime})$. Both  $\overline{y}^c$ and $C_y^c(\mathbf{x}, \mathbf{x^\prime})$ reflect the prior knowledge about $y(x)$ and are obtained via GPR equations, defined in \ref{sec:gpr}. In Eq.~\eqref{eq:CKLE_k}, $\boldsymbol{\xi} =[ \xi_1,..., \xi_{N_{\xi}}]^T$ is the vector of random variables with the prior independent standard normal distribution. For this prior of $\boldsymbol{\xi}$, the (prior) mean and covariance of $\hat{y}(\mathbf{x}, \boldsymbol\xi)$ are equal, up to the truncation error, to those of $y(\mathbf{x},\tilde{\omega})$.
Estimating the posterior distribution of $\boldsymbol\xi$, i.e., the distribution constrained by the governing PDE, is the goal of rPICKLE. 

Similarly, the CKLE of $u(\mathbf{x}, \tilde{\omega})$ has the form
\begin{eqnarray}\label{eq:CKLE_u}
    u(\mathbf{x},\tilde{\omega})\approx \hat{u}(\mathbf{x}, \boldsymbol{\eta}) &=& \overline{u}^c(\mathbf{x}) + \sum_{i = 1}^{N_{\eta}}\sqrt{\lambda^u_i}\psi_i^u(\mathbf{x})\eta_i, 
\end{eqnarray} 
where $\overline{u}^c$ is the conditional mean of $u$ and  $\{ \lambda_i^u \}_{i = 1}^{N_\eta}$ and $\{ \psi_i^u(\mathbf{x}) \}_{i = 1}^{N_\eta}$ are the leading eigenvalues and eigenfunctions of $C_u^c(\mathbf{x}, \mathbf{x^\prime})$, which is the prior covariance of $u$ conditioned on $u$ observations. 
The prior covariance of $u$ is obtained by Monte Carlo sampling of the solution of Eq.~\eqref{eq:PDE_general}  as described in \ref{sec:gpr}.
 As in the CKLE of $y$, $\{ \eta_i \}_{i = 1}^{N_{\eta}}$ are random variables with the independent standard normal prior distribution. The posterior distribution of $\boldsymbol{\eta}$ is estimated as part of the rPICKLE inversion. 

\subsection{Revisiting PICKLE}\label{sec:pickle}
We formulate rPICKLE  by randomizing a loss function in the PICKLE method, which was presented in \cite{tartakovsky2021PICKLE,Yeung2021PICKLE} for solving high-dimensional inverse PDE problems with unknown space-dependent parameters. In PICKLE, the unknown $\hat{y}(\mathbf{x},\tilde{\boldsymbol\xi})$ and $\hat{u}(\mathbf{x},\tilde{\boldsymbol\eta})$ fields are treated as  one realization of the random $y$ and $u$ fields, respectively, and $\tilde{\boldsymbol\xi}$ and $\tilde{\boldsymbol\eta}$ are (deterministic) parameters that are found as the solution of the minimization problem:

\begin{eqnarray}\label{eq:PICKLE_loss_w}
  (\tilde{\boldsymbol\xi}^*, \tilde{\boldsymbol\eta}^*) &=& \arg\min_{\tilde{\boldsymbol\xi}, \tilde{\boldsymbol\eta}} L(\tilde{\boldsymbol\xi}, \tilde{\boldsymbol\eta})  \\
 &=&  \arg\min_{\tilde{\boldsymbol\xi}, \tilde{\boldsymbol\eta}} \frac{\omega_r}{2}\|\mathcal{R}(\tilde{\boldsymbol\xi}, \tilde{\boldsymbol\eta})\|^2_2 +  \frac{\omega_\xi}{2}\| \tilde{\boldsymbol\xi} \|_2^2 + \frac{\omega_\eta}{2}\| \tilde{\boldsymbol\eta} \|_2^2  \nonumber
\end{eqnarray} 
where $\mathcal{R}$ is the vector of PDE residuals computed on a discretization mesh with a numerical method of choice  (finite volume discretization was used in \cite{tartakovsky2021PICKLE}) and the last two terms are $\ell_2$ regularization terms with respect to $\tilde{\boldsymbol\xi}$ and $\tilde{\boldsymbol\eta}$.  
In Section  \ref{sec:Bayesian_pickle}, we show that this PICKLE formulation provides the mode of the joint posterior distribution of ${\boldsymbol\xi}$ and ${\boldsymbol\eta}$ given that the likelihood and prior distributions of ${\boldsymbol\xi}$ and ${\boldsymbol\eta}$ are Gaussian. 

Weights $\omega_r$, $\omega_\xi$, and $\omega_\eta$ control the relative importance of each term in the loss function. Following \cite{tartakovsky2021PICKLE}, we set $\omega_\xi = \omega_\eta$.  This choice is justified in the Bayesian context because $\omega_\xi$ and $\omega_\eta$ are related to the variances of the prior distributions of $\boldsymbol\xi$ and $\boldsymbol\eta$, which are the same and equal to one as stated in Section \ref{sec:CKLE}. Then, the minimization problem \eqref{eq:PICKLE_loss_w} can be re-written as 
\begin{eqnarray}\label{eq:PICKLE_loss}
(\tilde{\boldsymbol\xi}^*, \tilde{\boldsymbol\eta}^*) 
 =  \arg\min_{\tilde{\boldsymbol\xi}, \tilde{\boldsymbol\eta}} \frac{1}{2}\|\mathcal{R}(\mathbf{x}; \tilde{\boldsymbol\xi}, \tilde{\boldsymbol\eta})\|^2_2 +  \frac{\gamma}{2}\|\tilde{\boldsymbol\xi} \|_2^2 + \frac{\gamma}{2}\|\tilde{\boldsymbol\eta} \|_2^2,
\end{eqnarray} 
where $\gamma = \omega_\xi/\omega_r=\omega_\eta/\omega_r$ is the regularization parameter controlling the relative magnitude of the regularization.  The value of $\gamma$ is selected to minimize the error in the PICKLE solution with respect to the reference field. If the reference field is unknown, $\gamma$ can be selected via cross-validation.

\subsection{Bayesian PICKLE}\label{sec:Bayesian_pickle}

The Bayesian estimate of the posterior distribution of $\boldsymbol{\xi}$ and $\boldsymbol{\eta}$ in the PICKLE model can be found from the Bayes rule: 
\begin{eqnarray}\label{eq:Bayesian_posterior}
P(\boldsymbol\xi, \boldsymbol\eta | \mathcal{D}_{res}) = \frac{P(\mathcal{D}_{res} | \boldsymbol\xi, \boldsymbol\eta)P(\boldsymbol\xi, \boldsymbol\eta)}
{\int\int P(\mathcal{D}_{res} | \boldsymbol\xi, \boldsymbol\eta )P(\boldsymbol\xi, \boldsymbol\eta ) d\boldsymbol\xi d\boldsymbol\eta},
\end{eqnarray}
where $P(\boldsymbol\xi, \boldsymbol\eta)$ is the joint prior distribution of $\boldsymbol{\xi}$ and $\boldsymbol{\eta}$. We assume that the prior distributions of $\boldsymbol{\xi}$ and $\boldsymbol{\eta}$ are independent, i.e., $P(\boldsymbol\xi, \boldsymbol\eta) = P(\boldsymbol\xi)P(\boldsymbol\eta)$.  $P(\mathcal{D}_{res} | \boldsymbol\xi, \boldsymbol\eta)$  is the likelihood function, where  $\mathcal{D}_{res}$ is the collection of PDE residuals evaluated on the numerical grid. The double integral in the denominator of Eq.~\eqref{eq:Bayesian_posterior} is the normalization coefficient for the posterior distribution to integrate to one.

In PICKLE, the CKLEs are conditioned on $y$ and $u$ observations. Therefore, the likelihood function only specifies the joint distribution of PDE residuals. The form of $P(\mathcal{D}_{res} | \boldsymbol\xi, \boldsymbol\eta)$ and $P(\boldsymbol\xi, \boldsymbol\eta)$ can be found from the PICKLE loss function  $L(\tilde{\boldsymbol\xi}, \tilde{\boldsymbol\eta})$ by requiring the PICKLE solution $\tilde{\boldsymbol\xi}^*$ and $\tilde{\boldsymbol\eta}^*$ to also maximize the posterior probability density $P(\boldsymbol\xi, \boldsymbol\eta | \mathcal{D}_{res})$, i.e., by requiring the PICKLE solution to be the mode of the posterior distribution, which is also known as the MAP (maximum a posteriori). 

This can be achieved by taking the negative logarithm of both sides of Eq.~\eqref{eq:Bayesian_posterior}, yielding
\begin{equation}\label{eq:log_posterior}
-\ln[P(\boldsymbol\xi, \boldsymbol\eta | \mathcal{D}_{res})] = 
\left [ \frac{1}{\gamma} L(\boldsymbol\xi, \boldsymbol\eta) + C \right ] + \ln \int\int P(\mathcal{D}_{res} | \boldsymbol\xi, \boldsymbol\eta )P(\boldsymbol\xi, \boldsymbol\eta ) d\boldsymbol\xi d\boldsymbol\eta],    
\end{equation}
where $C$ is a constant and $L(\boldsymbol\xi, \boldsymbol\eta)$ is the PICKLE loss defined in Eq.~\eqref{eq:PICKLE_loss}. The left-hand side of Eq.~\eqref{eq:log_posterior} is the negative logarithm of the posterior, and we postulated that 
\begin{equation}\label{eq:prior}
-\ln \left [ P(\mathcal{D}_{res} | \boldsymbol\xi, \boldsymbol\eta)P(\boldsymbol\xi, \boldsymbol\eta) \right ]  = 
\frac{1}{\gamma} L(\boldsymbol\xi, \boldsymbol\eta) + C. 
\end{equation}
Since the last term in Eq.~\eqref{eq:log_posterior} is independent of $\boldsymbol\xi$ and $\boldsymbol\eta$,  PICKLE solutions $\tilde{\boldsymbol\xi}^*$ and $\tilde{\boldsymbol\eta}^*$ that minimize $L(\tilde{\boldsymbol\xi}, \tilde{\boldsymbol\eta})$ also maximize $P(\boldsymbol\xi, \boldsymbol\eta | \mathcal{D}_{res})$. We can break down Eq.~\eqref{eq:prior} as
\begin{equation}\label{eq:log_likelihood}
-\ln \left [ P(\mathcal{D}_{res} | \boldsymbol\xi, \boldsymbol\eta)  \right ]  = 
  \frac{1}{2\gamma}\|\mathcal{R}(\mathbf{x}; \boldsymbol\xi, \boldsymbol\eta)\|^2_2 +  
  C_1. 
\end{equation}
and 
\begin{equation}\label{eq:log_prior}
-\ln \left [ P(\boldsymbol\xi, \boldsymbol\eta) \right ]  = 
   \frac{1}{2}\| \boldsymbol\xi\|_2^2 + \frac{1}{2}\| \boldsymbol\eta\|_2^2 
 + C_2, 
\end{equation}
where $C_1 + C_2 = C$.

In Eq.~\eqref{eq:log_likelihood}, we can choose $C_1$ such that the likelihood is
\begin{eqnarray}\label{eq:likelihood}
        P(\mathcal{D}_{res} | \boldsymbol\xi, \boldsymbol\eta) = \left ( \frac{1}{\sqrt{2\pi}\sigma_{r}} \right )^N \exp \left (-\frac{1}{2} \mathcal{R}(\mathbf{x}; \boldsymbol\xi, \boldsymbol\eta)^T\Sigma_r^{-1} \mathcal{R}(\mathbf{x}; \boldsymbol\xi, \boldsymbol\eta) \right ),
\end{eqnarray} 
where $\Sigma_r = \sigma_r^2\mathbf{I}$ and $\sigma_r^2 = \gamma$. As desired, this likelihood states that the PDE residuals have zero mean.  

Similarly, in  Eq.~\eqref{eq:log_prior}, we can choose $C_2$ such that the prior is 
\begin{eqnarray}\label{eq:l2_prior}
        P(\boldsymbol\xi, \boldsymbol\eta) = \left ( \frac{1}{\sqrt{2\pi}\sigma_\xi} \right )^{N_\xi} \left ( \frac{1}{\sqrt{2\pi}\sigma_\eta} \right )^{N_\eta} \exp \left ( -\frac{1}{2} \boldsymbol{\xi}^T\Sigma^{-1}_\xi\boldsymbol{\xi} - \frac{1}{2} \boldsymbol{\eta}^T\Sigma^{-1}_\eta\boldsymbol{\eta} \right ), 
\end{eqnarray}
where $\Sigma_\xi = \sigma_\xi^2 \mathbf{I}$ ($\sigma_\xi^2 = 1$) and $\Sigma_\eta = \sigma_\eta^2\mathbf{I}$ ($\sigma_\eta^2 = 1$).  
In other words, the prior distribution of $\boldsymbol{\xi}$ and $\boldsymbol{\eta}$ is independent and standard normal. Recall that this is consistent with the definition of $\boldsymbol{\xi}$ and $\boldsymbol{\eta}$ in Section \ref{sec:CKLE}.

As mentioned earlier, computing the double integral in Eq.~\eqref{eq:Bayesian_posterior} is computationally intractable for high-dimensional $\boldsymbol{\xi}$ and $\boldsymbol{\eta}$. At this point, we just note that the proposed rPICKLE method approximately samples the posterior distribution without computing this integral. Later, we compare rPICKLE with HMC, a common approach for sampling posterior distributions from the Bayes formula. Once obtained, the set of posterior samples $\{ (\boldsymbol\xi_i, \boldsymbol\eta_i)\}_{i=1}^{N_{ens}}$ ($N_{ens}$ is the size of the ensemble) can be used to compute the distributions of $u$ and $y$ or the leading moments of these distributions. For example, the first and second moments of $y$ can be estimated as   
\begin{align}
\mu_{\hat{y}} (\mathbf{x}| \mathcal{D}_{res}) & \approx \frac{1}{N_{ens}} \sum_{i=1}^{N_{ens}} \hat{y}(\mathbf{x}; \boldsymbol\xi_i) , \label{eq:BMA_ymean} \\ 
\sigma^2_{\hat{y} }( \mathbf{x} | \mathcal{D}_{res}) &\approx \frac{1}{N_{ens}-1} \sum_{i=1}^{N_{ens}} [\hat{y}(\mathbf{x}; \boldsymbol\xi_i) - \mu_{\hat{y}} (\mathbf{x}| \mathcal{D}_{res}) ]^2 . \label{eq:BMA_ystd}
\end{align}
An important question in rPICKLE and Bayesian PICKLE is the selection of $\sigma^2_r$. One criterion for selecting $\sigma^2_r$ is to minimize the distance between the MAP estimate $\hat{y}_{\mathrm{MAP}}$ or $\mu_{\hat{y}} (\mathbf{x}| \mathcal{D}_{res})$ and the reference $y$. If the reference field is not known, then the $y$ measurements must be divided into the training, validation, and testing data, and the distance is computed with respect to the validation dataset.  

 Another possible criterion is to select $\sigma^2_r$  that maximizes the log predictive probability (LPP), which is defined as the sum of the pointwise log probabilities of the reference being observed given the statistical forecast \cite{rasmussen2006gaussian}: 
\begin{eqnarray}\label{eq:lpp}
    \mathrm{LPP} = -\sum_{i = 1}^{N} \left\{ \frac{[ \mu_{\hat{y}} (\mathbf{x}_i| \mathcal{D}_{res}) - y(\mathbf{x}_i)]^2}{2\sigma^2_{\hat{y}}(\mathbf{x}_i)} +  \frac{1}{2}\log [2\pi \sigma^2_{\hat{y}}(\mathbf{x}_i)] \right\}.
\end{eqnarray} 
Here, the summation is over all points (elements) where the reference solution or the validation data are available.

\subsection{rPICKLE: Randomization of PICKLE Loss Function}

 In rPICKLE, we randomize the PICKLE loss function as
 \begin{eqnarray}\label{eq:rPICKLE_loss}
 L^r(\boldsymbol\xi, \boldsymbol\eta; \boldsymbol\omega, \boldsymbol\alpha, \boldsymbol\beta)
& = &
 \frac{1}{2}\|\mathcal{R}(\mathbf{x}; \boldsymbol\xi, \boldsymbol\eta) - \boldsymbol\omega\|_{\Sigma_r}^2 +  \frac{1}{2}\| \boldsymbol\xi -\boldsymbol\alpha\|_{\Sigma_\xi}^2 \\ \nonumber
& + &\frac{1}{2}\| \boldsymbol\eta  - \boldsymbol\beta\|_{\Sigma_\eta}^2,
\end{eqnarray}
where $\boldsymbol\omega$, $\boldsymbol\alpha$, and $\boldsymbol\beta$ are independent random noise vectors that have the same distributions as those of $\mathcal{R}$,  $\boldsymbol\xi$, and $\boldsymbol\eta$ in the Bayesian PICKLE, i.e., they have zero mean and the covariance functions $\Sigma_w = \Sigma_r = {\sigma_r^2}\mathbf{I}$, $\Sigma_\alpha = \Sigma_\xi = \mathbf{I}$, and $\Sigma_\beta = \Sigma_\eta = \mathbf{I}$, respectively. 
Then, the samples of the posterior distribution $P(\boldsymbol\xi, \boldsymbol\eta | \mathcal{D}_{res})$ are generated by solving the 
the minimization problem:
\begin{eqnarray}\label{eq:randomized_PICKLE_loss}
 (\boldsymbol\xi^*, \boldsymbol\eta^*) = \arg\min_{\boldsymbol\xi, \boldsymbol\eta} L^r(\boldsymbol\xi, \boldsymbol\eta; \boldsymbol\omega, \boldsymbol\alpha, \boldsymbol\beta). 
\end{eqnarray}

The rPICKLE method proceeds as follows. We draw $N_{ens}$ i.i.d. samples of $\boldsymbol\omega$, $\boldsymbol\alpha$, and $\boldsymbol\beta$ and, for each sample, minimize the loss \eqref{eq:randomized_PICKLE_loss} to obtain samples $(\boldsymbol{\xi}^*_i, \boldsymbol{\eta}^*_i)$ ($i=1,...,N_{ens}$), which approximate the posterior distribution of $\boldsymbol{\xi}$ and $\boldsymbol{\eta}$. Then, we use the CKLEs to obtain the samples $\hat{y}(\mathbf{x};\boldsymbol{\xi}^*_i)$ and $\hat{u}(\mathbf{x};\boldsymbol{\eta}_i^*)$ of the posterior distribution of $y$ and $u$, respectively. In Sections \ref{sec:rpickle_linear} and \ref{sec:low-dimensional},  for linear or low-dimensional problems, we analytically and via a comparison with HMC show that distributions approximated with the samples $  \{   (\boldsymbol\xi^*_i, \boldsymbol\eta^*_i) \}_{i = 1}^{N_{ens}} $ approach the \emph{true} posteriors of $(\boldsymbol\xi, \boldsymbol\eta)$ with increasing $N_{ens}$ regardless of the choice of $\sigma^2_r$. In Section \ref{sec:high-dimensional}, we numerically demonstrate that for high-dimensional problems, rPICKLE posteriors are highly informative (we cannot obtain the HMC posterior to validate rPICKLE's convergence to the true posterior due to the HMC's prohibitive computational cost for the considered high-dimensional problems).  We note that posterior distributions depend on the choice of $\sigma^2_r$. The value of $\sigma^2_r$ for obtaining the most informative posterior is chosen as described in Section \ref{sec:Bayesian_pickle}.

\subsubsection{rPICKLE for a linear model}\label{sec:rpickle_linear}

In this section, we prove that rPICKLE samples converge to the exact posterior as $N_{ens}\to \infty$ for the PDE residual of the linear form $\mathcal{R}(\mathbf{x}; \boldsymbol\xi, \boldsymbol\eta) = \mathbf{A}\boldsymbol\xi+ \mathbf{B}\boldsymbol\eta - \mathbf{c} \in \mathbb{R}^{N}$, where $\mathbf{A} \in \mathbb{R}^{N \times N_{\xi}}, \mathbf{B} \in \mathbb{R}^{N \times N_{\eta}}$, and $\mathbf{c} \in \mathbb{R}^N$. This proof is possible because in the case of a linear $\mathcal{R}$ with the Gaussian likelihood and prior, the Bayesian posterior \eqref{eq:Bayesian_posterior} is also Gaussian \cite{rasmussen2006gaussian}. 

First, we find the mean and covariance of the posterior given by the Bayes rule. 
Taking the first and second derivatives of both sides of Eq.~\eqref{eq:log_posterior} with respect to $[\boldsymbol\xi, \boldsymbol\eta]^T$ yields the relationships between the mean  $\boldsymbol\mu_{post}$ and covariance $\Sigma_{post}$ of the posterior distribution: 
\begin{eqnarray}\label{eq:linear_posterior_mode}
    \boldsymbol\mu_{post} = 
    \Sigma_{post}
    \begin{bmatrix} 
     \mathbf{A}^T\Sigma_r^{-1}\mathbf{c}\\
     \mathbf{B}^T\Sigma_r^{-1}\mathbf{c}
    \end{bmatrix}.
\end{eqnarray}
Then, Eq.~\eqref{eq:log_posterior} can be reformulated as:
\begin{align}\label{eq:lin_post}
 & \left ( \begin{bmatrix} 
    \boldsymbol\xi \\
    \boldsymbol\eta
    \end{bmatrix} - \boldsymbol\mu_{post} \right ) ^T \Sigma^{-1}_{post} \left ( \begin{bmatrix} 
    \boldsymbol\xi  \\
    \boldsymbol\eta 
    \end{bmatrix}  - \boldsymbol\mu_{post} \right ) \nonumber  \\
    &=  \left ( \begin{bmatrix} 
    \mathbf{A}\boldsymbol\xi \\
    \mathbf{B}\boldsymbol\eta 
    \end{bmatrix} - \mathbf{c} \right )^T \Sigma_r^{-1} \left ( \begin{bmatrix} 
    \mathbf{A}\boldsymbol\xi \\
    \mathbf{B}\boldsymbol\eta
    \end{bmatrix} - \mathbf{c} \right ) + \begin{bmatrix} 
    \boldsymbol\xi \\
    \boldsymbol\eta
    \end{bmatrix}^T\begin{bmatrix} 
    \Sigma_\xi^{-1} & \mathbf{0} \\
    \mathbf{0} & \Sigma_\eta^{-1}
    \end{bmatrix}\begin{bmatrix} 
    \boldsymbol\xi \\
    \boldsymbol\eta 
    \end{bmatrix}.
\end{align}
Differentiating  Eq.~\eqref{eq:lin_post} twice with respect to $[\boldsymbol\xi, \boldsymbol\eta]^T$ yields the expression for $\Sigma_{post}$: 
\begin{eqnarray}\label{eq:gaussian_post}
    \Sigma_{post}  =  \begin{bmatrix} 
    \mathbf{A}^T\Sigma_r^{-1}\mathbf{A} +  \Sigma_\xi^{-1} & \mathbf{B}^T\Sigma_r^{-1}\mathbf{A} \\
    \mathbf{A}^T\Sigma_r^{-1}\mathbf{B} & \mathbf{B}^T\Sigma_r^{-1}\mathbf{B} +  \Sigma_\eta^{-1} 
    \end{bmatrix}^{-1}. 
\end{eqnarray}

Next, we derive the mean and covariance of $\boldsymbol\xi^*$ and $\boldsymbol\eta^*$ given by the rPICKLE Eq.~\eqref{eq:rPICKLE_loss}. For a linear $\mathcal{R}$, the randomized minimization problem \eqref{eq:randomized_PICKLE_loss} is reduced to a linear least-square problem, and its solution is given by the system of linear equations: 
\begin{eqnarray}\label{eq:randomized_loss_grad}
\frac{\partial L^r}{\partial \boldsymbol\xi}
=  \left[ (\mathbf{A}\boldsymbol\xi + \mathbf{B}\boldsymbol\eta - \mathbf{c})^T\Sigma_r^{-1}\mathbf{A} - \boldsymbol\omega^T\Sigma_r^{-1}\mathbf{A} \right] +  \left[ (\boldsymbol\xi- \boldsymbol\alpha)^T\Sigma_\xi^{-1}  \right]=0, \\
\frac{\partial L^r}{\partial \boldsymbol\eta}
=  \left[ (\mathbf{A}\boldsymbol\xi + \mathbf{B}\boldsymbol\eta - \mathbf{c})^T\Sigma_r^{-1}\mathbf{B} - \boldsymbol\omega^T\Sigma_r^{-1}\mathbf{B} \right] +  \left[ (\boldsymbol\eta - \boldsymbol\beta)^T\Sigma_\eta^{-1}  \right]=0,
\end{eqnarray}
or
\begin{eqnarray}
    \begin{bmatrix} 
    \mathbf{A}^T\Sigma_r^{-1}\mathbf{A} +  \Sigma_\xi^{-1} & \mathbf{B}^T\Sigma_r^{-1}\mathbf{A} \\
    \mathbf{A}^T\Sigma_r^{-1}\mathbf{B} & \mathbf{B}^T\Sigma_r^{-1}\mathbf{B} + \Sigma_\eta^{-1}
    \end{bmatrix}
    \begin{bmatrix} 
    \boldsymbol\xi^* \\
    \boldsymbol\eta^*
    \end{bmatrix}
    =
    \begin{bmatrix} 
     \mathbf{A}^T\Sigma_r^{-1}(\mathbf{c} + \boldsymbol\omega)+  \Sigma_\xi^{-1}\boldsymbol\alpha  \\
     \mathbf{B}^T\Sigma_r^{-1}(\mathbf{c} + \boldsymbol\omega) +  \Sigma_\eta^{-1}\boldsymbol\beta
    \end{bmatrix}.
\end{eqnarray}

The solution of this equation is:
\begin{eqnarray}\label{eq:rPICKLE_solution}
    \begin{bmatrix} 
    \boldsymbol\xi^* \\
    \boldsymbol\eta^*
    \end{bmatrix}
    =\Sigma
    \begin{bmatrix} 
     \mathbf{A}^T\Sigma_r^{-1}(\mathbf{c} + \boldsymbol\omega)+  \Sigma_\xi^{-1}\boldsymbol\alpha  \\
     \mathbf{B}^T\Sigma_r^{-1}(\mathbf{c} + \boldsymbol\omega) +  \Sigma_\eta^{-1}\boldsymbol\beta
    \end{bmatrix},
\end{eqnarray}
where
\begin{eqnarray}\label{eq:rPICKLE_cov}
   \Sigma =  \begin{bmatrix} 
    \mathbf{A}^T\Sigma_r^{-1}\mathbf{A} +  \Sigma_\xi^{-1}& \mathbf{B}^T\Sigma_r^{-1}\mathbf{A} \\
    \mathbf{A}^T\Sigma_r^{-1}\mathbf{B} & \mathbf{B}^T\Sigma_r^{-1}\mathbf{B} +  \Sigma_\eta^{-1}
    \end{bmatrix}^{-1}.
\end{eqnarray}
Comparing Eqs.~\eqref{eq:rPICKLE_cov} and~\eqref{eq:gaussian_post} yields $\Sigma = \Sigma_{post}$.
Recall that $[\boldsymbol\xi^*,\boldsymbol\eta^*]^T$ is a function of the random noises $(\boldsymbol\alpha, \boldsymbol\beta, \boldsymbol\omega)$. Taking the expectation of  $[\boldsymbol\xi^*,\boldsymbol\eta^*]^T$, we get
\begin{eqnarray}\label{eq:rPICKLE_mean}
\mathbb{E} \begin{bmatrix} 
    \boldsymbol\xi^* \\
    \boldsymbol\eta^*
    \end{bmatrix} &=& \mathbb{E} \left[ \Sigma
    \begin{bmatrix} 
   \mathbf{A}^T\Sigma_r^{-1}(\mathbf{c} + \boldsymbol\omega)+  \Sigma_\xi^{-1}\boldsymbol\alpha \\
    \mathbf{B}^T\Sigma_r^{-1}(\mathbf{c} + \boldsymbol\omega) +  \Sigma_\eta^{-1}\boldsymbol\beta
    \end{bmatrix} \right] \nonumber \\
        &=& \Sigma \begin{bmatrix} 
    \mathbf{A}^T\Sigma_r^{-1}(\mathbf{c} + \mathbb{E}[\boldsymbol\omega]) + \Sigma_\xi^{-1}\mathbb{E}[\boldsymbol\alpha] \\
    \mathbf{B}^T\Sigma_r^{-1}(\mathbf{c} + \mathbb{E}[\boldsymbol\omega]) + \Sigma_\eta^{-1}\mathbb{E}[\boldsymbol\beta] 
    \end{bmatrix} \nonumber \\
    &=& \Sigma \begin{bmatrix} 
    \mathbf{A}^T\Sigma_r^{-1}\mathbf{c} \\
    \mathbf{B}^T\Sigma_r^{-1}\mathbf{c} 
    \end{bmatrix}  = \boldsymbol\mu.
\end{eqnarray}
Comparing Eqs.~\eqref{eq:rPICKLE_mean} and \eqref{eq:linear_posterior_mode} yields $\boldsymbol\mu =  \boldsymbol\mu_{post}$. Next, we prove that $\Sigma = \Sigma_{post}$ is the covariance of $[\boldsymbol\xi^*,\boldsymbol\eta^*]^T$. 
The covariance of $\boldsymbol\zeta = [\boldsymbol\xi^*,\boldsymbol\eta^*]^T$ can be calculated as 
\begin{eqnarray}\label{eq:rPICKLE_cov2}
&&\mathbb{E}[(\boldsymbol\zeta - \mathbb{E}[\boldsymbol\zeta])(\boldsymbol\zeta - \mathbb{E}[\boldsymbol\zeta])^T] \\
&=& \Sigma\mathbb{E} \left[
\begin{bmatrix} 
     \mathbf{A}^T\Sigma_r^{-1}\boldsymbol\omega  +  \Sigma_\xi^{-1}\boldsymbol\alpha  \\
     \mathbf{B}^T\Sigma_r^{-1}\boldsymbol\omega +  \Sigma_\eta^{-1}\boldsymbol\beta
    \end{bmatrix}
\begin{bmatrix}
\mathbf{A}^T\Sigma_r^{-1}\boldsymbol\omega  +  \Sigma_\xi^{-1}\boldsymbol\alpha & \mathbf{B}^T\Sigma_r^{-1}\boldsymbol\omega +  \Sigma_\eta^{-1}\boldsymbol\beta
\end{bmatrix}
\right] \Sigma \nonumber \\
&=& 
\Sigma
\begin{bmatrix} 
      M_{11} &  M_{12} \\
      M_{21} &  M_{22}
\end{bmatrix}
\Sigma,  \nonumber 
\end{eqnarray}
where 
\begin{align}
    M_{11} &= \mathbf{A}^T\Sigma_r^{-1}\Sigma_r^{-1}\mathbf{A} \mathbb{E}[\boldsymbol\omega\boldsymbol\omega^T]  +  \Sigma_\xi^{-1}\Sigma_\xi^{-1} \mathbb{E} [\boldsymbol\alpha\boldsymbol\alpha^T] \nonumber \\
    M_{12} &= \mathbf{B}^T\Sigma_r^{-1}\Sigma_r^{-1}\mathbf{A} \mathbb{E}[\boldsymbol\omega\boldsymbol\omega^T] \nonumber \\
    M_{21} &= \mathbf{A}^T\Sigma_r^{-1}\Sigma_r^{-1}\mathbf{B} \mathbb{E}[\boldsymbol\omega\boldsymbol\omega^T] \nonumber \\
    M_{22} &= \mathbf{B}^T\Sigma_r^{-1}\Sigma_r^{-1}\mathbf{B} \mathbb{E}[\boldsymbol\omega\boldsymbol\omega^T]  +  \Sigma_\eta^{-1}\Sigma_\eta^{-1}\mathbb{E} [\boldsymbol\beta\boldsymbol\beta^T]. \nonumber
\end{align}
For $\Sigma_\omega = \mathbb{E}[\boldsymbol\omega\boldsymbol\omega^T] = \Sigma_r$, $\Sigma_\alpha = \mathbb{E} [\boldsymbol\alpha\boldsymbol\alpha^T] = \Sigma_\xi$, and $\Sigma_\beta = \mathbb{E} [\boldsymbol\beta\boldsymbol\beta^T] = \Sigma_\eta$, Eq.~\eqref{eq:rPICKLE_cov2} reduces to
\begin{equation}
\mathbb{E}[(\boldsymbol\zeta - \boldsymbol\mu)(\boldsymbol\zeta - \boldsymbol\mu)^T]= \Sigma\Sigma^{-1}\Sigma = \Sigma_{post}.
\end{equation}
This ends the proof that, for the linear $\mathcal{R}(\mathbf{x}; \boldsymbol\xi, \boldsymbol\eta)$, the ensemble mean and covariance of $\boldsymbol\zeta$ in rPICKLE are equal to the mean and covariance of posterior distribution given by the Bayes rule. 

Finally, we note that in rPICKLE,  $N_{ens}$ samples of $\boldsymbol\zeta$ are obtained and used to compute the \emph{sample} mean and covariance of $\boldsymbol\zeta$. As $N_{ens}\to \infty$, the sample mean and covariance converge to their ensemble counterparts.  

\subsubsection{Metropolis rejection}\label{sec:metropolis}
When the residual $\mathcal{R}(\mathbf{x}; \boldsymbol{\xi}, \boldsymbol{\eta})$ does not have a linear form, rPICKLE samples may deviate from the true posterior. This deviation, however, can be corrected with a Metropolis procedure. 

Recall that the random noise vectors $\boldsymbol\omega$, $\boldsymbol\alpha$, and $\boldsymbol\beta$  have an independent joint distribution:
\begin{eqnarray}
    p(\boldsymbol\omega, \boldsymbol\alpha, \boldsymbol\beta) \propto \exp(-\boldsymbol\omega^T \Sigma^{-1}_\omega \boldsymbol\omega -  \boldsymbol\alpha^T \Sigma^{-1}_\alpha \boldsymbol\alpha -\boldsymbol\beta^T \Sigma^{-1}_\beta \boldsymbol\beta).
\end{eqnarray}
Next, we define  a random vector $\boldsymbol\delta = \mathcal{R}(\mathbf{x}; \boldsymbol{\xi}^*, \boldsymbol{\eta}^*) - \boldsymbol\omega$ and assume that there exists an invertible map $\mathcal{G}: (\boldsymbol\omega, \boldsymbol\alpha, \boldsymbol\beta) \rightarrow (\boldsymbol\delta, \boldsymbol\xi^*, \boldsymbol\eta^*)$, where 
$\boldsymbol\xi^*$ and $\boldsymbol\eta^*$ minimize the rPICKLE loss function $L^r$.
Because $p(\boldsymbol\omega, \boldsymbol\alpha, \boldsymbol\beta)$ is known, the joint distribution of $(\boldsymbol\delta, \boldsymbol\xi^*, \boldsymbol\eta^*)$ can be computed as
\begin{eqnarray}\label{eq:inverse_transformation}
    f(\boldsymbol\delta, \boldsymbol\xi^*, \boldsymbol\eta^*) = p(\boldsymbol\omega, \boldsymbol\alpha, \boldsymbol\beta)|\mathrm{det}(\mathbf{J})|,
\end{eqnarray}
where $f$ is the probability density function of $(\boldsymbol\delta, \boldsymbol\xi^*, \boldsymbol\eta^*)$ and $\mathbf{J}$ is the Jacobian of the map $\mathcal{G}$ defined as
\begin{equation}
    \mathbf{J} := \frac{\partial(\boldsymbol\omega, \boldsymbol\alpha, \boldsymbol\beta) }{\partial (\boldsymbol\delta, \boldsymbol\xi^*, \boldsymbol\eta^*)}.
\end{equation} 
 To find this Jacobian, we note that for a general residual operator $\mathcal{R}$, 
\begin{eqnarray}\label{eq:randomized_loss_grad_nonlinear}
\frac{\partial L^r}{\partial \boldsymbol\xi}
=  \boldsymbol\delta^T\Sigma_r^{-1}\frac{\partial \mathcal{R}}{\partial \boldsymbol\xi}  +   (\boldsymbol\xi - \boldsymbol\alpha)^T\Sigma_\xi^{-1}  \\
\frac{\partial L^r}{\partial \boldsymbol\eta}
=  \boldsymbol\delta^T\Sigma_r^{-1}\frac{\partial \mathcal{R}}{\partial \boldsymbol\eta} +   (\boldsymbol\eta - \boldsymbol\beta)^T\Sigma_\eta^{-1}
\end{eqnarray}
and $(\boldsymbol\xi^*, \boldsymbol\eta^*)$ is the solution of $(\frac{\partial L^r}{\partial \boldsymbol\xi}, \frac{\partial L^r}{\partial \boldsymbol\eta})^T = \mathbf{0}$. Then, $\mathcal{G}$ is implicitly expressed as:
\begin{eqnarray}\label{eq:rv_mapping}
\begin{cases}
    \boldsymbol\alpha = \Sigma_\xi \frac{\partial \mathcal{R}}{\partial \boldsymbol\xi}^T\Sigma_r^{-1}\boldsymbol\delta + \boldsymbol\xi^* \\
    \boldsymbol\beta = \Sigma_\eta \frac{\partial \mathcal{R}}{\partial \boldsymbol\eta}^T\Sigma_r^{-1}\boldsymbol\delta + \boldsymbol\eta^* \\
    \boldsymbol\omega = \mathcal{R}(\mathbf{x}; \boldsymbol{\xi}^*, \boldsymbol{\eta}^*) - \boldsymbol\delta .
\end{cases}
\end{eqnarray}
The explicit form of the Jacobian can then be obtained as
\begin{eqnarray}
        \mathbf{J} &:=& \begin{bmatrix} 
        \frac{\partial\boldsymbol\omega}{\partial\boldsymbol\delta} & \frac{\partial\boldsymbol\omega}{\partial\boldsymbol\xi^*} & \frac{\partial\boldsymbol\omega}{\partial\boldsymbol\eta^*}  \\
        \frac{\partial\boldsymbol\alpha}{\partial\boldsymbol\delta}  & \frac{\partial\boldsymbol\alpha}{\partial\boldsymbol\xi^*} & \frac{\partial\boldsymbol\alpha}{\partial\boldsymbol\eta^*} \\
        \frac{\partial\boldsymbol\beta}{\partial\boldsymbol\delta}  & \frac{\partial\boldsymbol\beta}{\partial\boldsymbol\xi^*} & \frac{\partial\boldsymbol\beta}{\partial\boldsymbol\eta^*} 
    \end{bmatrix} \nonumber  \\
        &=& \begin{bmatrix} 
        -\mathbf{I} & \frac{\partial \mathcal{R}}{\partial \boldsymbol\xi} & \frac{\partial \mathcal{R}}{\partial \boldsymbol\eta} \\
        \Sigma_\xi \frac{\partial \mathcal{R}}{\partial \boldsymbol\xi}^T\Sigma_r^{-1}  &   \Sigma_\xi (\frac{\partial^2 \mathcal{R}}{\partial \boldsymbol\xi^2})^T\otimes (\Sigma_r^{-1}\boldsymbol\delta) &  \Sigma_\xi (\frac{\partial^2 \mathcal{R}}{\partial \boldsymbol\xi\partial\boldsymbol\eta})^T\otimes (\Sigma_r^{-1}\boldsymbol\delta) \\
        \Sigma_\eta \frac{\partial \mathcal{R}}{\partial \boldsymbol\eta}^T\Sigma_r^{-1} &  \Sigma_\eta (\frac{\partial^2 \mathcal{R}}{\partial \boldsymbol\eta\partial\boldsymbol\xi})^T\otimes (\Sigma_r^{-1}\boldsymbol\delta) & \Sigma_\eta (\frac{\partial^2 \mathcal{R}}{\partial \boldsymbol\eta^2})^T\otimes (\Sigma_r^{-1}\boldsymbol\delta) 
    \end{bmatrix} \nonumber \\
     &=& \begin{bmatrix} 
        -\mathbf{I} & \frac{\partial \mathcal{R}}{\partial \boldsymbol\zeta} \\
        \widehat{\Sigma}\frac{\partial \mathcal{R}}{\partial \boldsymbol\zeta}^T\Sigma_r^{-1}  &  \mathbf{I} + \widehat{\Sigma} (\frac{\partial^2 \mathcal{R}}{\partial \boldsymbol\zeta^2})^T\otimes (\Sigma_r^{-1}\boldsymbol\delta) 
    \end{bmatrix}, 
\end{eqnarray}
where $\otimes$ denotes the tensor product and
\begin{eqnarray}
 \widehat{\Sigma} = \begin{bmatrix} 
        \Sigma_\xi & \mathbf{0} \\
        \mathbf{0} & \Sigma_\eta
    \end{bmatrix} .
\end{eqnarray}
The determinant of $\mathbf{J}$ is 
\begin{eqnarray}\label{eq:det_jacobian}
    |\mathrm{det}(\mathbf{J})|  = \left| \mathrm{det} \left[\mathbf{I} + \widehat{\Sigma} (\frac{\partial^2 \mathcal{R}}{\partial \boldsymbol\zeta^2})^T\otimes(\Sigma_r^{-1}\boldsymbol\delta) + \widehat{\Sigma}(\frac{\partial \mathcal{R}}{\partial \boldsymbol\zeta})^T\Sigma_r^{-1}\frac{\partial \mathcal{R}}{\partial \boldsymbol\zeta}\right] \right|.
\end{eqnarray}

With the expression for $|\mathrm{det}(\mathbf{J})|$, we formulate the Metropolis rejection method for the rPICKLE that we encapsulate in Algorithm \ref{alg:randomized_pickle}. 
In this method, the samples $\boldsymbol\omega_k$, $\boldsymbol\alpha_k $, and $\boldsymbol\beta_k$ $(k=1,...,N_{ens})$ are independently drawn.
Then, the samples $\boldsymbol\zeta_k = (\boldsymbol\xi_k^*, \boldsymbol\eta_k^*)$ of the posterior distribution are found as the solution of the rPICKLE minimization problem  \eqref{eq:randomized_PICKLE_loss} for $(\boldsymbol\omega, \boldsymbol\alpha, \boldsymbol\beta)=(\boldsymbol\omega_k, \boldsymbol\alpha_k, \boldsymbol\beta_k)$.  

The first sample is automatically accepted. The acceptance or rejection of the $\boldsymbol\zeta_k$ ($k>1$) sample can be done according to the independent Metropolis-Hastings (IMH) method \cite{tierney1994markov} with the acceptance ratio $\alpha$:
\begin{eqnarray}\label{eq:metropolis-hasting}
    \alpha(\boldsymbol\zeta_{k-1}, \boldsymbol\zeta_{k}) = \min \left\{ 1, \frac{\pi(\boldsymbol\zeta_{k})q(\boldsymbol\zeta_{k-1})}{\pi(\boldsymbol\zeta_{k-1})q(\boldsymbol\zeta_{k})} \right\},
\end{eqnarray}
where $\pi(\cdot)$ is a function proportional to the Bayesian posterior and  $q(\cdot)$ is the proposal density function defined as
\begin{align}\label{eq:proposal_distribution}
    q(\boldsymbol\zeta)  &= \int f(\boldsymbol\delta, \boldsymbol\xi^*, \boldsymbol\eta^*) d\boldsymbol\delta \nonumber \\
    &= \int p(\boldsymbol\omega, \boldsymbol\alpha, \boldsymbol\beta)|\mathrm{det}(\mathbf{J})| d\boldsymbol\delta,
\end{align}
where the mapping from $(\boldsymbol\omega, \boldsymbol\alpha, \boldsymbol\beta)$ to $(\boldsymbol\xi, \boldsymbol\eta, \boldsymbol\delta)$ is given by Eq.~\eqref{eq:rv_mapping}. In most cases, this integration is computationally prohibitive. Following \cite{wang2018rmap}, we use an approximate expression for the acceptance ratio:
\begin{eqnarray}\label{eq:approximate_mh}
    \alpha(\boldsymbol\zeta_{k-1}, \boldsymbol\zeta_{k}) \approx \min \left\{ 1, \frac{ |\mathrm{det}(\mathbf{J}_{\zeta_{k}})|^{1/2}}{|\mathrm{det}(\mathbf{J}_{\zeta_{k-1}})|^{1/2}} \right\}.
\end{eqnarray}
For each $k$ sample, the number $g$ is drawn from the continuous uniform distribution $U(0,1)$ and the kth sample of $\boldsymbol\zeta$ is accepted if $g \leq \alpha(\boldsymbol\zeta_k, \boldsymbol\zeta_{k-1})$. Otherwise, the $k$th sample is rejected, i.e., $\boldsymbol\zeta_k$ is replaced with $\boldsymbol\zeta_{k-1}$. 

 It should be noted that in this IMH algorithm, the proposal function $q$ is not conditioned on the previous sample as in the HMC method. However, the acceptance ratio depends on the previous sample, and the total transition obeys the Markov property. In the case of the linear $\mathcal{R}$  model considered in Section \ref{sec:rpickle_linear}, every proposal is accepted because the determinant of the Jacobian is a constant and independent of $\boldsymbol{\zeta}$.

\begin{algorithm}
\caption{Randomized PICKLE Algorithm.}
\label{alg:randomized_pickle}
\begin{algorithmic}[1]
\Require{number of samples $N_{ens}$}
\For{$k = 1, \dots, N_{ens}$}
    \State Sample random noises $\boldsymbol\omega_k \sim \mathcal{N}(\mathbf{0}, \Sigma_r)$, $\boldsymbol\alpha_k \sim \mathcal{N}(\mathbf{0}, \Sigma_\xi)$, $\boldsymbol\beta_k \sim \mathcal{N}(\mathbf{0}, \Sigma_\eta)$
    \State Propose $\boldsymbol\zeta^\prime = (\boldsymbol\xi^*, \boldsymbol\eta^*)$ by optimizing the randomized loss \eqref{eq:randomized_PICKLE_loss}
    \State Calculate acceptance ratio $\alpha(\boldsymbol\zeta^\prime, \boldsymbol\zeta_{k-1}) = \min \left\{ 1, \frac{|\det(\mathbf{J}_{\boldsymbol\zeta^\prime})|^{1/2}}{|\det(\mathbf{J}{\zeta_{k-1}})|^{1/2}} \right\}$
    \State Sample $g \sim U(0, 1)$
    \If{$g \leq \alpha(\boldsymbol\zeta^\prime, \boldsymbol\zeta_{k-1})$}
        \State $\boldsymbol\zeta_k \gets \boldsymbol\zeta^\prime$ (accept $\boldsymbol\zeta^\prime$ and set it as the new state)
    \Else
        \State $\boldsymbol\zeta_k \gets \boldsymbol\zeta_{k-1}$ (reject $\boldsymbol\zeta^\prime$ and copy the old state forward)
    \EndIf
\EndFor
\end{algorithmic}
\end{algorithm}

\section{Application to the Groundwater Flow Hanford Site Model}\label{sec:Hanford_problem_formulation}

\subsection{Governing Equations}\label{sec:GWF}
We test the rPICKLE method for a two-dimensional steady-state groundwater flow model, described by the boundary value problem (BVP)
\begin{eqnarray}\label{eq:GWF_eqns}
    \nabla \cdot (e^{y(\mathbf{x})}\nabla u(\mathbf{x})) &=& 0 \quad\quad \mathbf{x} \in \Omega \label{eq:eq21} \\
    e^{y(\mathbf{x})}\nabla u(\mathbf{x}) \cdot n(\mathbf{x}) &=& q_\mathcal{N}(\mathbf{x})  \quad \mathbf{x} \in \partial\Omega_\mathcal{N} \label{eq:eq22} \\
        u(\mathbf{x}) &=& u_\mathcal{D}(\mathbf{x})  \quad \mathbf{x} \in \partial\Omega_\mathcal{D},
 \label{eq:eq23}
\end{eqnarray}
where $T(\mathbf{x}) = e^{y(\mathbf{x})}$ is the transmissivity field, $u(\mathbf{x})$ is the hydraulic head, $\partial\Omega_\mathcal{N}$ is the known Neumann boundary,  $\partial\Omega_\mathcal{D}$ is the known Dirichlet boundary, $\partial\Omega_\mathcal{N} \cap \partial\Omega_\mathcal{D} = \varnothing$, $q_\mathcal{N}(\mathbf{x})$ is the prescribed normal flux at the Neumann boundary, $n(\mathbf{x})$ is the unit normal vector, and $u_\mathcal{D}(\mathbf{x})$ is the prescribed hydraulic head at the Dirichlet boundary. It is common to treat $y(\mathbf{x})$ as a realization of a correlated Gaussian field. Also, it was found that solving an inverse problem for $y$ instead of $T$ decreases the level of non-convexity in optimization problems \cite{Carrera1986estimation}. 

We use a cell-centered finite volume (FV) discretization for Eqs.~\eqref{eq:eq21}--\eqref{eq:eq23} and the two-point flux approximation (TPFA) to compute residuals in the rPICKLE objective function.
The numerical domain $\Omega$ is discretized into $N$ finite volume cells, and $u(\mathbf{x})$ and $y(\mathbf{x})$ are approximated with $\mathbf{u} = \{ u_1, ..., u_N\}$ and $\mathbf{y} = \{y_1, ..., y_N \}$ vectors of their respective values evaluated at the cell centers $\{\mathbf{x}_1, ..., \mathbf{x}_N \}$. For the inverse problem, we assume there are $N^{\text{obs}}_y$ observations of $y$, $\mathbf{y}^{\text{obs}} = \{y_i^{\text{obs}}\}_{i = 1}^{N_y^{\text{obs}}}$, and $N_u^{\text{obs}}$ observations of $u$, $\mathbf{u}^{\text{obs}} = \{u^{\text{obs}}_{i}\}_{i = 1}^{N_u^{\text{obs}}}$. 

\subsection{Hanford Site Case Study}\label{sec:hanford_case}

\begin{figure}[ht]
    \centering
    \includegraphics{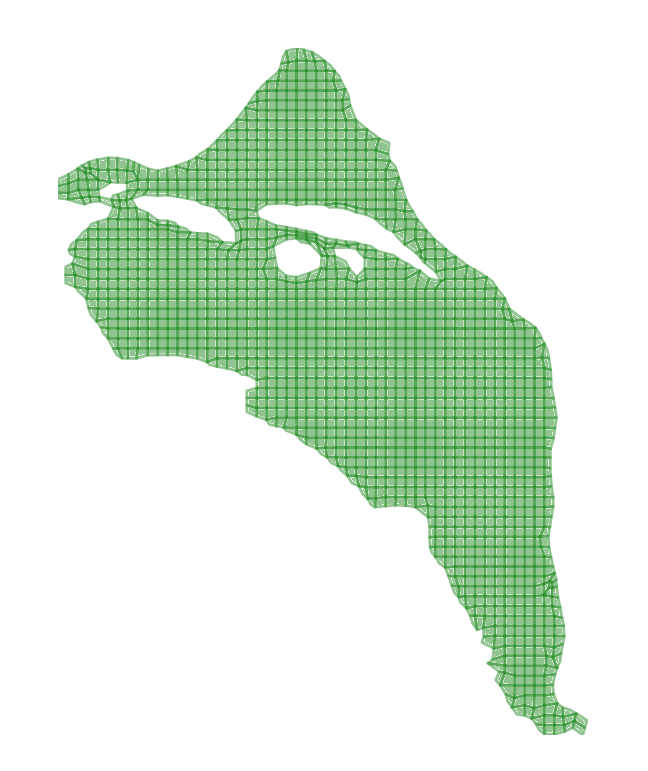}
    \caption{
    A quasi-uniform coarse mesh with $N = 1475$ cells for computing PDE residuals in the rPICKLE model.  The figure is adapted from \cite{Yeung2021PICKLE}.
    }
    \label{fig:hanford_mesh}
\end{figure}

The rPICKLE method is tested for estimating subsurface parameters at the Hanford Site, a United States Department of Energy site situated on the Columbia River in Washington State. We use Eqs.~\eqref{eq:eq21}--\eqref{eq:eq23} to describe the two-dimensional (depth-averaged) groundwater flow at the Hanford Site. 
The ground truth conductivity and transmissivity fields are taken from a previous calibration study reported in \cite{cole2001transient}. Following  \cite{Yeung2021PICKLE}, we employ the unstructured quadrilateral mesh with $N = 1475$ cells (see Figure~\ref{fig:hanford_mesh}) to compute PDE residuals. 
The rPICKLE estimates of the posterior distribution of log-transmissivity are compared with that of the HMC and, for the mode of the distribution, with PICKLE. 

\begin{figure}[!ht]
    \centering
    \begin{subfigure}[]{0.4\textwidth}
        \includegraphics[width=\textwidth]{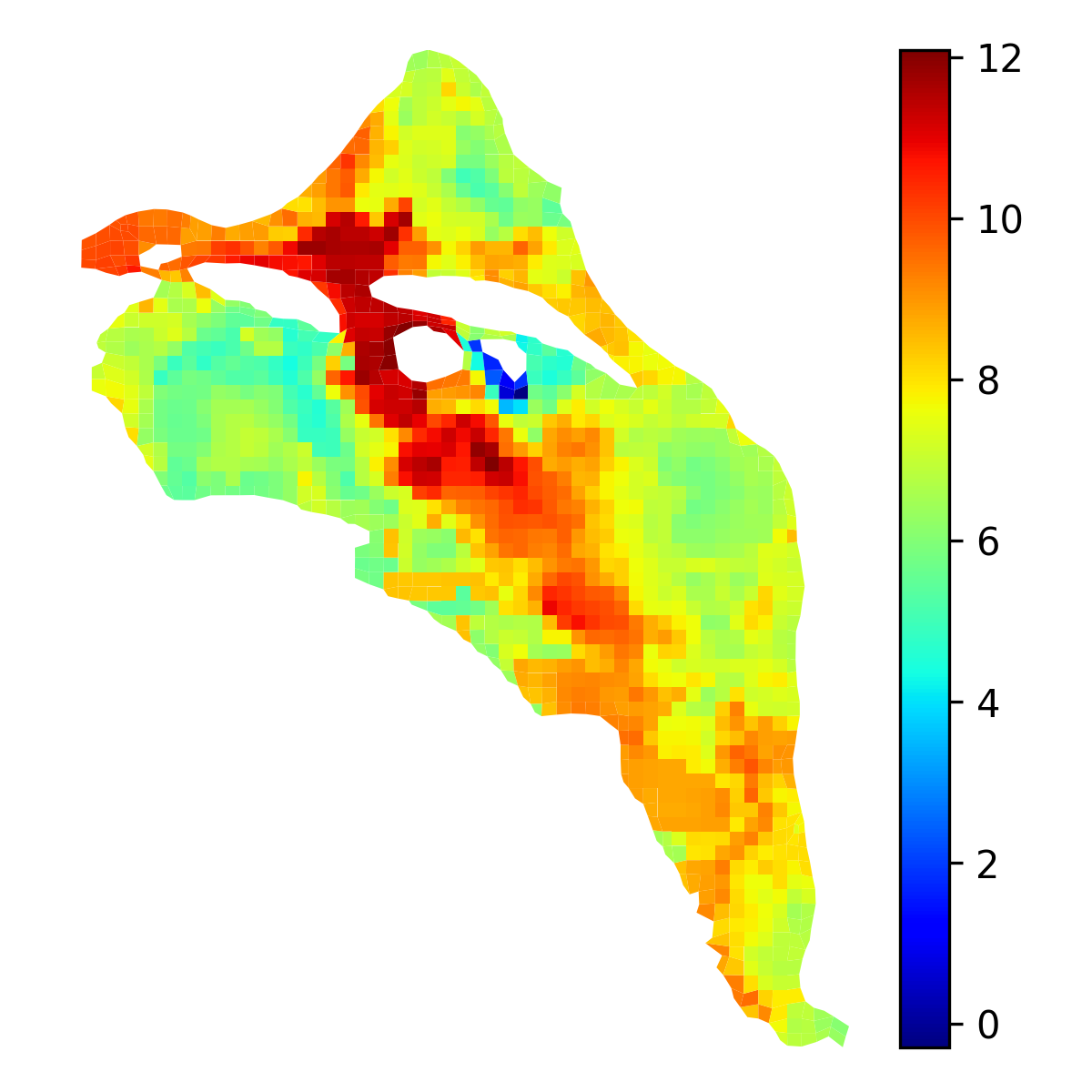}
        \caption{$y_{\mathrm{HD}}$}
        \label{fig:Yref}
    \end{subfigure}
    \hspace{20pt}
    \begin{subfigure}[]{0.4\textwidth}
        \includegraphics[width=\textwidth]{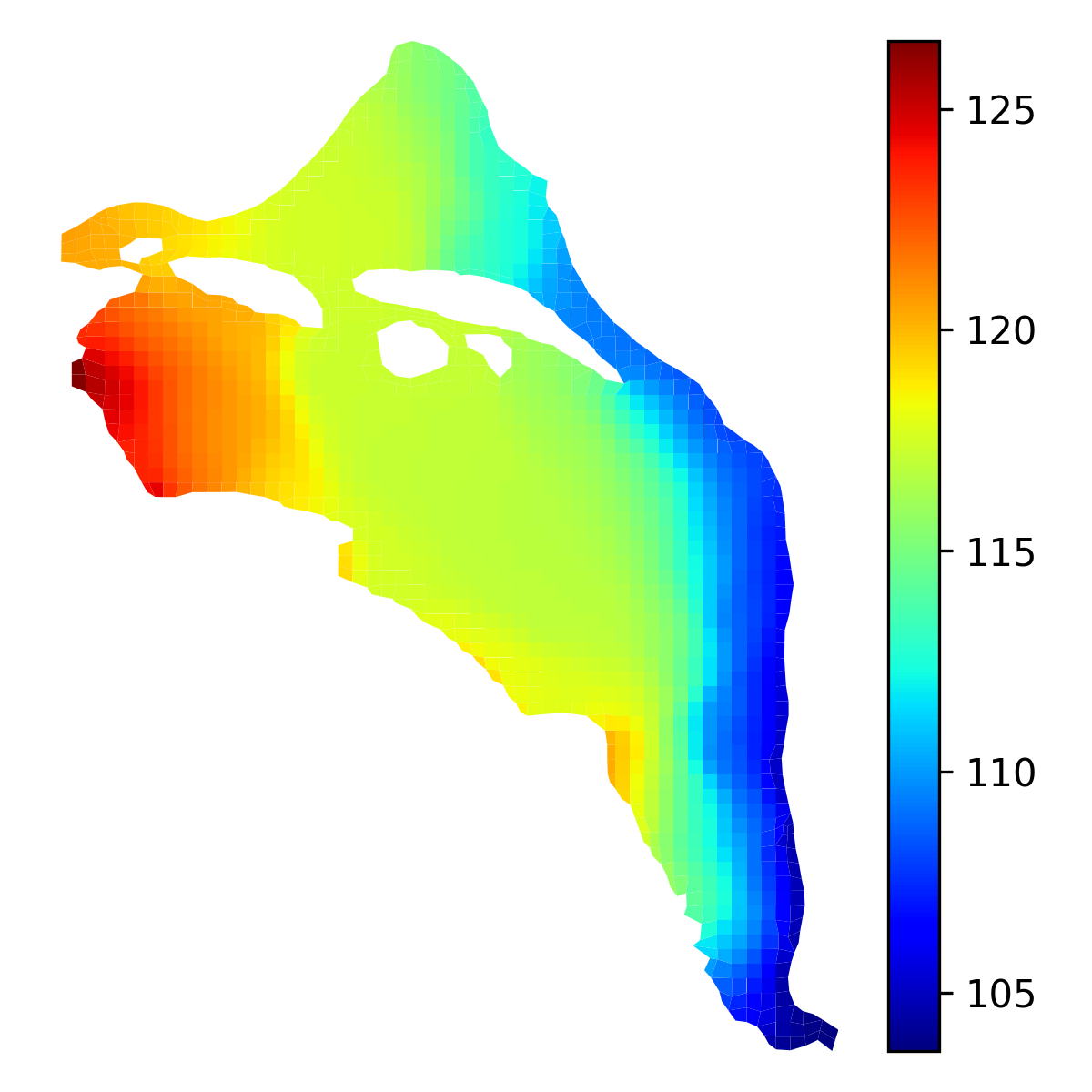}
        \caption{$u_{\mathrm{HD}}$}
        \label{fig:uref}
    \end{subfigure}
    
    \vspace{10pt}
    
    \begin{subfigure}[]{0.4\textwidth}
        \includegraphics[width=\textwidth]{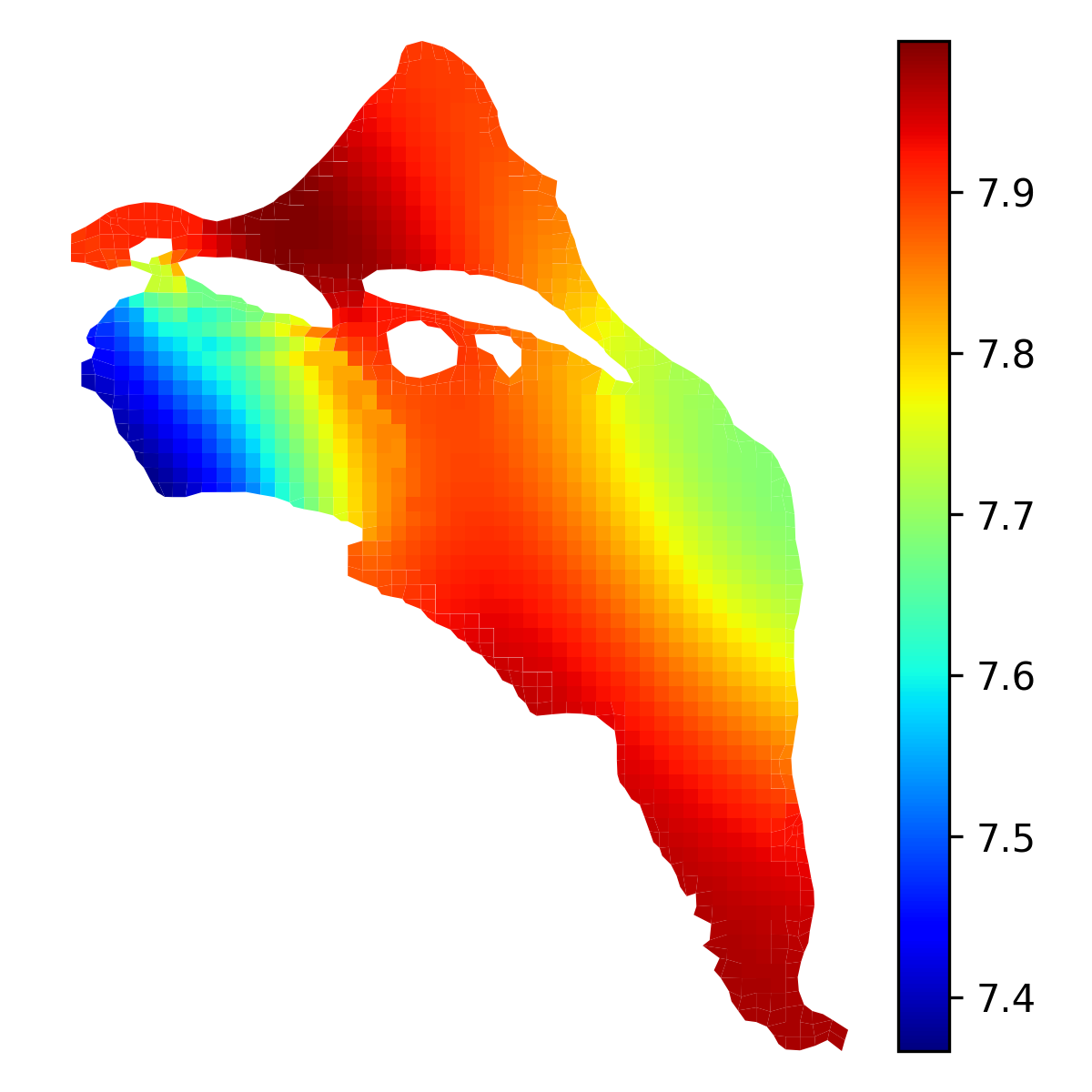}
        \caption{$y_{\mathrm{LD}}$}
        \label{fig:Ysmooth}
    \end{subfigure}
    \hspace{20pt}
    \begin{subfigure}[]{0.4\textwidth}
        \includegraphics[width=\textwidth]{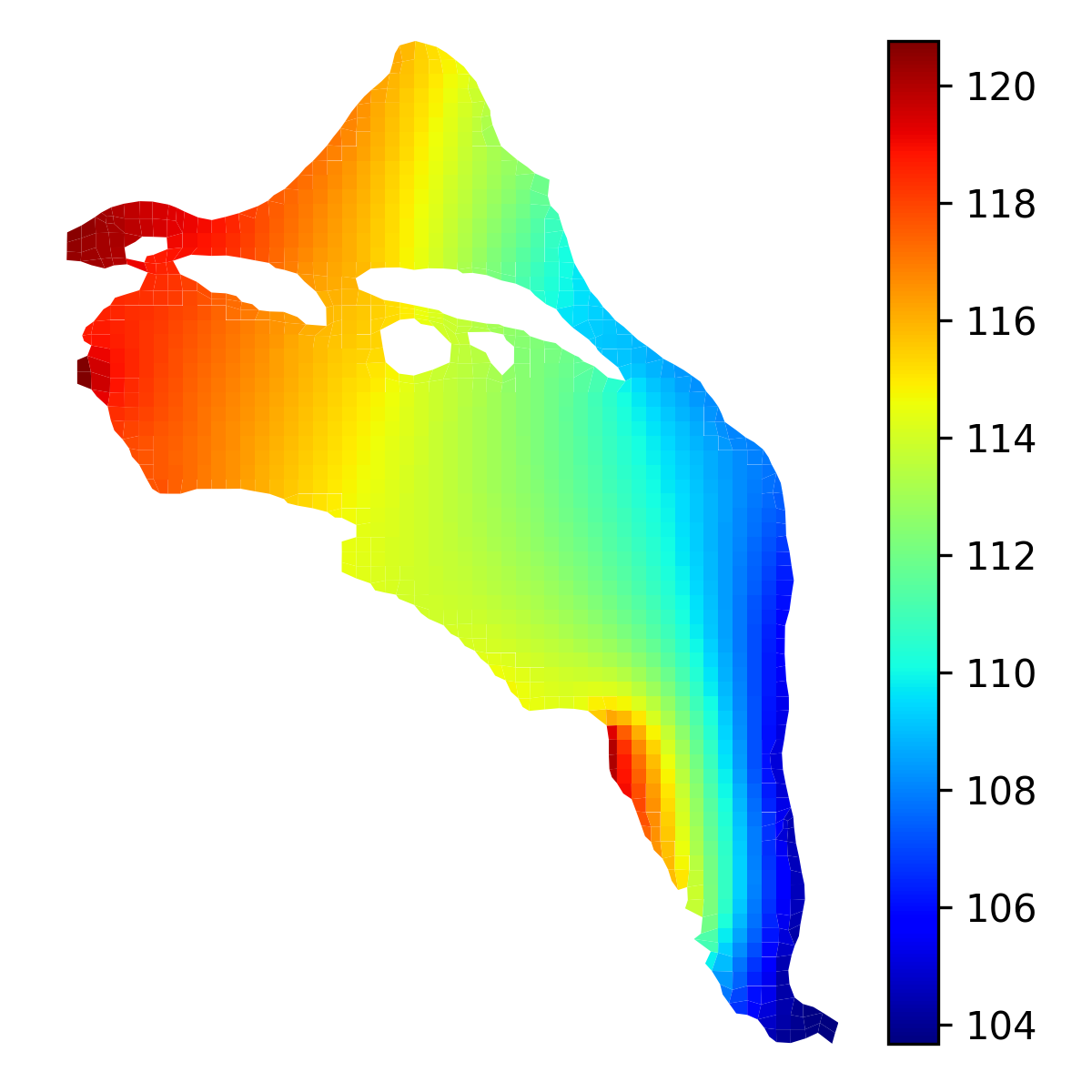}
        \caption{$u_{\mathrm{LD}}$}
        \label{fig:usmooth}
    \end{subfigure}
    \caption{(a) The reference log-transmissivity field $y_{\mathrm{HD}}$ and (b) the reference hydraulic head field $u_{\mathrm{HD}}$ corresponding to $y_{\mathrm{HD}}$ obtained by solving Eqs. \eqref{eq:eq21}--\eqref{eq:eq23} with the finite volume method. (c) The low-dimensional field $y_{\mathrm{LD}}$ obtained by iterative local averaging of $y_{\mathrm{HD}}$ and  (d) the hydraulic head field  $u_{\mathrm{LD}}$ corresponding to $y_{\mathrm{LD}}$.}
     \label{fig:ref_field}
\end{figure}

Figure \ref{fig:ref_field}a shows the ground truth $y(\mathbf{x})$ field. To retain 95\% of the variance of this field, a 1000-term CKLE ($N_{\xi} = 1000$) is needed. We denote the ground truth $y$ field as the high-dimensional or $y_{\mathrm{HD}}$ field. In the following, we will show that such high dimensionality in combination with relatively small $\sigma_r$ reduces the efficiency of HMC, resulting in prohibitively large computational costs.

To enable a comparison between HMC and rPICKLE,  we generate a lower-dimensional (smoother) $y$ field $y_{\mathrm{LD}}$ using iterative local averaging of $y_{\mathrm{HD}}$. Such averaging reduces the variance and increases the correlation length of the field \cite{Tart2017WRR} and, therefore, reduces the dimensionality of the inverse problem. At the $k$th iteration, for FV element $i$, we calculate the geometric mean of $y^{(k)}_j$  over the adjacent  $j$ elements and assign this value to $y^{(k+1)}_i$. Here, $y_{\mathrm{HD}} = y^{(k = 0)}$ and $y_{\mathrm{LD}} = y^{(k=30)}$. We find that the ten-term CKLE of $y_{\mathrm{LD}}$  retains 95\% of the total variance of this field.

We generate the hydraulic head fields $u_{\mathrm{HD}}$ and $u_{\mathrm{LD}}$ corresponding to the $y_{\mathrm{HD}}$ and $y_{\mathrm{LD}}$ fields by solving the Darcy flow equation on the mesh shown in Figure \ref{fig:hanford_mesh} with the Dirichlet and Neumann boundary conditions from the calibration study \cite{cole2001transient} using the finite volume method as described in \cite{Yeung2021PICKLE}. 

Figure \ref{fig:ref_field} shows the $y_{\mathrm{HD}}$ and $y_{\mathrm{LD}}$ fields and the corresponding $u$ fields. In the PICKLE representation, we set $N_\eta = 1000$ and $N_\eta = 5$ in CKLEs of $u_{\mathrm{HD}}$ and $u_{\mathrm{LD}}$, respectively, to retain no less than $95 \%$ of the total variance of the hydraulic head field. 

In the Hanford Site calibration study \cite{cole2001transient}, coordinates of $558$ wells are provided with some of these wells located within the same cells. In this work, we only allow to have a single well per cell resulting in a total of $323$ wells. We assume that  $N^{\text{obs}}_u$ of $u$ and $N^{\text{obs}}_y$ of $y$ are available. The locations of measurements are randomly selected from the well locations. 
The measurements of $y$ and $u$ at the selected locations are drawn from the $y_{\mathrm{HD}}$ and $u_{\mathrm{HD}}$ fields for the high-dimensional case and the $y_{\mathrm{LD}}$ and $u_{\mathrm{LD}}$ fields for the low-dimensional case. 

The HMC and randomized PICKLE algorithms are implemented in TensorFlow 2 and TensorFlow-Probability. All simulations are performed on a workstation with Intel$^{\circledR}$ Xeon$^{\circledR}$ Gold 6230R CPU @ 2.10GHz.

\subsection{Prior Mean and Covariance Models of $y$ and $u$}\label{sec:mean_covariance}

Following \cite{Yeung2021PICKLE}, for the log-transmissivity field $y(\mathbf{x})$, we select the 5/2-Matérn type prior covariance kernel:
    \begin{eqnarray}\label{eq:cov_kernel}
        C_y(\mathbf{x}, \mathbf{x}^\prime;\theta) = \sigma^2(1 + \sqrt{5}\frac{\| \mathbf{x} - \mathbf{x}^\prime\|}{l} +\frac{5}{3}\frac{\| \mathbf{x} - \mathbf{x}^\prime\|^2}{l^2} )\exp(-\sqrt{5}\frac{\| \mathbf{x} - \mathbf{x}^\prime\|}{l}) ,
    \end{eqnarray}
where $\sigma$ and $l$ are the standard deviation and correlation length of $y(\mathbf{x})$ that, for a given set of $y$ field measurements, are found by minimizing the negative marginal log-likelihood function \cite{rasmussen2006gaussian}. Then, the CKLE of $y$ is constructed by first computing the mean and covariance of $y$ conditioned on the $y$ measurements using GPR (Eqs.~\eqref{eq:gpr_k} and \eqref{eq:gpr_cov}), and then evaluating the eigenvalues and eigenfunctions by solving the eigenvalue problem \eqref{eq:eigenvalue_problem_Y}. 

Next, we generate $N^{\text{MC}}$ number of realizations of the stochastic $y$ field by independently sampling $ \{ \xi_i \}_{i = 1}^{N_{\xi}}$ from the normal distribution and solving Eqs.~\eqref{eq:GWF_eqns}--\eqref{eq:eq23} for each realization of $y$ using the FV method described in \cite{Yeung2021PICKLE}. The ensemble of $u$ solutions is used to compute the (ensemble) mean and covariance of $u$ using Eqs.~\eqref{eq:mean_u} and \eqref{eq:cov_u}. Then, the mean and covariance of $u$ are conditioned on $u$ observations using the GPR equations \eqref{eq:gpr_u}. Finally, the CKLE of $u$ in Eq. \eqref{eq:CKLE_u} is constructed by performing the eigenvalue decomposition of the conditional covariance of $u$, i.e., by solving the eigenvalue problem similar to Eq.~\eqref{eq:eigenvalue_problem_Y}. In this work, we set $N^{\text{MC}} = 5,000$. 

\section{Numerical Results}\label{sec:numerical_results}
\subsection{Low-Dimensional Problem}\label{sec:low-dimensional}

We first present results for the low-dimensional case with the ground truth $y$ field given by $y_\mathrm{LD}$. Here, we assume that 10 $y_\mathrm{LD}$ and $u_\mathrm{LD}$ observations are available. The locations of observations are randomly selected from the Hanford Site well locations. We use the HMC-sampled posterior distribution of $\boldsymbol\xi$ (and $y$), and the PICKLE-estimated MAP to benchmark the rPICKLE method with and without Metropolization. As stated earlier, we set $\sigma^2_\xi= \sigma^2_\eta = 1$. The value of $\sigma^2_r$ is chosen to maximize the LPP of the posterior. However, the joint problem of minimizing the rPICKLE loss over $\boldsymbol\xi$ and $\boldsymbol\eta$ and maximizing the LPP over $\sigma^2_r$ is computationally challenging. Instead, we compute the inverse solutions for several values of $\sigma^2_r$ in the range $[10^{-5}, 1]$ and select the value producing the largest LPP.  
We also study the effect of $\sigma^2_r$ on uncertainty in the inverse solution and the performance of HMC and rPICKLE methods. 

For rPICKLE, we compute $N_{ens}=10^4$ samples from the posterior of $\boldsymbol\xi$ by solving the rPICKLE minimization problem \eqref{eq:rPICKLE_loss} for $N_{ens}$ different realizations of $\boldsymbol\omega$, $\boldsymbol\alpha$, and $\boldsymbol\beta$. We find Section \ref{sec:convergence}, we show that this number of samples is sufficient for the first two moments of the rPICKLE-sampled posterior distribution to converge. For HMC, we initialize three dispersed Markov chains over the parameter space. We set the number of HMC burn-in steps to $2 \times 10^{4}$ and the number of samples to $N_{ens}=10^4$ as the stopping criterion. We adopt the No-U-Turn Sampler (NUTS) HMC method \cite{hoffman2014NUTS}, which adaptively determines the number of integration steps taken within one HMC iteration. Furthermore, we use the dual averaging algorithms \cite{hoffman2014NUTS} to determine the optimal step size for NUTS to maintain a reasonable acceptance rate. Following \cite{betancourt2015optimizing}, we set the target acceptance rate to $70\%$. 

\begin{figure}[!htb]
    \centering
    \begin{subfigure}{0.45\textwidth}
    \centering
        \includegraphics[width=0.95\textwidth]{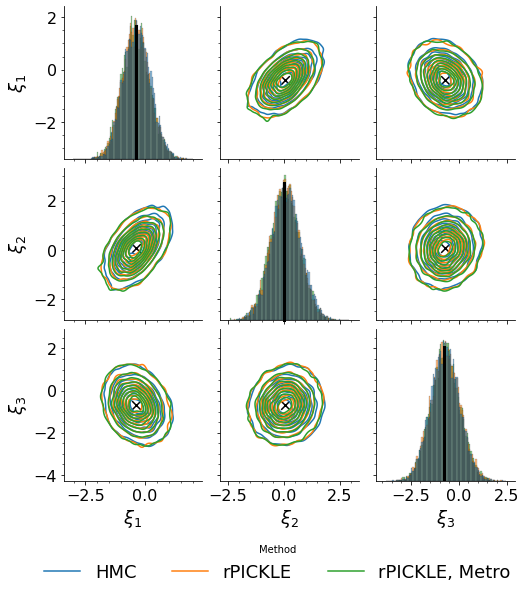}
        \caption{$\sigma_r^2 = 10^{-2}$, $\xi_1$ -  $\xi_3$}
    \end{subfigure}
    \hfill
    \begin{subfigure}{0.45\textwidth}
    \centering
        \includegraphics[width=0.95\textwidth]{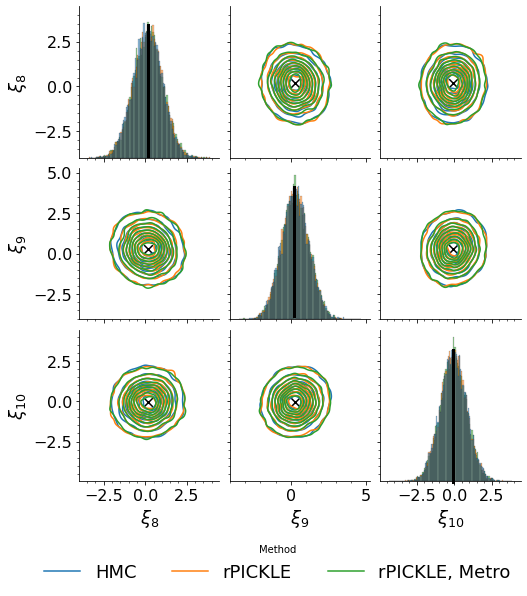}
        \caption{$\sigma_r^2 = 10^{-2}$, $\xi_{8}$ -  $\xi_{10}$}
    \end{subfigure}
    
    \vspace{10pt}

    \centering
    \begin{subfigure}{0.45\textwidth}
    \centering
        \includegraphics[width=0.95\textwidth]{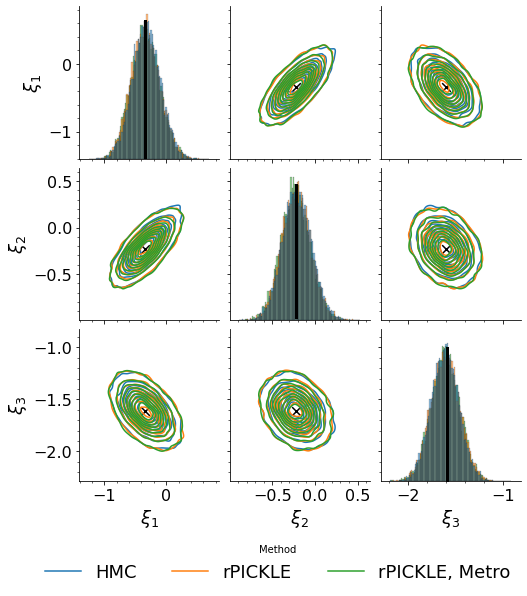}
        \caption{$\sigma_r^2 = 10^{-4}$, $\xi_1$ -  $\xi_3$}
    \end{subfigure}
    \hfill
    \begin{subfigure}{0.45\textwidth}
    \centering
        \includegraphics[width=0.95\textwidth]{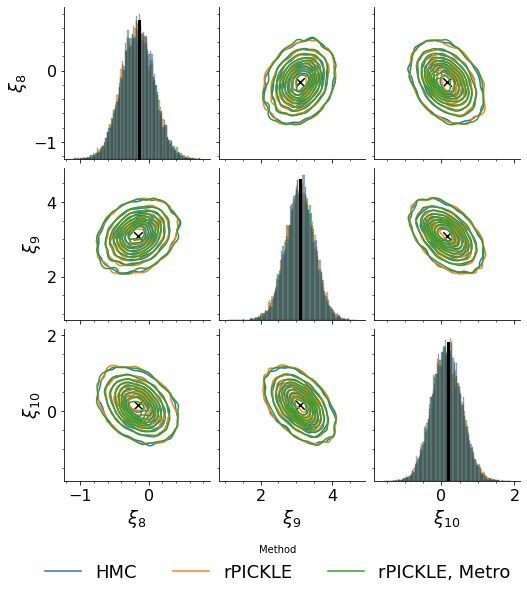}
        \caption{$\sigma_r^2 = 10^{-4}$, $\xi_{8}$ -  $\xi_{10}$}
    \end{subfigure}
    
    \caption{Bivariate joint and marginal distributions of the first (first column) and last (second column) three components of $\boldsymbol\xi$ obtained from HMC, rPICKLE, and rPICKLE with Metropolization for the low-dimensional case. The black cross symbols and lines indicate the coordinates of the mode of the joint posterior distribution computed from  PICKLE. The top and bottom rows show results for $\sigma^2_r = 10^{-2}$ and $10^{-4}$, respectively.}
    \label{fig:Ysmooth_joint_posterior} 
\end{figure}

Figure \ref{fig:Ysmooth_joint_posterior} depicts the marginal and bivariate distributions of the first and last three components of the $\boldsymbol{\xi}$ vector computed from HMC  and rPICKLE with and without Metropolization using the kernel density estimation (KDE) for $\sigma^2_r = 10^{-4}$ and $10^{-2}$. The distributions produced by three different methods have very similar shapes. 
We find that the marginal and bivariate joint distributions are approximately symmetric.
For $\sigma^2_r = 10^{-4}$, the bivariate joint distributions are narrower than for $\sigma^2_r = 10^{-2}$, i.e., stronger physical constraints result in more certain predictions. Also, we see that for the smaller $\sigma^2_r$, the bivariate distributions exhibit stronger correlations between the $\boldsymbol{\xi}$ components. These correlations are much stronger for the first three components of  $\boldsymbol{\xi}$ than for the last three components. 

In Figure  \ref{fig:Ysmooth_joint_posterior}, we also show the coordinates of the joint posterior mode given by the PICKLE solution. We can see that the coordinates of the modes of the marginal and bivariate distributions obtained from HMC and rPICKLE are in good agreement with the coordinates of the joint distribution mode. It should be noted that unless the posterior is Gaussian, the coordinates of the modes of marginal distributions and the corresponding coordinates of the joint distribution mode may not coincide.  Because the coordinates of the marginal and joint distribution modes are close to each other, this indicates that the posterior distribution is close to Gaussian.

\begin{figure}[!htb]

    \centering

    \begin{subfigure}{\textwidth}
        \includegraphics[width=0.24\textwidth]{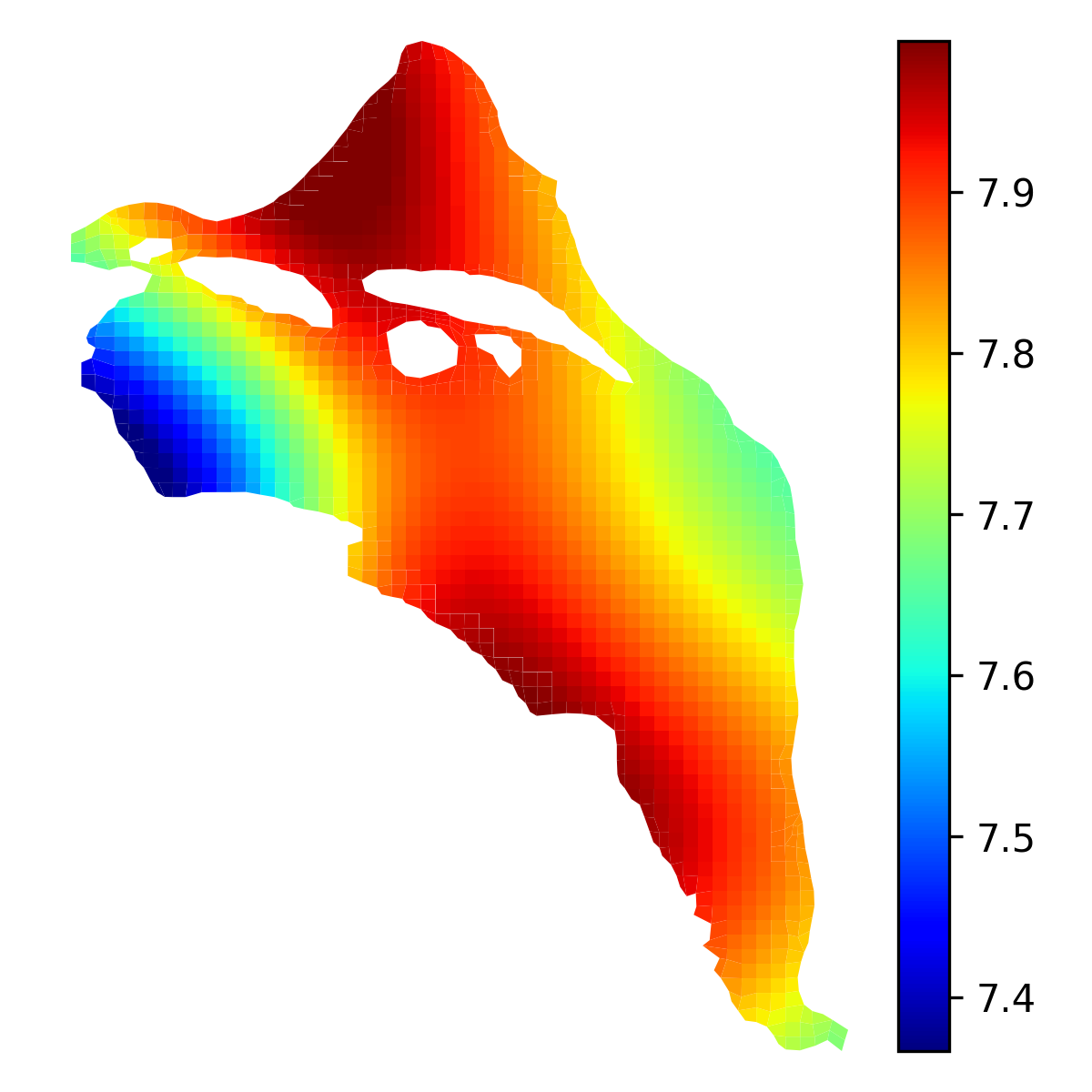}
        \includegraphics[width=0.24\textwidth]{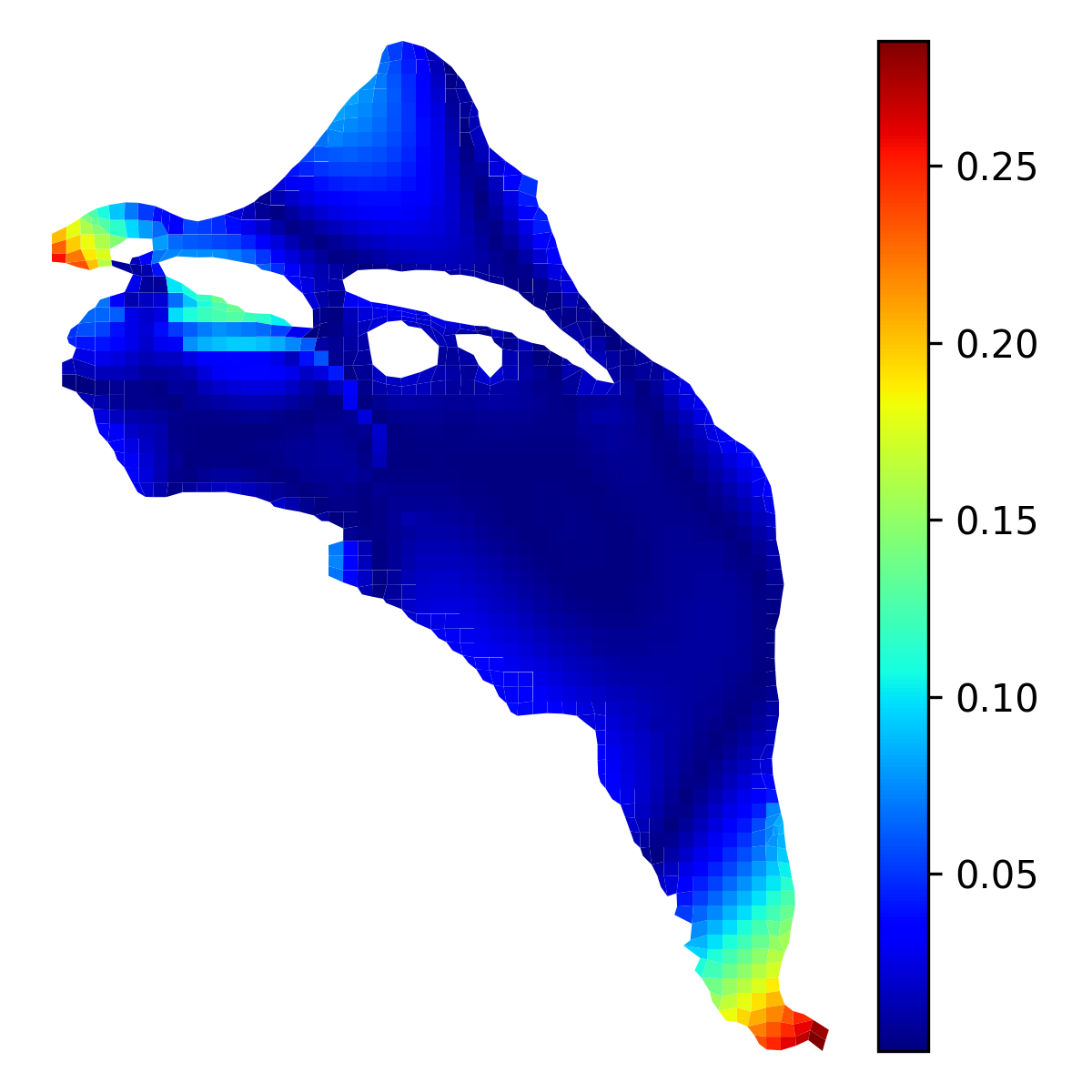}
        \includegraphics[width=0.24\textwidth]{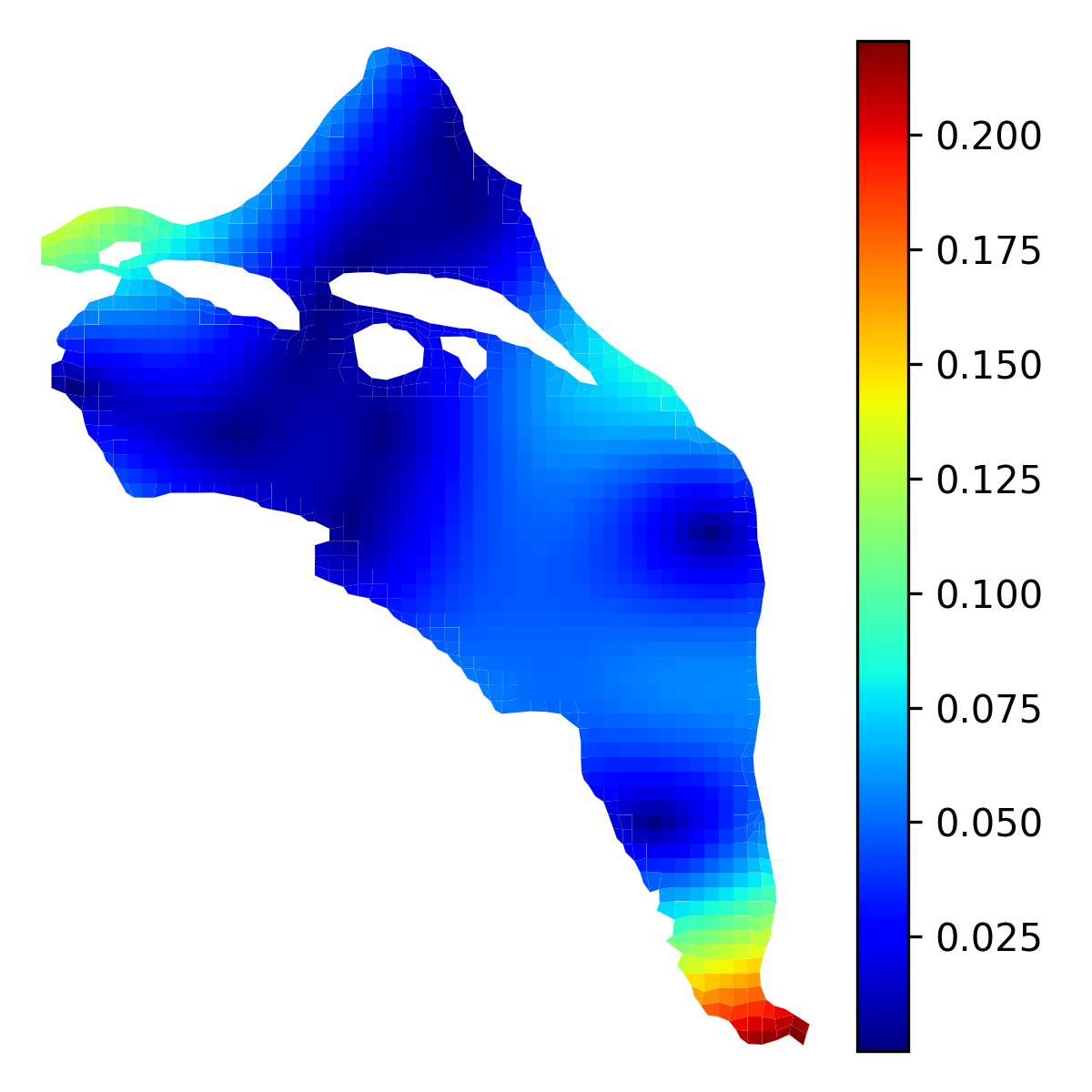}
        \includegraphics[width=0.24\textwidth]{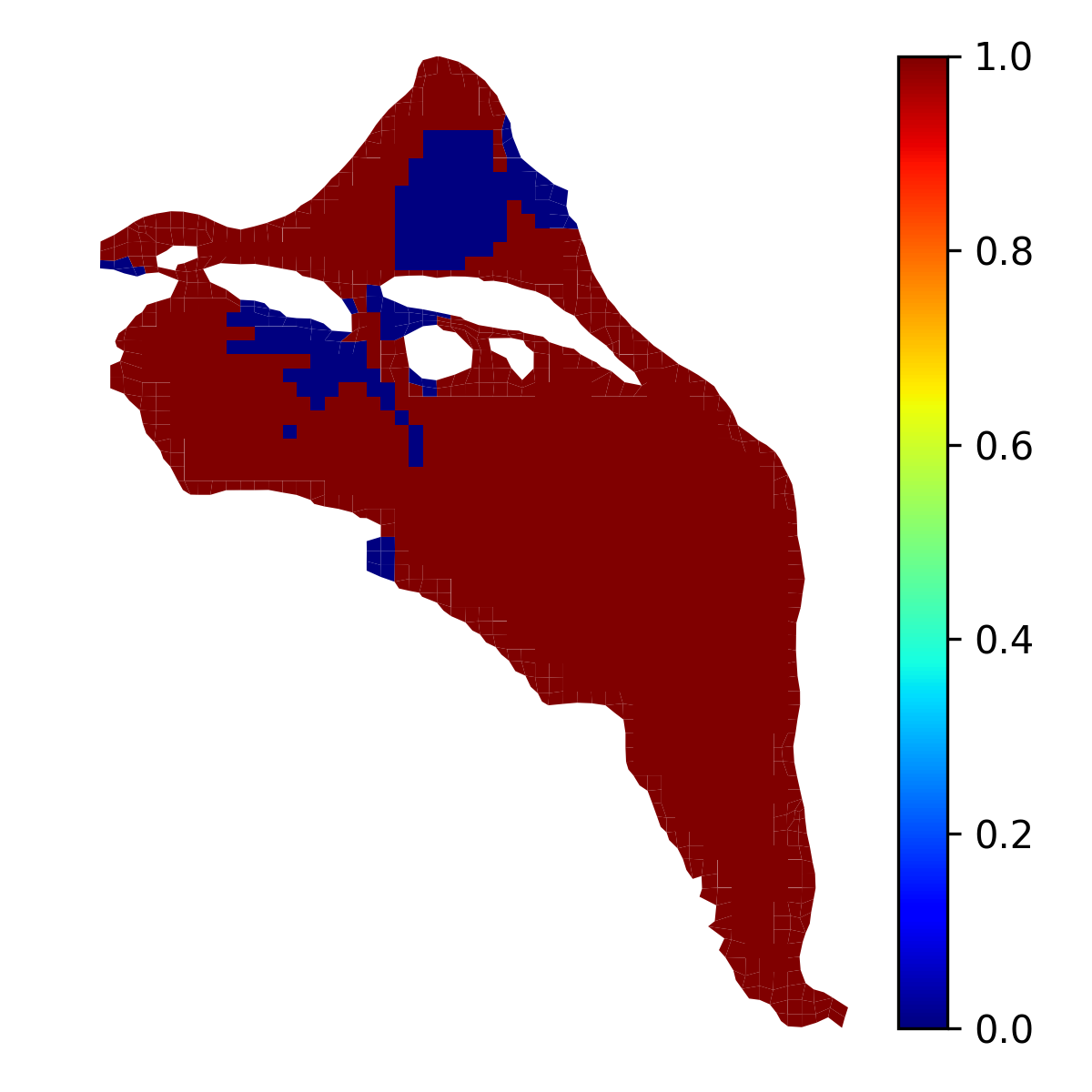}
        \caption{HMC}
    \end{subfigure}
    
    \vspace{10pt}

    \begin{subfigure}{\textwidth}
        \includegraphics[width=0.24\textwidth]{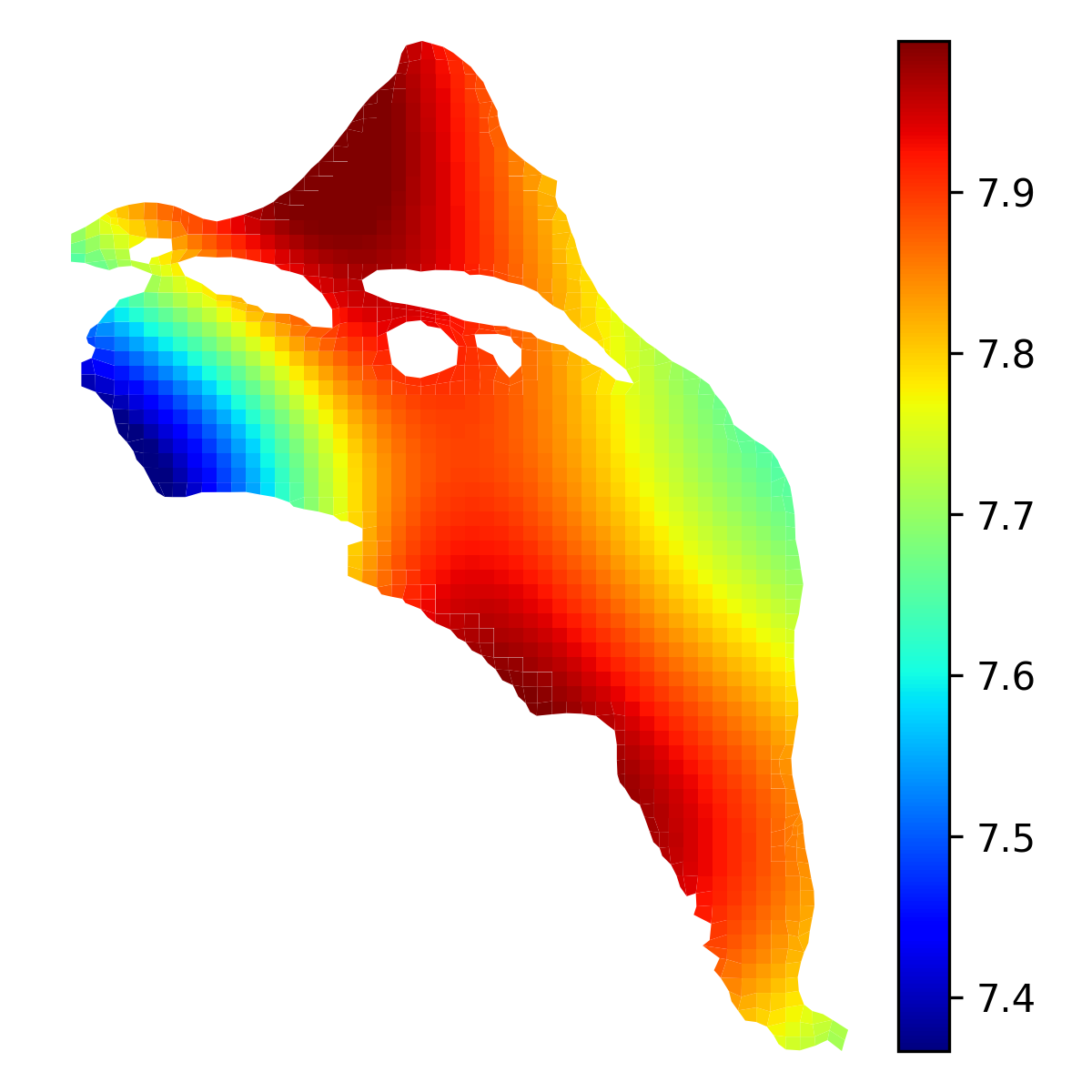}
        \includegraphics[width=0.24\textwidth]{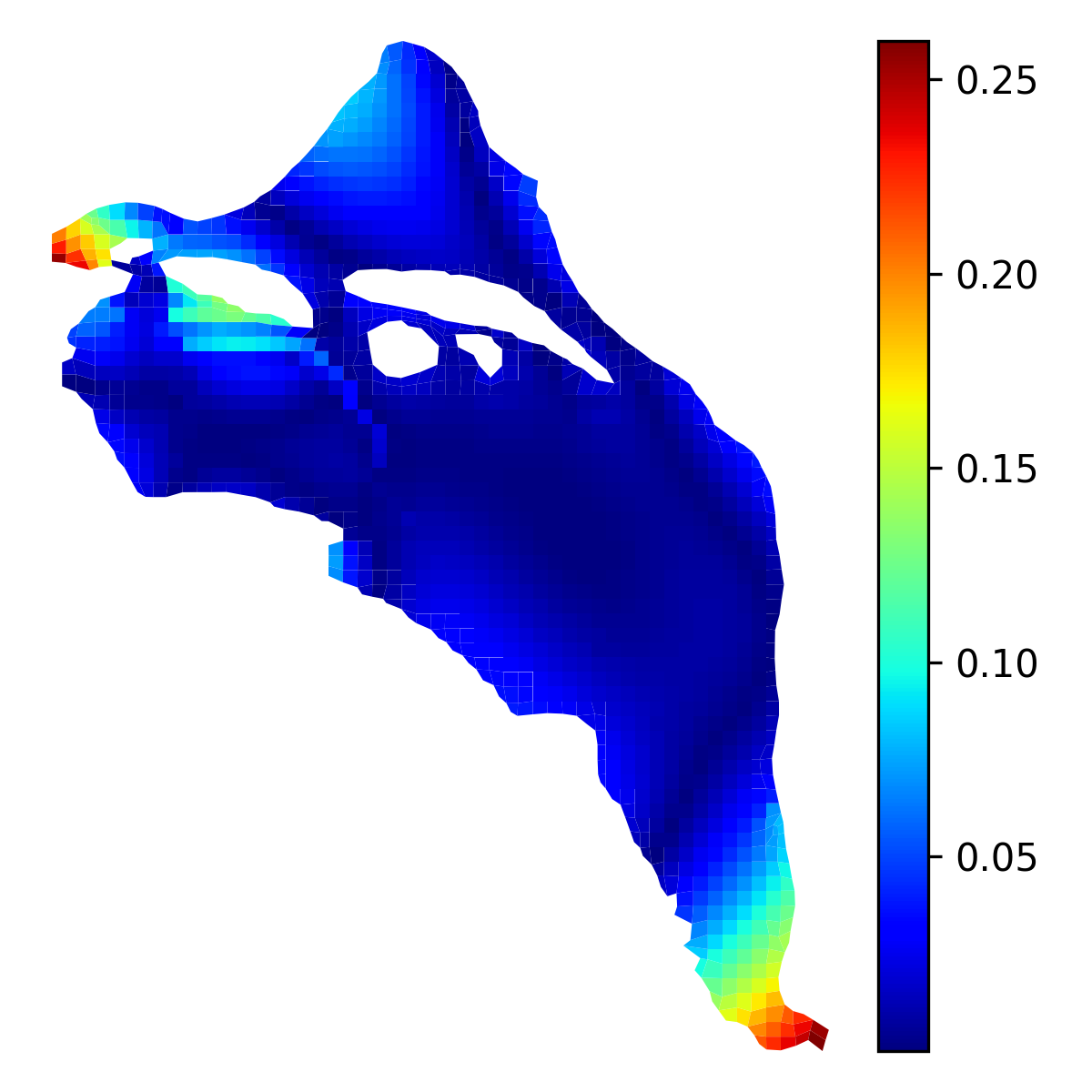}
        \includegraphics[width=0.24\textwidth]{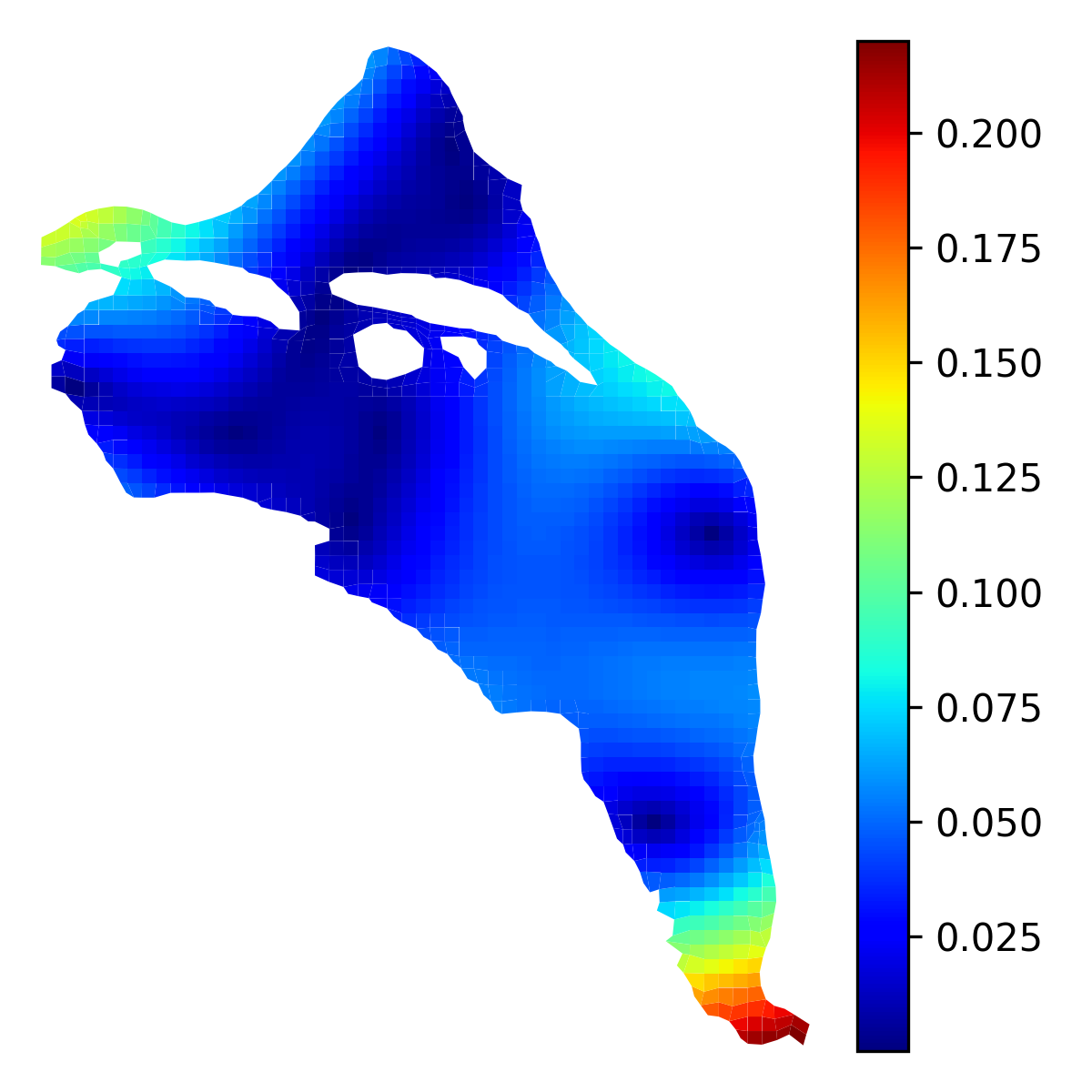}
        \includegraphics[width=0.24\textwidth]{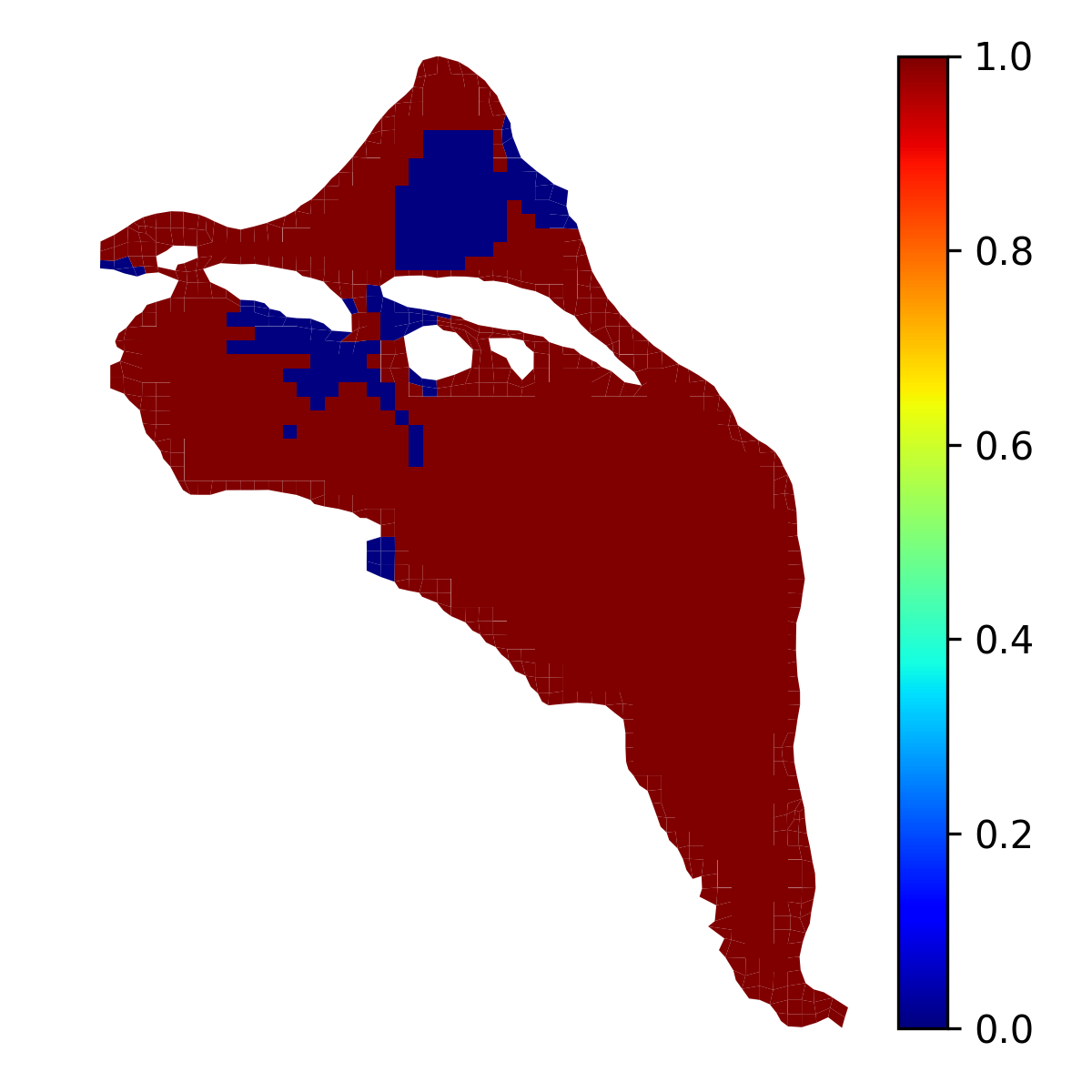}
        \caption{rPICKLE}
    \end{subfigure}

    \vspace{10pt}
    
    \begin{subfigure}{\textwidth}
        \includegraphics[width=0.24\textwidth]{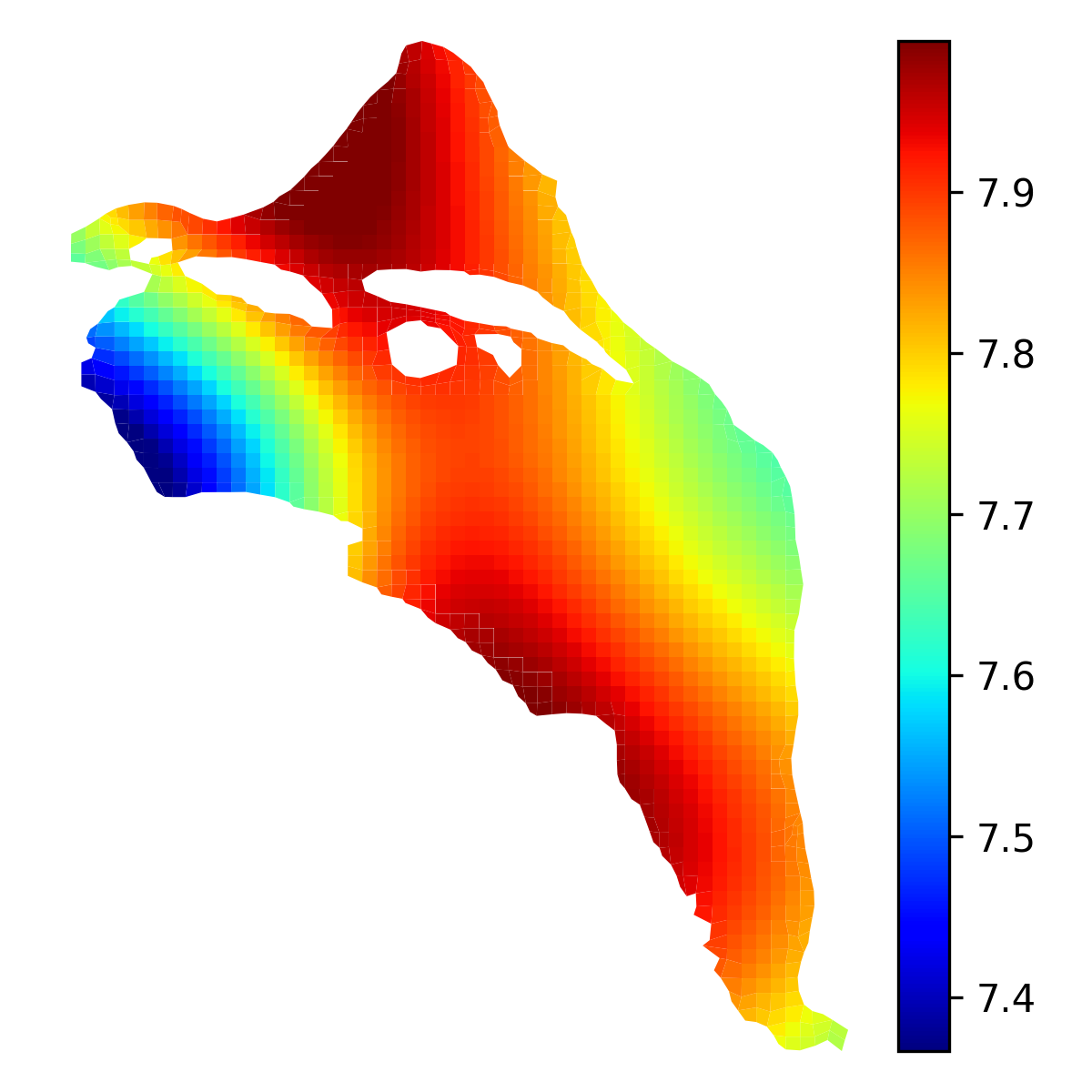}
        \includegraphics[width=0.24\textwidth]{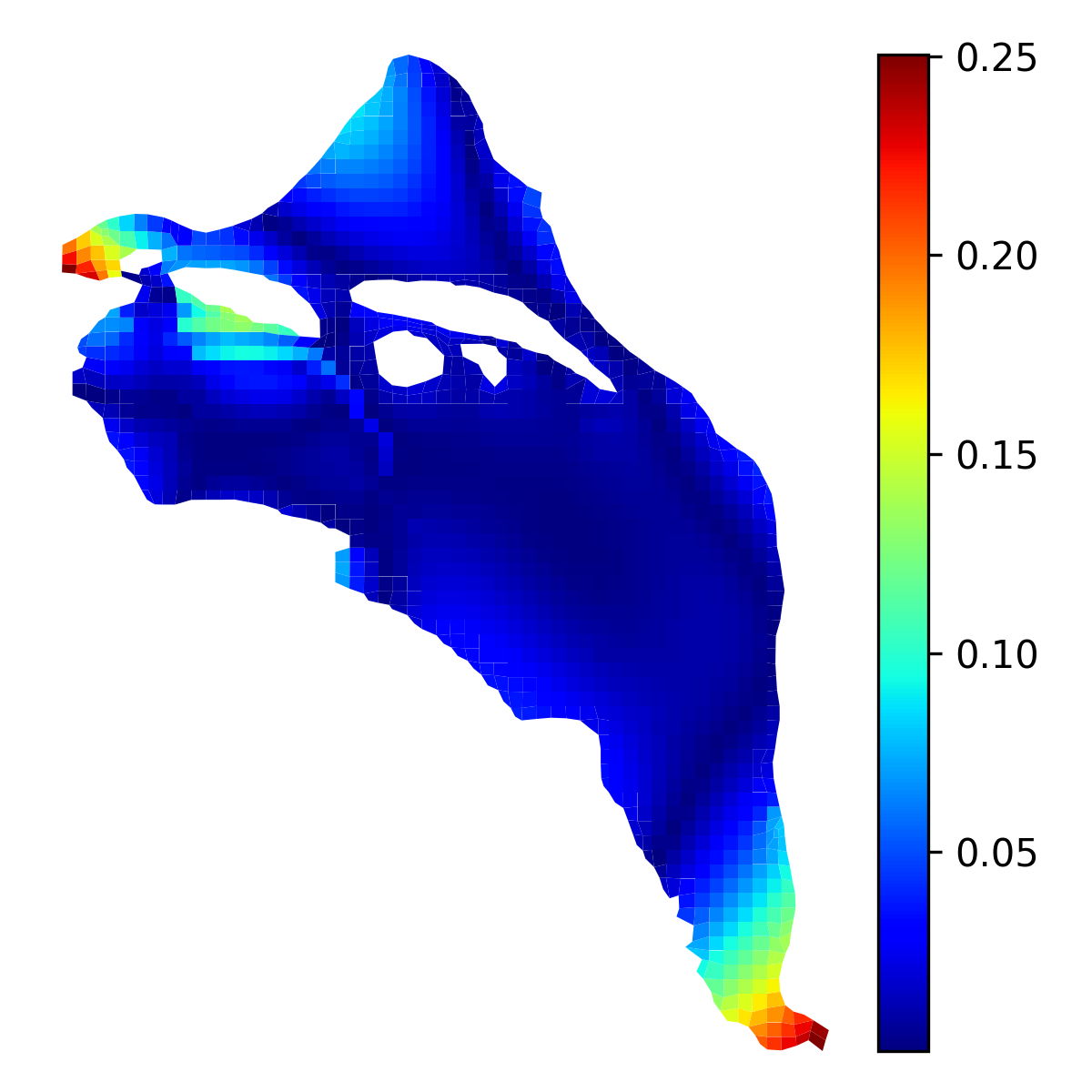}
        \includegraphics[width=0.24\textwidth]{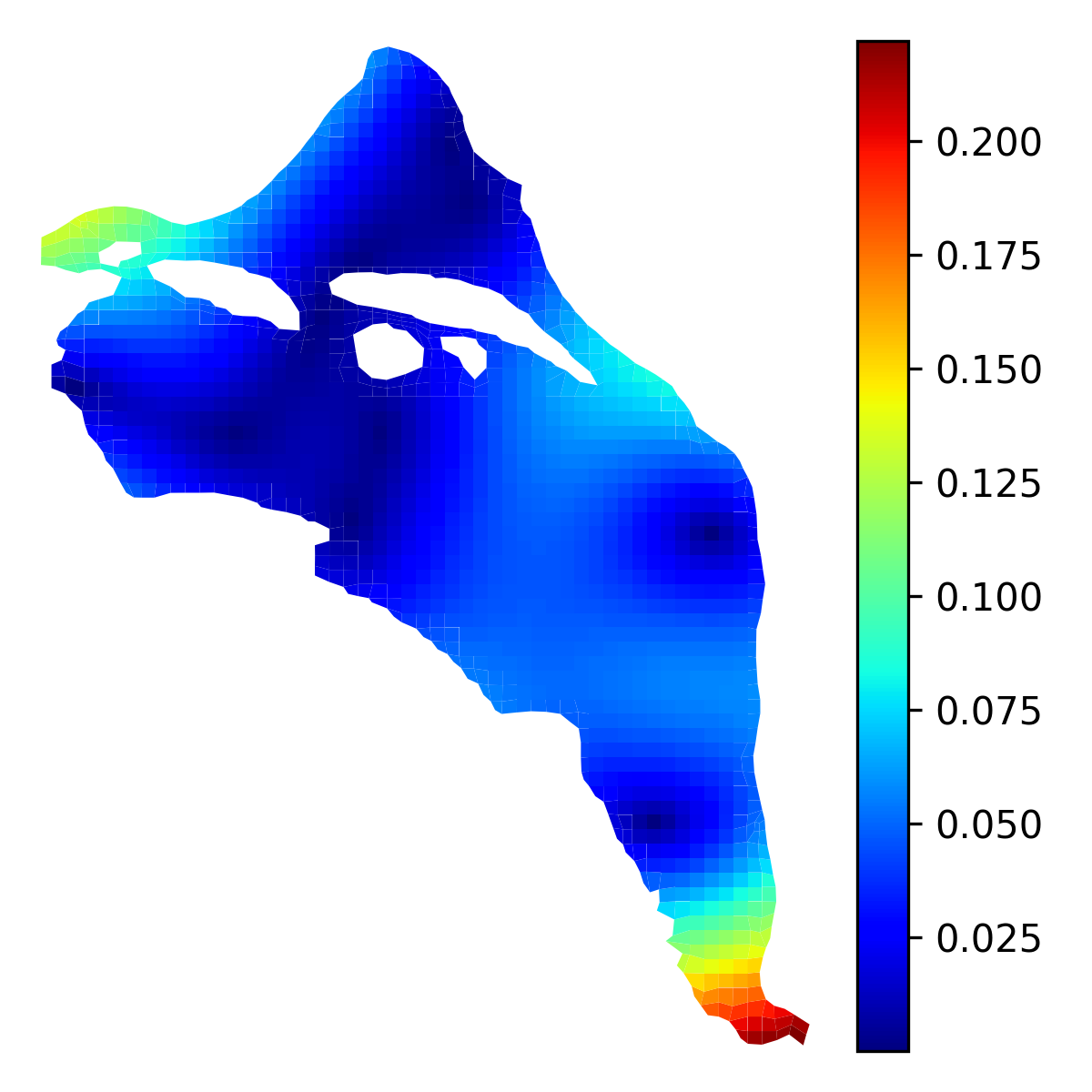}
        \includegraphics[width=0.24\textwidth]{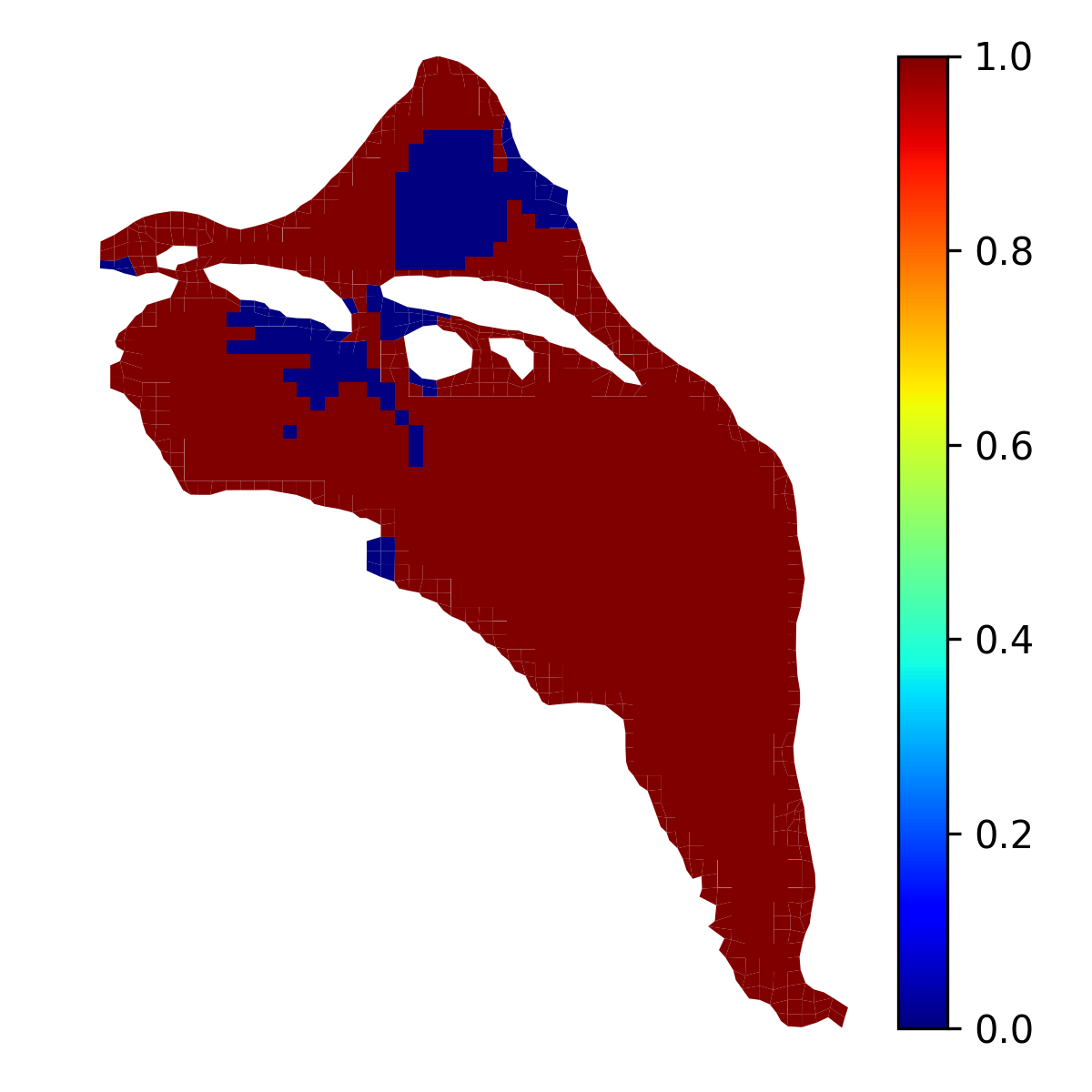}
        \caption{rPICKLE with Metropolization}
    \end{subfigure}
    
    	\caption{The low-dimensional $y$ field estimated from (a) HMC, (b) rPICKLE, and (c) rPICKLE with Metropolization. The first column shows the posterior mean estimates; the second column shows the point errors in the predicted mean with respect to the reference $y_{LD}$ field; the third column shows the posterior standard deviation of  $y$; and the fourth column shows the coverage of the reference field with the $95\%$ credibility interval, where the zero and one values correspond to the reference field being outside and inside the credibility interval, respectively.}
	\label{fig:Ysmooth_results_sigmar_01}
\end{figure}

Finally, we present the Bayesian predictions of $y$. For each of the three methods and $\sigma^2_r = 10^{-2}$, Figure \ref{fig:Ysmooth_results_sigmar_01} shows the estimated mean $\mu_{\hat{y}} (\mathbf{x}| \mathcal{D}_{res}) $ field, the absolute difference between the mean and the reference field $|y - \mu_{\hat{y}} (\mathbf{x}| \mathcal{D}_{res}) |$, the standard deviation $\sigma_{\hat{y}}(\mathbf{x}| \mathcal{D}_{res})$, and the coverage plot showing the locations where the reference solution is within the $95\%$ confidence interval.  In Table \ref{tab:ysmooth}, we summarize results in terms of the relative $\ell_2$ error and $\ell_\infty$ error between the predicted mean and the reference $y_\mathrm{LD}$ fields, LPP, and the coverage (percentage of nodes where the reference solution is within the $95\%$ confidence interval) for 
$\sigma^2_r = 10^{-5}$, $10^{-4}$, $10^{-2}$, $10^{-1}$, and $10^{0}$.  
 For a given value of $\sigma^2_r$, all three methods produce a posterior mean close to the PICKLE's estimate of MAP, and similar LPPs and coverages. This indicates that (i) the HMC and rPICKLE sampled distributions converge to the same posterior (note that in Section \ref{sec:rpickle_linear}, we only prove the consistency of rPICKLE in the linear case), and (ii) Metropolization makes the posterior more descriptive of the reference field, i.e., it yields larger LPP and smaller $\ell_2$ error, but the improvements are less than $1\%$. We also find that,  for this problem, the total acceptance rate for Metropolization is above $95\%$, which agrees with the findings in \cite{oliver1996conditioning, wang2018rmap} that the acceptance rate in randomized algorithms is, in general, very high. 
 It should be noted that the Metropolization requires the estimation of the Jacobian, which becomes computationally expensive for high-dimensional problems. For this reason, in the high-dimensional case presented in Section \ref{sec:high-dimensional}, we do not perform Metropolization and accept all samples generated by the rPICKLE algorithm. 

According to Table \ref{tab:ysmooth}, for a given value of $\sigma^2_r$, HMC and rPICKLE yield similar mean estimates of $y$ in terms of the point errors and the relative $\ell_2$ errors. 
Both, the HMC and rPICKLE errors are smallest for $\sigma^2_r = 10^{-2}$. We find that the LPPs in these methods are also the largest for this value of $\sigma^2_r$. This indicates that $\sigma^2_r = 10^{-2}$ provides the most informative posterior distribution of $y$. 
It is worth reminding that $\sigma^2_r = \gamma$, the regularization parameter in PICKLE. Because PICKLE's $\ell_2$ and $\ell_\infty$ errors are also smallest for $\sigma^2_r = 10^{-2}$, we conclude that this value of $\sigma^2_r$ also provides the optimal regularization for this problem.  

\begin{table}[!ht]
\centering
\caption{The summary of the rPICKLE, HMC, and PICKLE results for the low-dimensional problem with different values of $\sigma^2_r$. Shown are the relative $\ell_2$ and $\ell_{\infty}$ errors in the estimated $y$ field with respect to the reference field, the LPP, and the percentage of coverage of $y_\mathrm{LD}$ by the $95\%$ confidence interval. Note that the PICKLE solution only provides MAP, and LPP and coverage for the PICKLE solution are not given here.}
\label{tab:ysmooth}
\begin{tabular}{p{1cm}|p{2.5cm}p{2cm}p{2cm}p{2cm}p{2cm}}
\toprule
\textbf{$\sigma^2_r$}& \textbf{Method} & \textbf{$\ell_2$ error} & \textbf{$\ell_{\infty}$ error} & \textbf{LPP} & \textbf{Coverage} \\
\midrule
\multirow{4}{*}{\textbf{$10^{0}$}}& PICKLE & $8.05 \times 10^{-3}$ & $3.14 \times 10^{-1}$ & -- & -- \\
        & HMC & $8.20 \times 10^{-3}$ & $3.15 \times 10^{-1}$ & $6972$ & $92 \%$\\
        & rPICKLE & $8.08 \times 10^{-3}$ & $3.14 \times 10^{-1}$ & $6971$  & $92 \%$ \\
        & Metropolized rPICKLE & $7.92 \times 10^{-3}$ & $3.15 \times 10^{-1}$ & $6984$ & $92\%$\\
\hline
\multirow{4}{*}{\textbf{$10^{-1}$}}& PICKLE & $7.28 \times 10^{-3}$ & $3.06 \times 10^{-1}$ & -- & -- \\
        & HMC & $7.55 \times 10^{-3}$ & $3.07 \times 10^{-1}$ & $7206$ & $90 \%$\\
        & rPICKLE & $7.35 \times 10^{-3}$ & $3.08 \times 10^{-1}$ & $7226$  & $90 \%$ \\
        & Metropolized rPICKLE & $7.17 \times 10^{-3}$ & $3.04 \times 10^{-1}$ & $7240$ & $90\%$\\
\hline
\multirow{4}{*}{\textbf{$10^{-2}$}}& PICKLE & $5.90 \times 10^{-3}$ & $2.50 \times 10^{-1}$ & -- & -- \\
        & HMC & $6.33 \times 10^{-3}$ & $2.85 \times 10^{-1}$ & $7561$ & $89 \%$\\
        & rPICKLE & $6.03 \times 10^{-3}$ & $2.60 \times 10^{-1}$ & $7564$ & $89\%$\\
        & Metropolized rPICKLE & $5.91 \times 10^{-3}$ & $2.50 \times 10^{-1}$ & $7546$  & $89 \%$ \\
\hline
\multirow{4}{*}{$10^{-4}$}& PICKLE & $7.14 \times 10^{-3}$ & $3.30 \times 10^{-1}$ & -- & -- \\
        & HMC  & $7.25 \times 10^{-3}$ & $3.40 \times 10^{-1}$ & $-21314$ & $41 \%$ \\
        & rPICKLE   & $7.17 \times 10^{-3}$  & $3.33 \times 10^{-1}$ & $-20725$ & $41 \%$    \\
        & Metropolized rPICKLE  &  $7.17 \times 10^{-3}$& $3.32 \times 10^{-1}$ & $-20232$ & $41 \%$\\
\hline
\multirow{4}{*}{$10^{-5}$}& PICKLE & $9.36 \times 10^{-3}$ & $4.62 \times 10^{-1}$ & -- & -- \\
        & HMC  & $9.36 \times 10^{-3}$ & $4.63 \times 10^{-1}$ & $-275219$ & $16 \%$ \\
        & rPICKLE   & $9.37 \times 10^{-3}$  & $4.62 \times 10^{-1}$ & $-277842$ & $16 \%$     \\
        & Metropolized rPICKLE  &  $9.37 \times 10^{-3}$& $4.62 \times 10^{-1}$ & $-282418$ & $16 \%$\\
\bottomrule
\end{tabular}
\end{table}

Finally, we find that in this low-dimensional problem, the runtime of rPICKLE (i.e., the time to obtain a solution of the rPICKLE minimization problem) is independent of the value of $\sigma^2_r$. For all considered values of $\sigma^2_r$, the runtime per sample was approximately  $0.02$ seconds. On the other hand, the HMC runtime is found to increase with decreasing $\sigma^2_r$ from $1.12$ seconds per sample for $\sigma^2_r = 1$ to $2.01$ for $\sigma^2_r = 10^{-5}$. 

\subsection{High-Dimensional Problem}\label{sec:high-dimensional}

Here, we consider the IUQ problem where the $y$ field is given by $y_{\mathrm{HD}}$. 
First, we estimate the posterior distributions for $\sigma^2_r = 10^{-4}$, $10^{-2}$, and $10^{-1}$ given $N_y^{\text{obs}}=100$ observations of the $y_{\mathrm{HD}}$ field. Later, we estimate posteriors for $N_y^{\text{obs}}=50$ and $200$ to study the dependence of the posterior on the number of $y$ measurements. In all simulations in this section, we assume that $N_u^{\text{obs}}=323$, i.e., $u$ measurements are available at all wells. 

Based on the results in the previous section, we do not perform Metropolis rejection in rPICKLE.  We find that for this high-dimensional problem, the HMC step size, computed from the dual averaging step size adaptation algorithm (which is designed to maintain a prescribed acceptance rate) becomes extremely small. As a result, for some values of $\sigma^2_r $,
 the HMC code fails to reach the stopping criterion ($10^4$ samples) after running for more than 30 days. For comparison, rPICKLE generates the same number of samples in four to five days depending on $\sigma^2_r$. 
 Therefore, for the high-dimensional case, we only present rPICKLE and PICKLE results. We attribute the HMC's large computational time to the high condition number of the posterior covariance matrix, which we find to increase with increasing dimensionality and decreasing $\sigma^2_r$. We investigate this dependence in detail in Section \ref{sec:hmc_failure}. 

\begin{table}[!htp]
\centering
\caption{The summary of the rPICKLE and PICKLE results for the high-dimensional problem for the priors based on different numbers of $y$ observations and $\sigma^2_r$. Shown are the relative $\ell_2$ and $\ell_{\infty}$ errors in the estimated $y$ field with respect to the reference field, the LPP, and the percentage of coverage of $y_\mathrm{LD}$ by the $95\%$ confidence interval.}
\label{tab:yref}
\begin{tabular}{p{1.5cm}|p{2.5cm}p{2cm}p{2cm}p{2cm}p{2cm}}
\toprule
\textbf{$\mathrm{N_y^{obs}}$} & \textbf{Method} & \textbf{$r\ell_2$ error} & \textbf{$\ell_{inf}$ error} & \textbf{LPP} & \textbf{Coverage} \\
\midrule
{} & \multicolumn{5}{c}{$\sigma^2_r = 10^{-1}$}\\
\hline
\multirow{3}{*}{\parbox{1.5cm}{100}}  & PICKLE & $1.08 \times 10^{-1}$ & $4.38 \times 10^{0}$ & -- & --\\
    & rPICKLE & $1.11 \times 10^{-1}$ & $4.41 \times 10^{0}$ & $1730$ & $87 \%$\\
\hline
{} & \multicolumn{5}{c}{$\sigma^2_r = 10^{-2}$} \\
\hline 
\multirow{3}{*}{\parbox{1.5cm}{100}}  & PICKLE & $1.01 \times 10^{-1}$ & $4.14 \times 10^{0}$ & -- & -- \\
    & rPICKLE & $1.04 \times 10^{-1}$ & $4.13 \times 10^{0}$ & $2032$ & $82 \%$\\

\hline
{} & \multicolumn{5}{c}{$\sigma^2_r = 10^{-4}$} \\
\hline
\multirow{2}{*}{\parbox{1.5cm}{50}}  & PICKLE & $1.91 \times 10^{-1}$ & $6.49 \times 10^{0}$ & -- & --\\
    & rPICKLE   & $1.71 \times 10^{-1}$  & $6.50 \times 10^{0}$ & $-2792$ &  65\%     \\
\hline
\multirow{2}{*}{\parbox{1.5cm}{100}}  & PICKLE & $1.26 \times 10^{-1}$ & $6.14 \times 10^{0}$ & -- & --\\
     & rPICKLE   & $1.10 \times 10^{-1}$  & $5.33 \times 10^{0}$ & 1070 & 67\%      \\
\hline
\multirow{2}{*}{\parbox{1.5cm}{200}}  & PICKLE & $7.91 \times 10^{-2}$ & $5.34 \times 10^{0}$ & -- & -- \\
     & rPICKLE   & $7.99 \times 10^{-2}$  & $5.38 \times 10^{0}$ & 5220 & 75\%\\
\bottomrule
\end{tabular}
\end{table}

Table \ref{tab:yref} summarizes the relative $\ell_2$ and $\ell_{\infty}$ errors, LPP, and coverage in rPICKLE estimates of $y$ for $\sigma^2_r = 10^{-4}$,  $10^{-2}$, and  $10^{-1}$. We find that the smallest rPICKLE errors and the largest LPP are achieved for  $\sigma^2_r = 10^{-2}$. However, we find that LPP is more sensitive to  $\sigma^2_r$ than to $\ell_2$ error--$\ell_2$ errors vary by less than 7\% for the considered $\sigma^2_r$  values, while LPP values change by more than 100\%. Also, the $\ell_2$ error is 1\% smaller for $\sigma^2_r = 10^{-4}$ than for  $\sigma^2_r = 10^{-1}$ but the LPP  is 70\% larger for  $\sigma^2_r = 10^{-1}$ than for $\sigma^2_r = 10^{-4}$. Therefore, we conclude that LPP provides a better criterion for selecting $\sigma^2_r$ than $\ell_2$ errors. Errors in the PICKLE and rPIKCLE predictions of $y$ are very similar, and $\sigma^2_r = 10^{-2}$ also provides the best value of the regularization coefficient for PICKLE, i.e., the PICKLE error is smallest for  $\gamma=\sigma^2_r = 10^{-2}$.

\begin{figure}[!htb]
    \centering
    \begin{subfigure}{0.45\textwidth}
    \centering
        \includegraphics[width=0.99\textwidth]{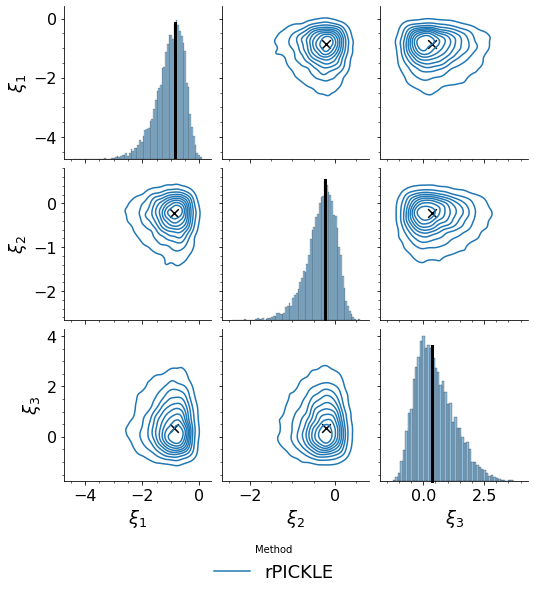}
        \caption{$\sigma_r^2 = 10^{-1}$, $\xi_1$ -  $\xi_3$}
    \end{subfigure}
    \hfill
    \begin{subfigure}{0.45\textwidth}
    \centering
        \includegraphics[width=0.99\textwidth]{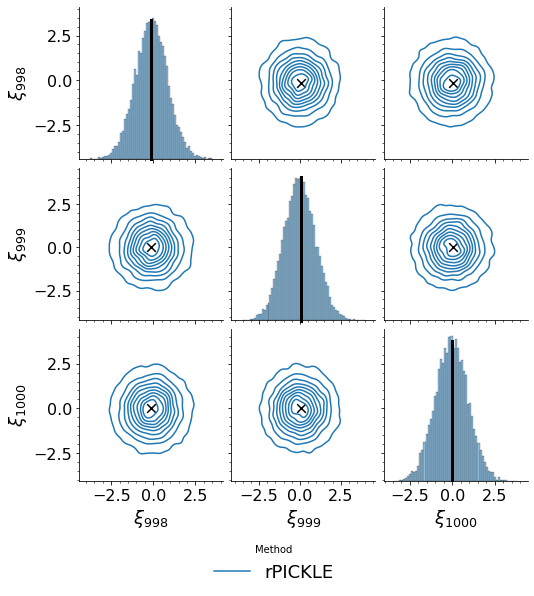}
        \caption{$\sigma_r^2 = 10^{-1}$, $\xi_{998}$ -  $\xi_{1000}$}
    \end{subfigure}
    
    \vspace{10pt}

    \centering
    \begin{subfigure}{0.45\textwidth}
    \centering
        \includegraphics[width=0.99\textwidth]{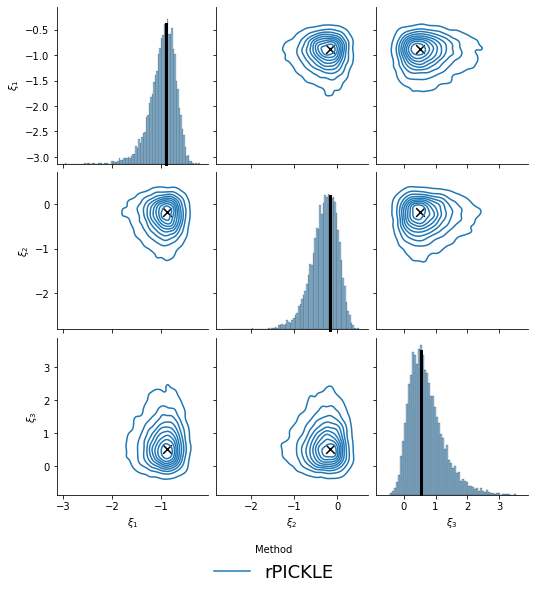}
        \caption{$\sigma_r^2 = 10^{-2}$, $\xi_1$ -  $\xi_3$}
    \end{subfigure}
    \hfill
    \begin{subfigure}{0.45\textwidth}
    \centering
        \includegraphics[width=0.99\textwidth]{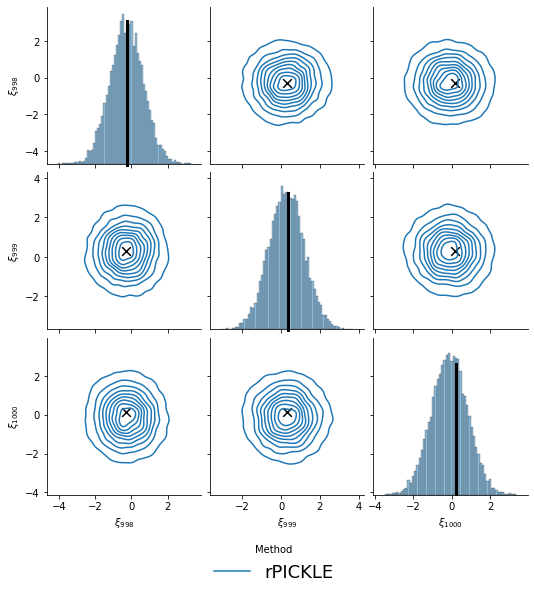}
        \caption{$\sigma_r^2 = 10^{-2}$, $\xi_{998}$ -  $\xi_{1000}$}
    \end{subfigure}
    
    \caption{Bivariate joint and marginal distributions of the first (first column) and last (second column) three components of $\boldsymbol\xi$ obtained from rPICKLE for the high-dimensional case. The black cross symbols and lines indicate the coordinates of the mode of the joint posterior distribution computed from  PICKLE. The top and bottom rows show results for $\sigma^2_r = 10^{-1}$ and $10^{-2}$, respectively.}
    \label{fig:Yref_joint_posterior} 
\end{figure}

Figure \ref{fig:Yref_joint_posterior} depicts the marginal and bivariate distributions of the first and last three components of $\boldsymbol\xi$ for $\sigma^2_r=10^{-1}$ and $10^{-2}$. Compared with the low-dimensional case, we observe that the posterior of the first three components is more non-symmetric and correlated. The posterior distributions become narrower as $\sigma^2_r$ becomes smaller.  The modes of these distributions have non-zero coordinates (the prior distributions are centered at zero).  On the other hand, the last three terms have symmetric marginal distributions approximately centered at zero and circular-shaped bi-variate distributions, the latter indicating the lack of cross-correlation.

Figure \ref{fig:Yref_joint_posterior} also shows the coordinates of the joint posterior distribution mode obtained from PICKLE. The coordinates of the joint distribution slightly deviate from the coordinates of the marginal and bivariate distributions because of the non-Gaussianity of the posterior.

\begin{figure}[!htb]

    \begin{subfigure}{\textwidth}
        \includegraphics[width=0.24\textwidth]{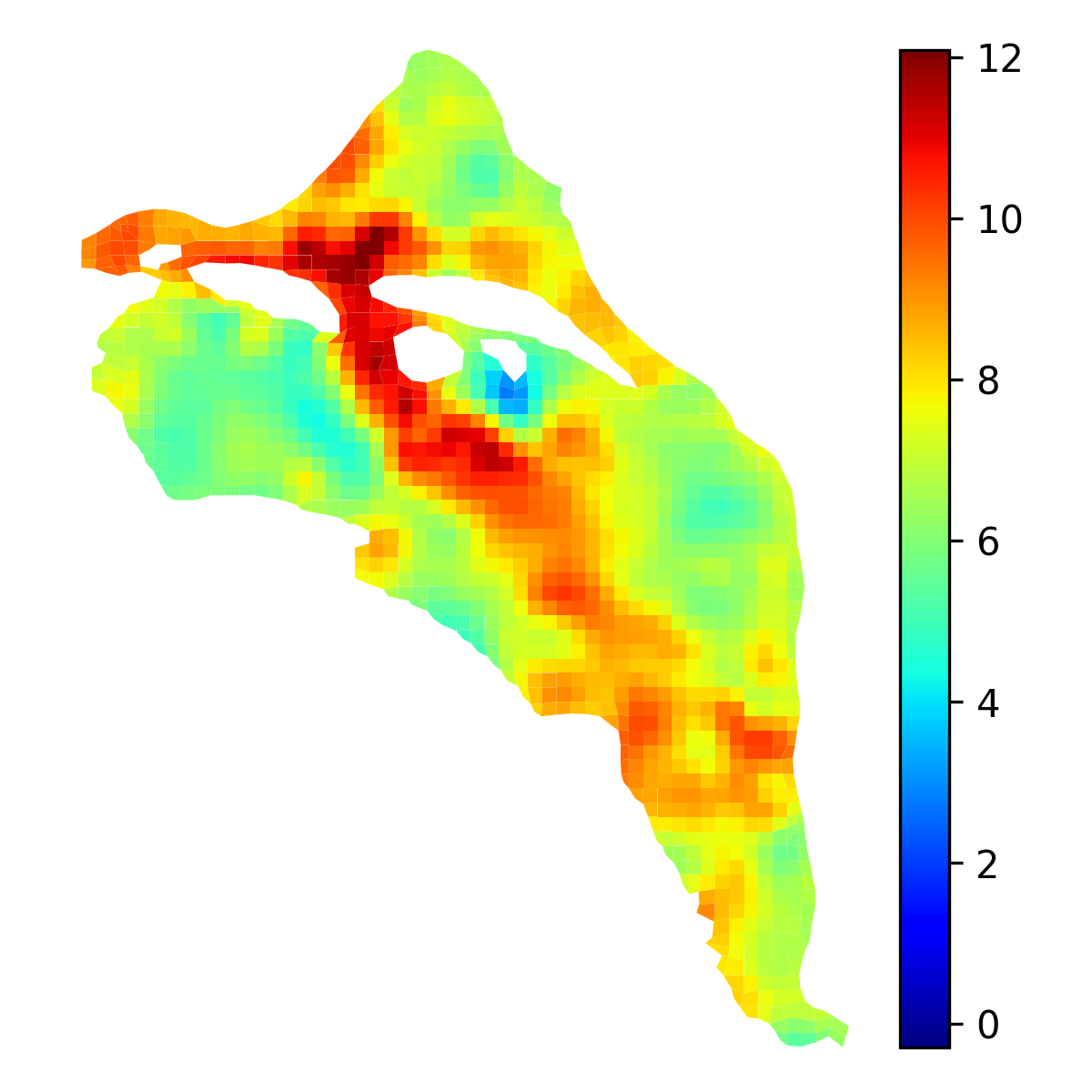}
        \includegraphics[width=0.24\textwidth]{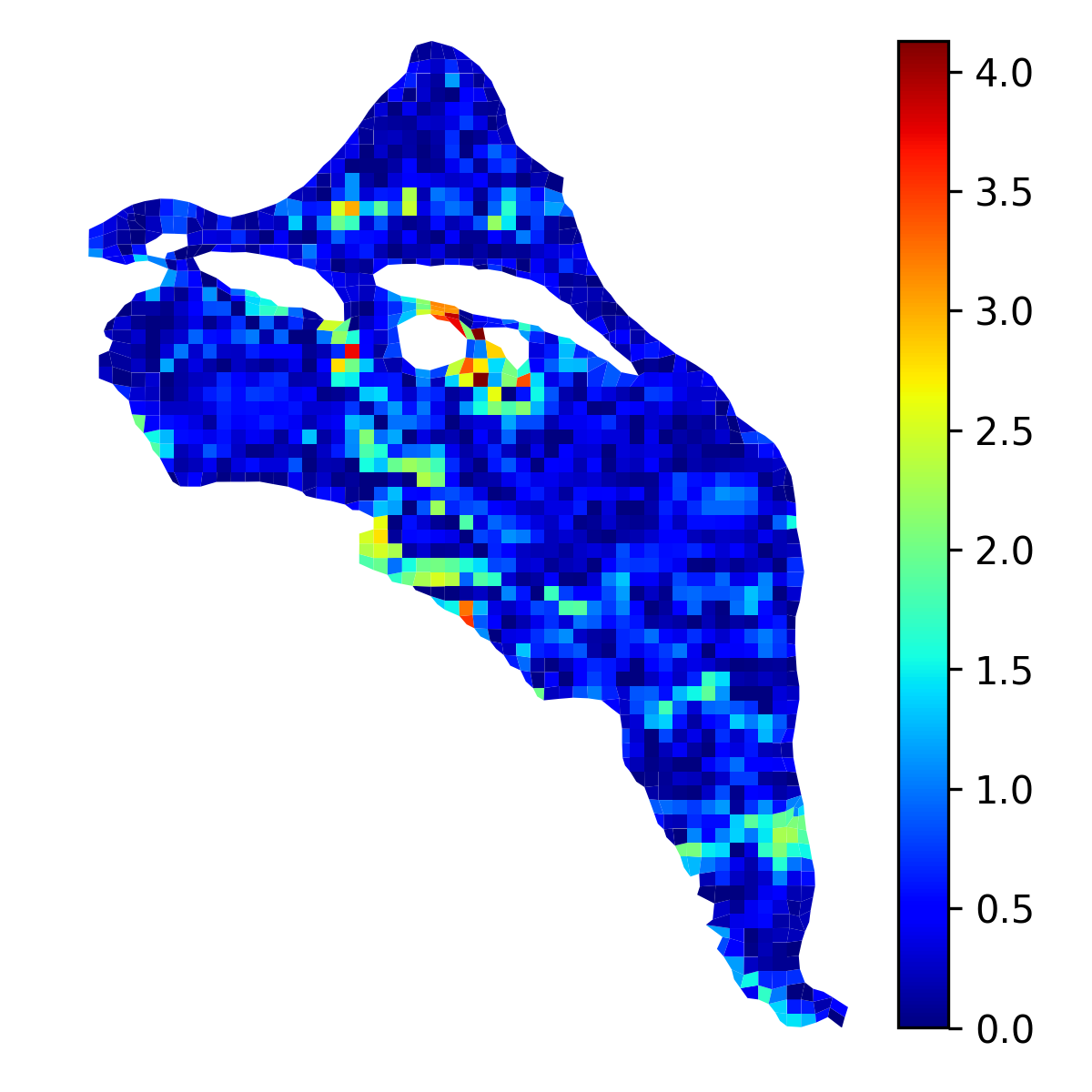}
        \includegraphics[width=0.24\textwidth]{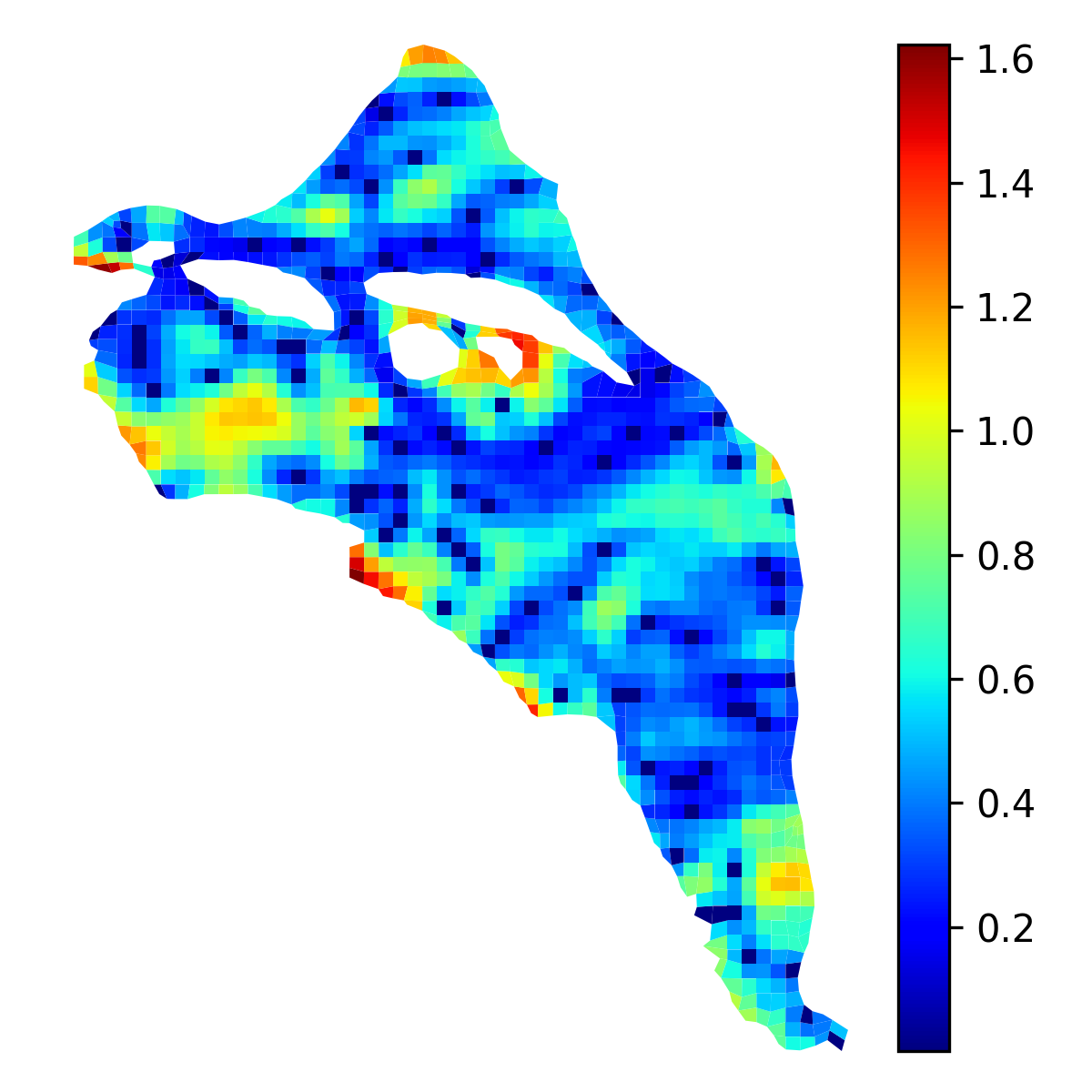}
        \includegraphics[width=0.24\textwidth]{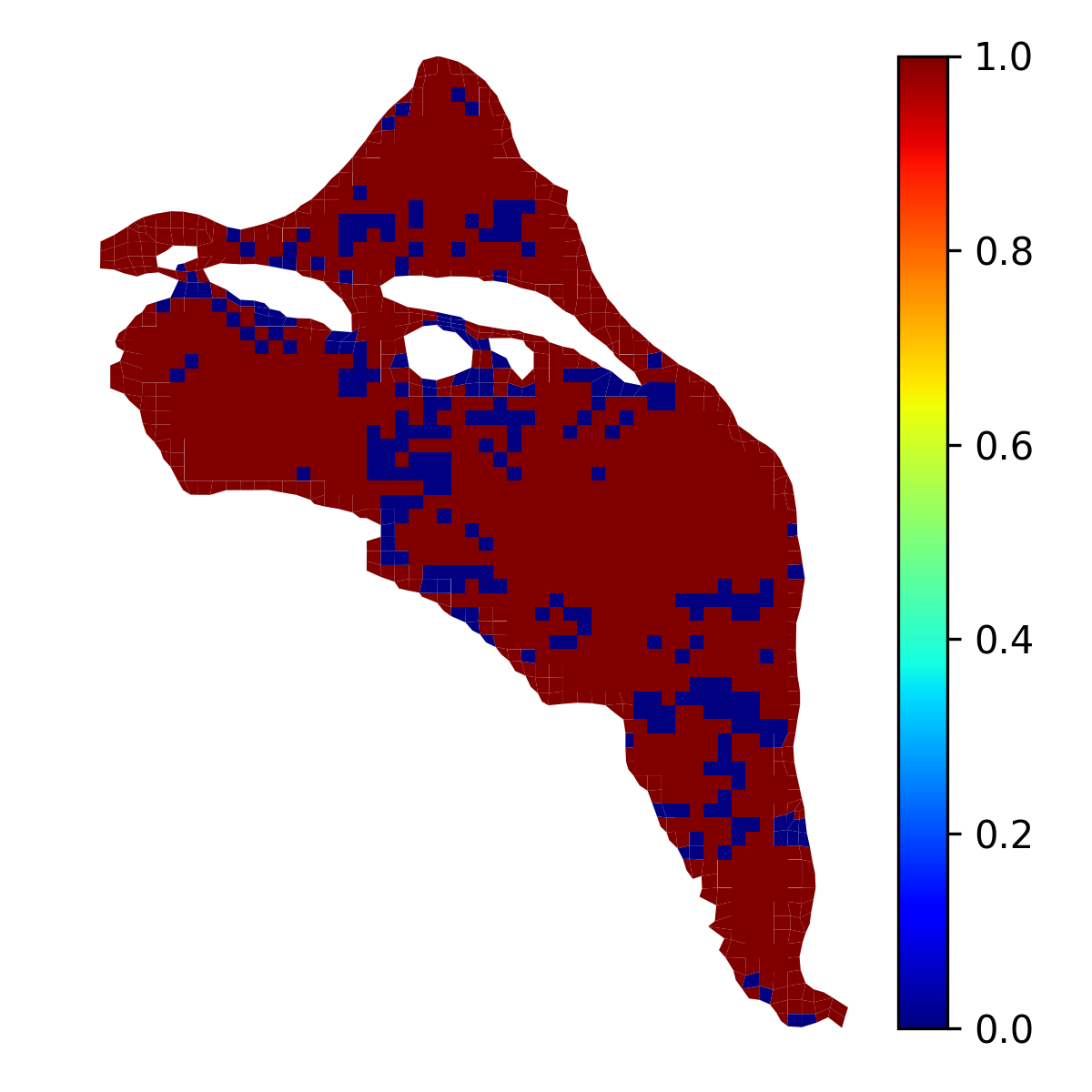}
        \caption{ $\sigma^2_{r} = 10^{-2}$}
    \end{subfigure}
    
        \vspace{10pt}
    
    \begin{subfigure}{\textwidth}
        \includegraphics[width=0.24\textwidth]{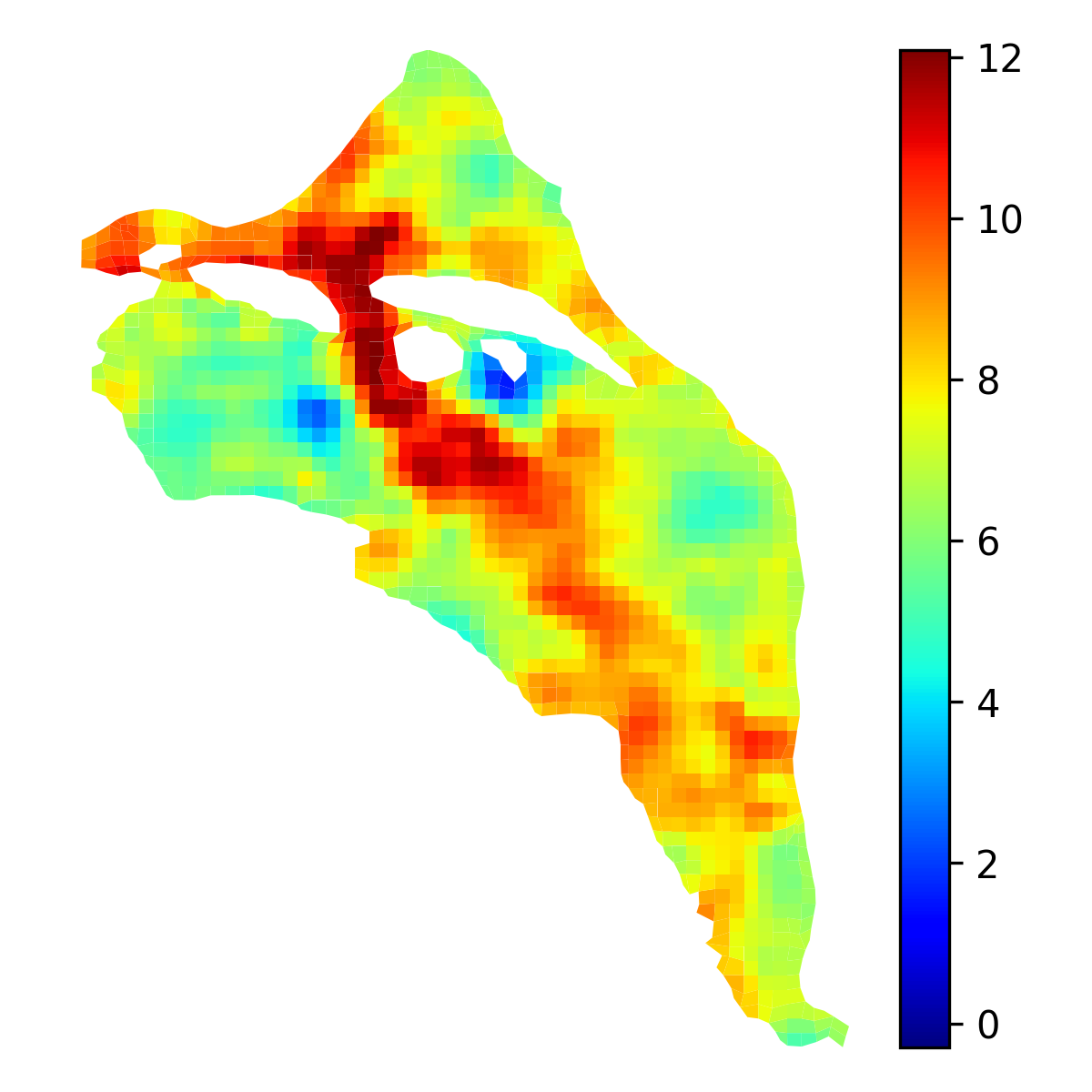}
        \includegraphics[width=0.24\textwidth]{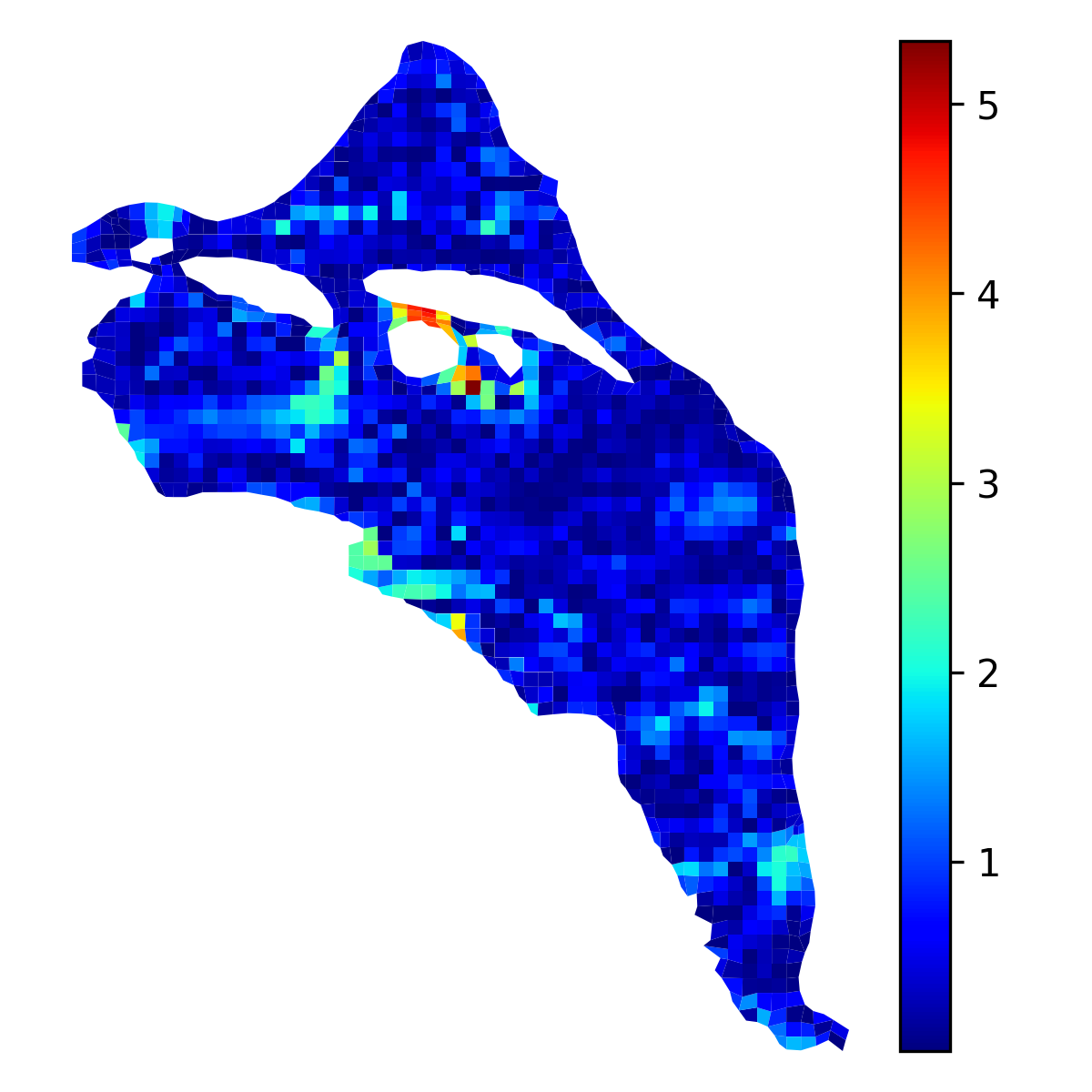}
        \includegraphics[width=0.24\textwidth]{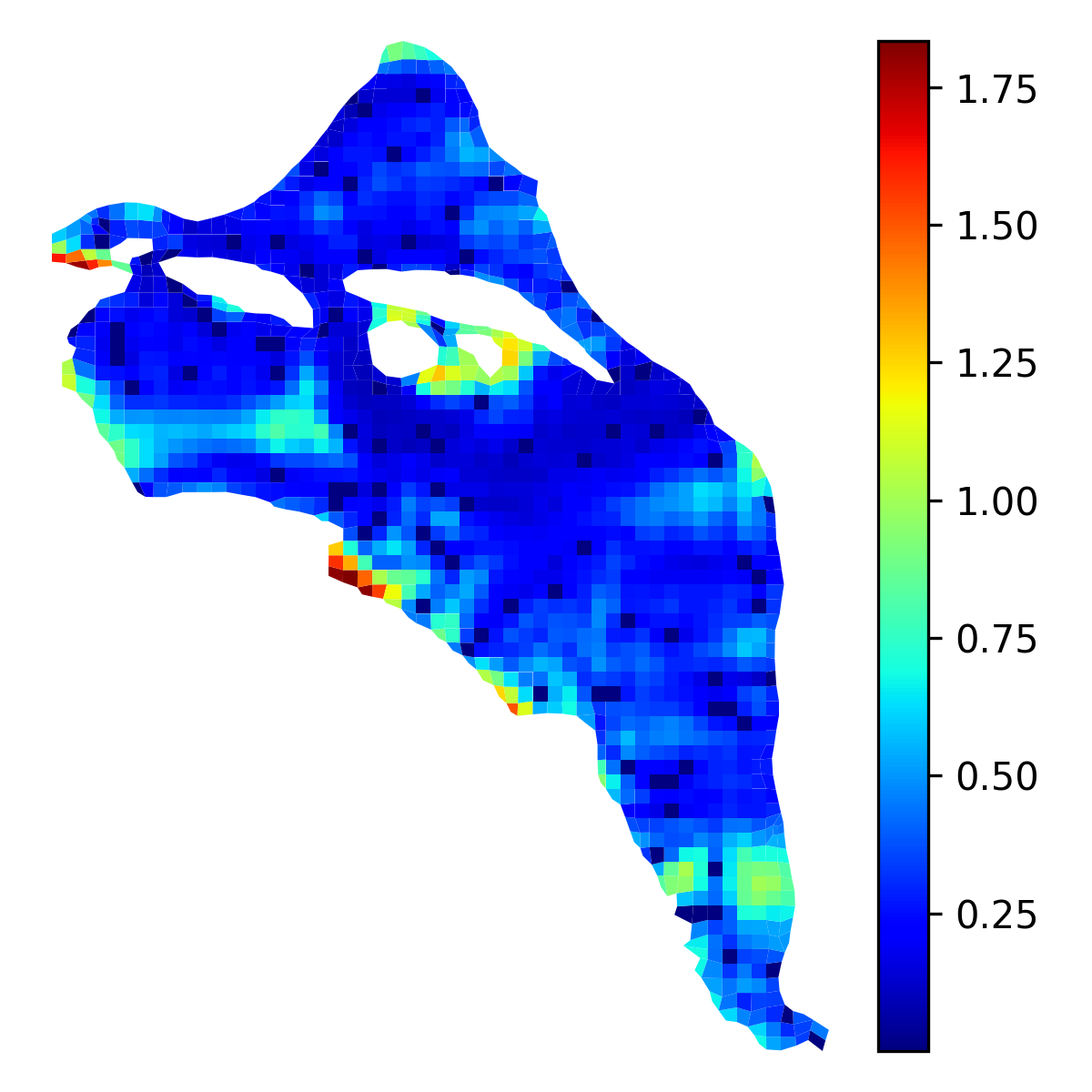}
        \includegraphics[width=0.24\textwidth]{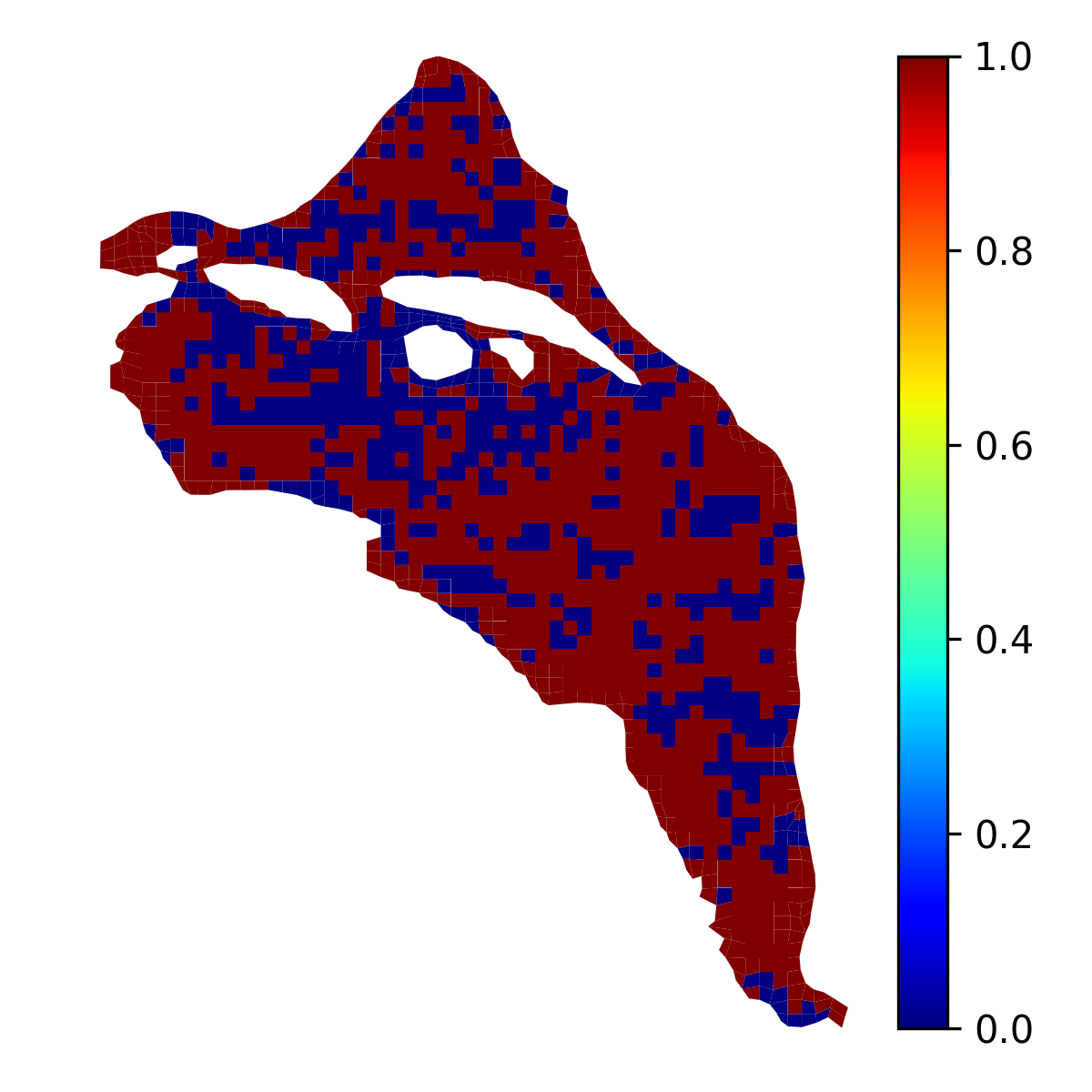}
        \caption{ $\sigma^2_{r} = 10^{-4}$}
    \end{subfigure}
    
    	\caption{rPICKLE estimates of $y_{\mathrm{HD}}$ with $\sigma^2_{r} = 10^{-2}$  and $\sigma^2_{r} = 10^{-4}$ and 100 measurements of  $y_{\mathrm{HD}}$. (First column) the posterior mean of $y$,  (second column) point errors computed as the difference between the posterior mean of $y$ and the reference $y$ field, (third column) the posterior standard deviation of $y$, and (fourth column) the coverage of the reference $y$ by the $95\%$ credibility interval.}
	\label{fig:Yref_results_sigmar_01}
\end{figure}

Figure \ref{fig:Yref_results_sigmar_01} shows the rPICKLE estimate of the posterior mean of $y$, the absolute point difference between the mean and reference $y_{\mathrm{HD}}(x)$ fields, the posterior standard deviation of $y(x)$, and the coverage for $\sigma^2_r = 10^{-2}$ and $\sigma^2_r = 10^{-4}$. We see significant differences in rPICKLE predictions for different $\sigma^2_r$. Errors in the predictions with $\sigma^2_r = 10^{-4}$  are in general larger than in the predictions with $\sigma^2_r = 10^{-2}$ with the maximum point error being 50\% larger. As expected, the posterior standard deviations are generally larger in the prediction with the larger $\sigma^2_r$. However, the maximum point standard deviation is larger in the simulation with the smaller $\sigma^2_r$.   
We also see that $\sigma^2_r = 10^{-2}$ produces a better coverage--the reference $y_{\mathrm{HD}}$ field is within the confidence interval in 82\% of all predicted locations versus 65\% for  $\sigma^2_r = 10^{-2}$. 
We reiterate that the LPP for $\sigma^2_r = 10^{-4}$ is -2792, which is significantly smaller than that for $\sigma^2_r = 10^{-2}$ (2032).

\begin{figure}[!htb]
    \begin{subfigure}{\textwidth}
        \includegraphics[width=0.24\textwidth]{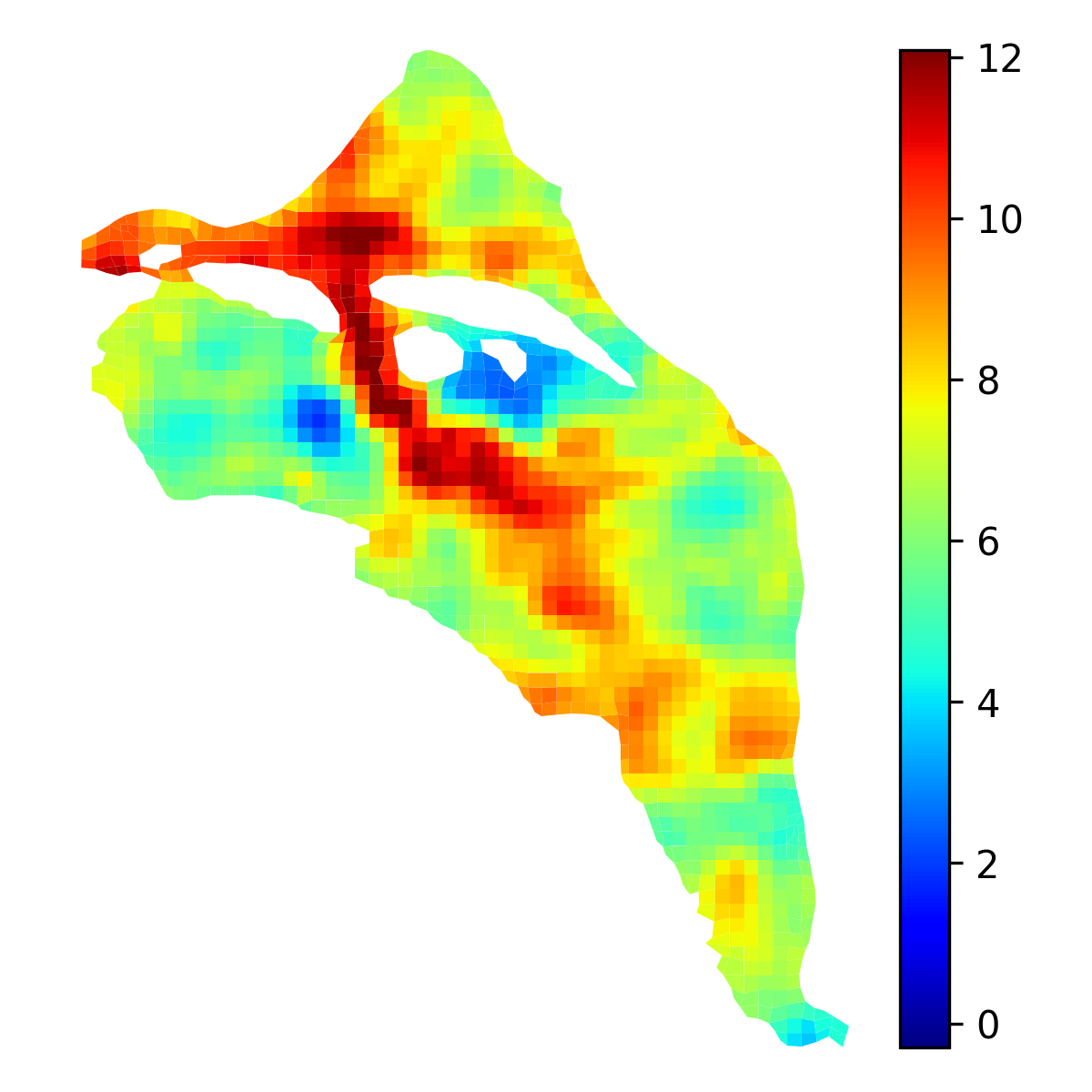}
        \includegraphics[width=0.24\textwidth]{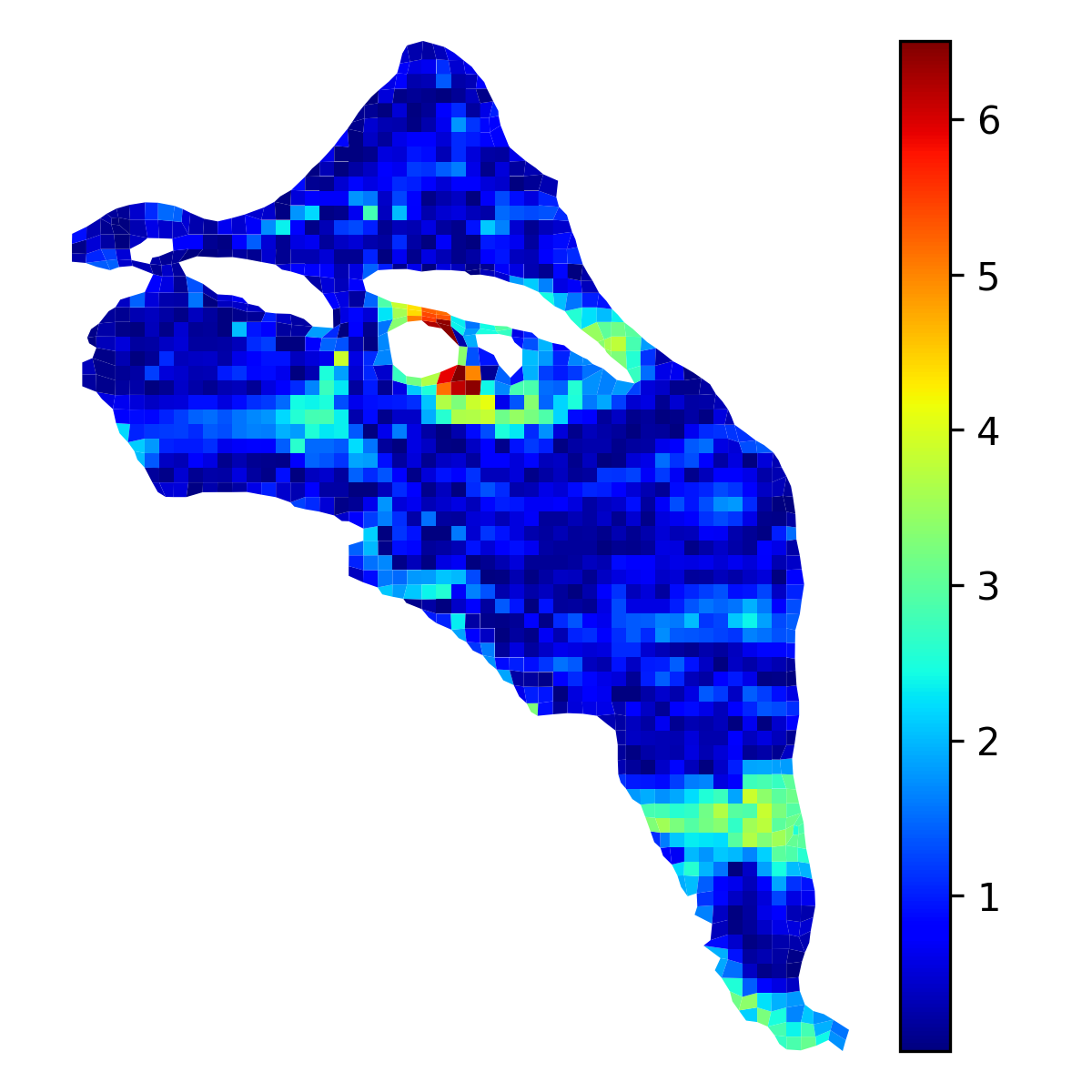}
        \includegraphics[width=0.24\textwidth]{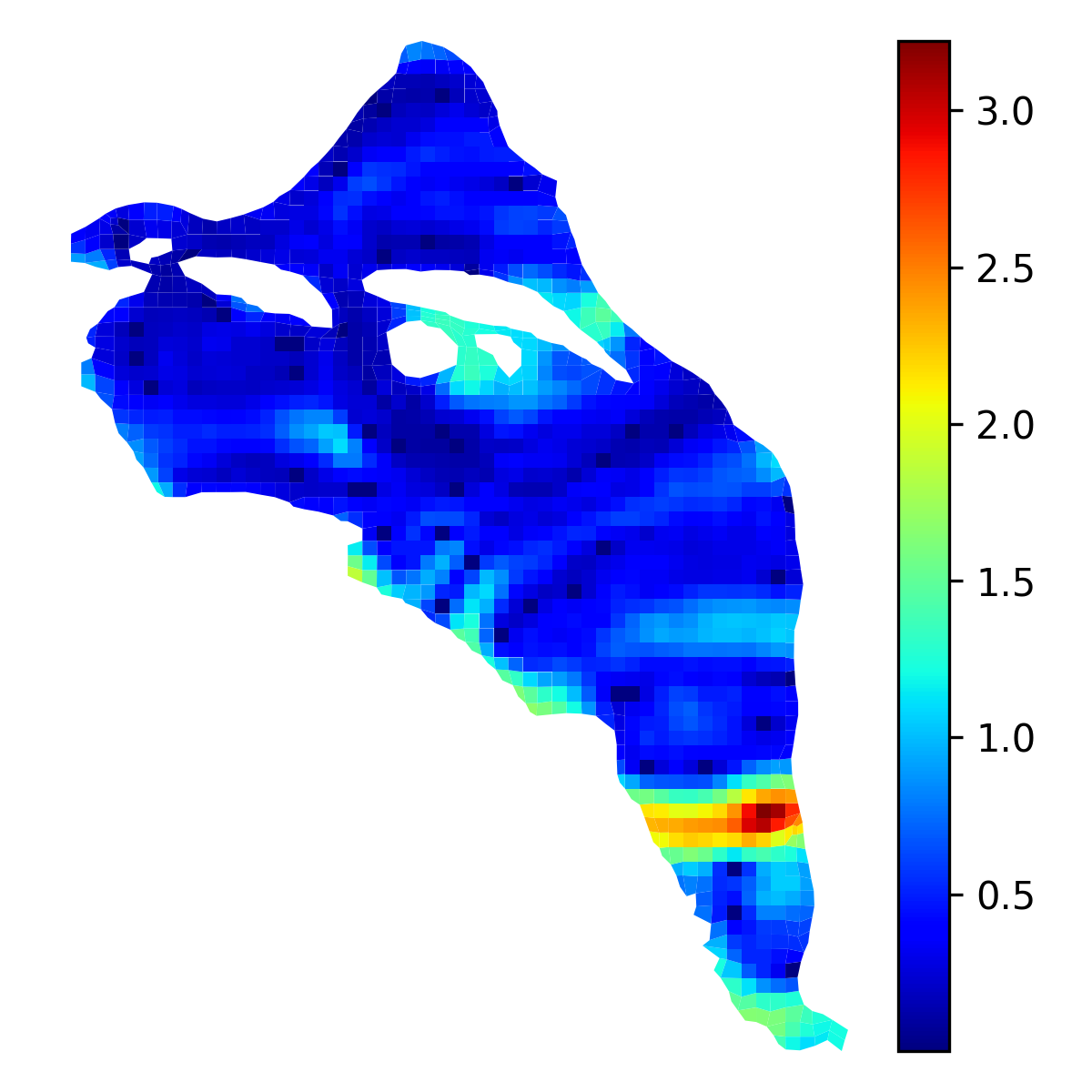}
        \includegraphics[width=0.24\textwidth]{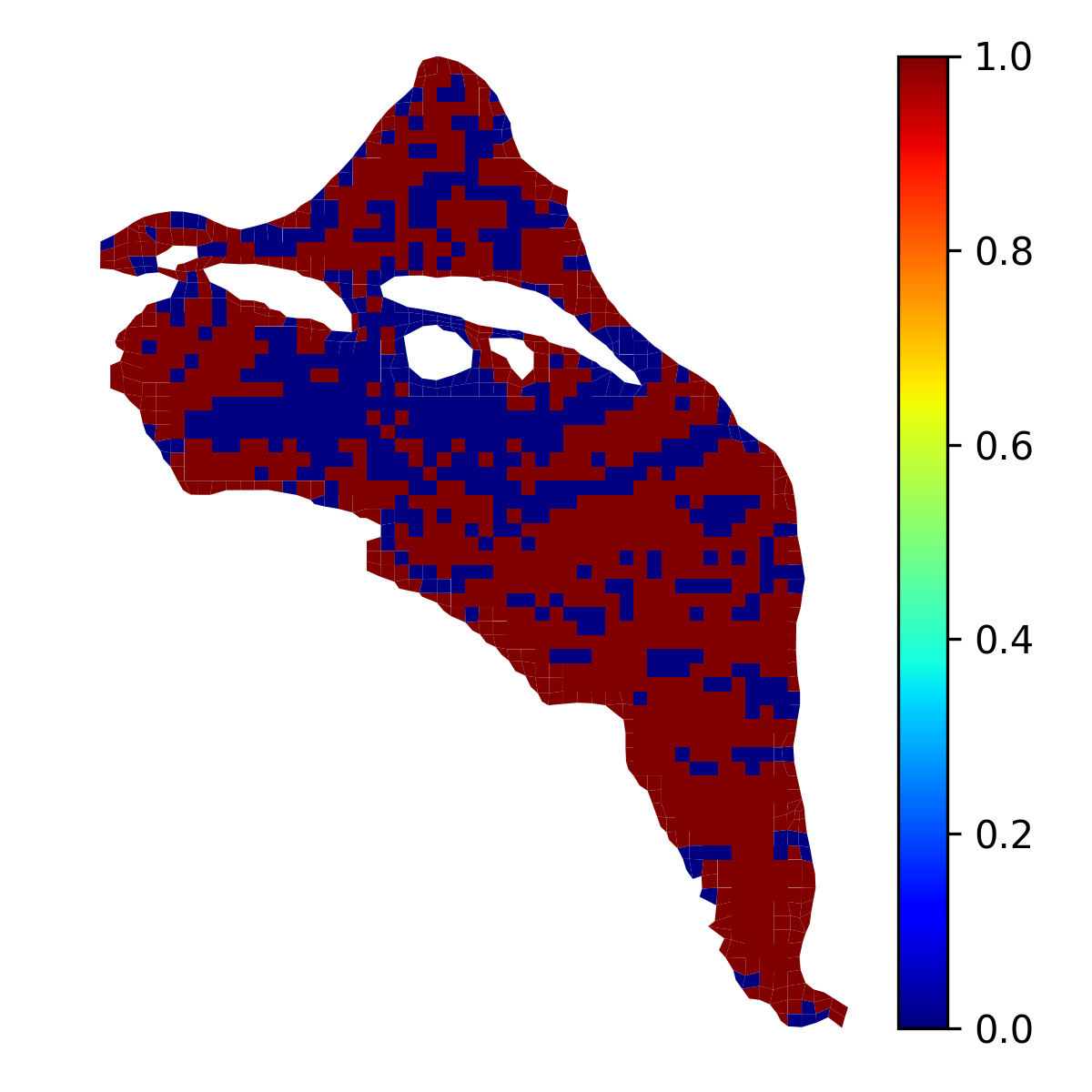}
        \caption{$\mathrm{N_y^{\text{obs}} = 50}$}
    \end{subfigure}

    \vspace{10pt}
    
    \begin{subfigure}{\textwidth}
        \includegraphics[width=0.24\textwidth]{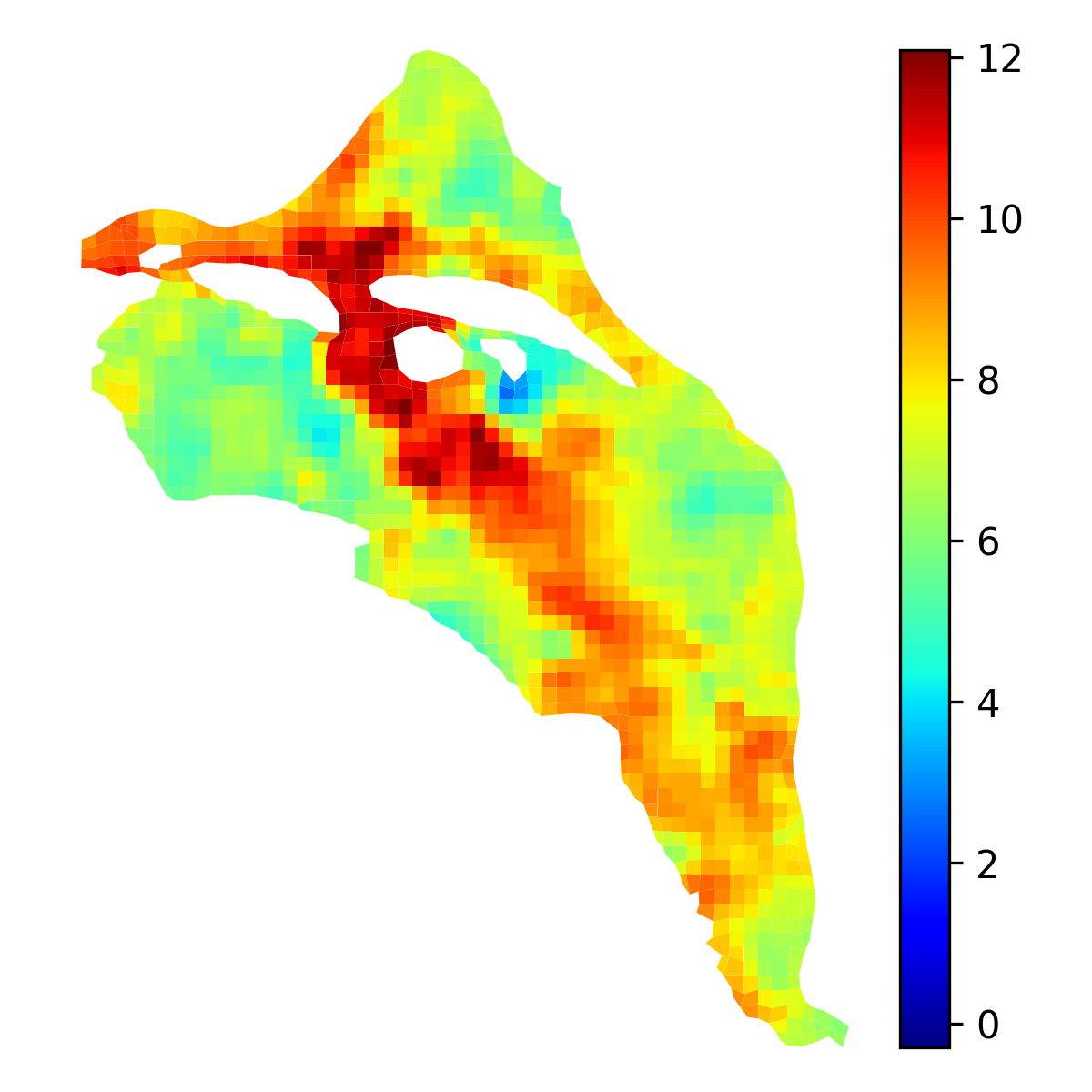}
        \includegraphics[width=0.24\textwidth]{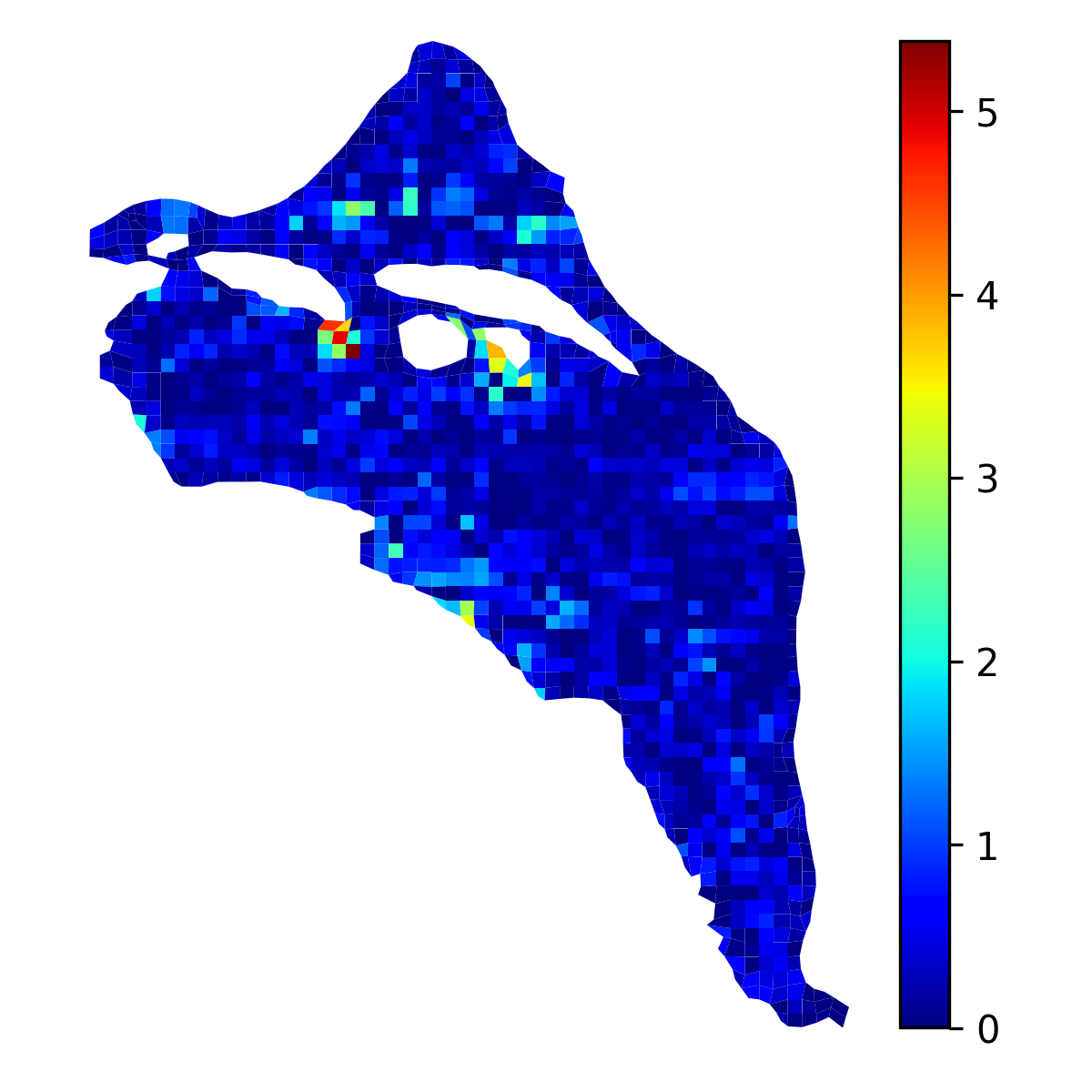}
        \includegraphics[width=0.24\textwidth]{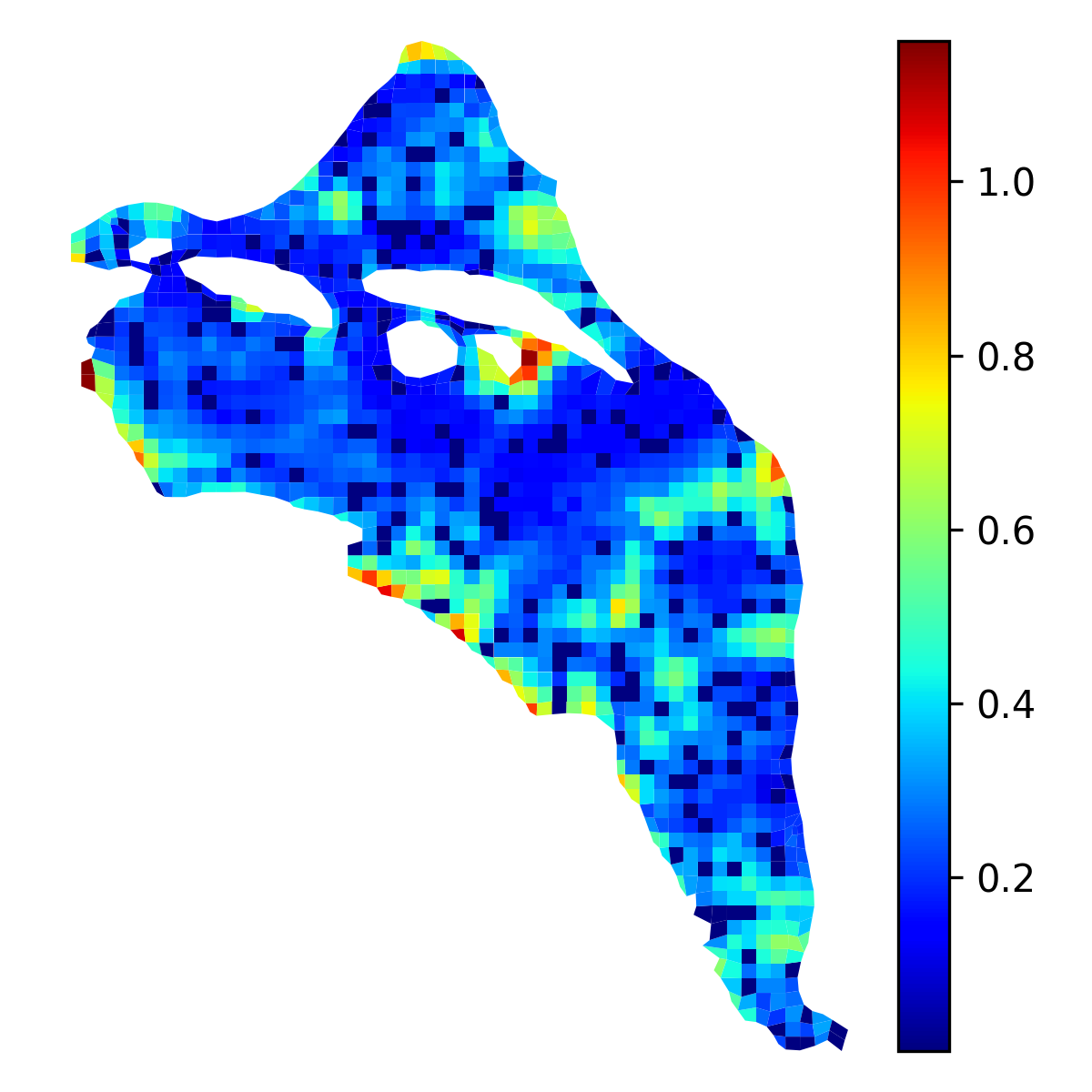}
        \includegraphics[width=0.24\textwidth]{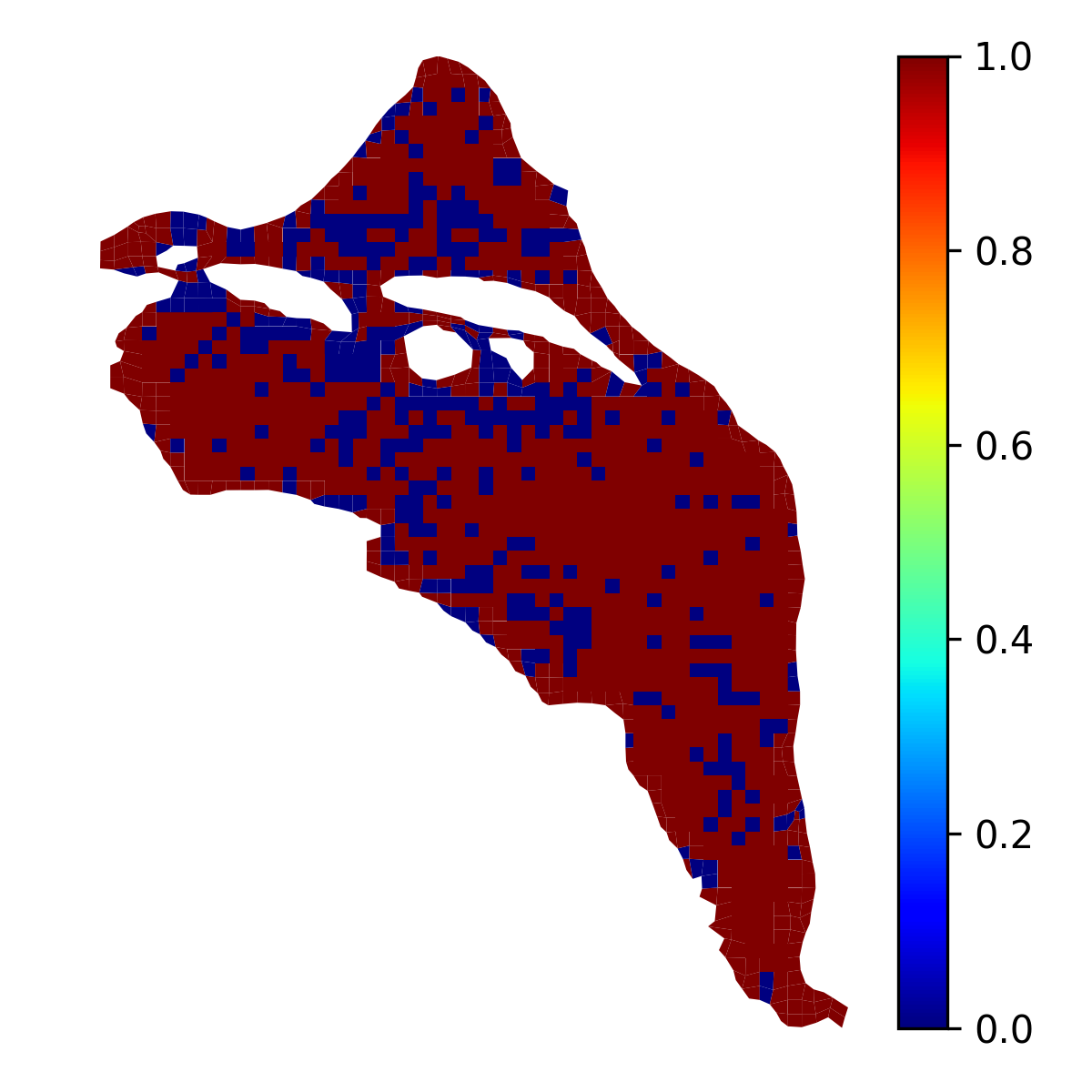}
        \caption{$\mathrm{N_y^{\text{obs}} = 200}$}
    \end{subfigure}
    
    	\caption{rPICKLE estimates of the high-dimensional  $y$ field obtained with the prior based on $\sigma^2_{r} = 10^{-4}$ and (a) $N_y^{\text{obs}} = 50$ and (b) $N_y^{\text{obs}} = 200$ observations of $y$. (First column) the posterior mean of $y$,  (second column) point errors computed as the difference between the posterior mean of $y$ and the reference $y$ field, (third column) the posterior standard deviation of $y$, and (fourth column) the coverage of the reference $y$ by the $95\%$ credibility interval.}
    \label{fig:Yref_results_sigmar_001_l2} 
\end{figure}

Next, we study uncertainty in the inverse rPICKLE solution as a function of $N_y^{\text{obs}}$. Figure \ref{fig:Yref_results_sigmar_001_l2} depicts the estimates of $y$ obtained with rPICKLE for $N_y^{\text{obs}} = 50$ and $200$ and $\sigma^2_r = 10^{-4}$. The estimates for $N_y^{\text{obs}} = 100$ are given in Figure \ref{fig:Yref_results_sigmar_01}. Table \ref{tab:yref} summarizes the relative $\ell_2$ and $\ell_{\infty}$ errors, LPP, and the percent of coverage of the corresponding posteriors. As $N_y^{\text{obs}}$ increases, the posterior mean becomes closer to $y_{\mathrm{HD}}$, and the posterior variance of $y$ decreases. The LPP increases with $N_y^{\text{obs}}$, indicating that the posterior distribution becomes more informative. Also, we see that for all values of $N_y^{\text{obs}}$, the coverage for the $y_{\mathrm{HD}}$ field is good, with the best coverage ($75 \%$) achieved for $N_y^{\text{obs}} = 200$. 

\subsection{Convergence of rPICKLE and HMC with the Ensemble Size}\label{sec:convergence}

\begin{figure}[!htb]
    \centering
    \begin{subfigure}{0.48\textwidth}
    \centering
        \includegraphics[width=0.48\textwidth]{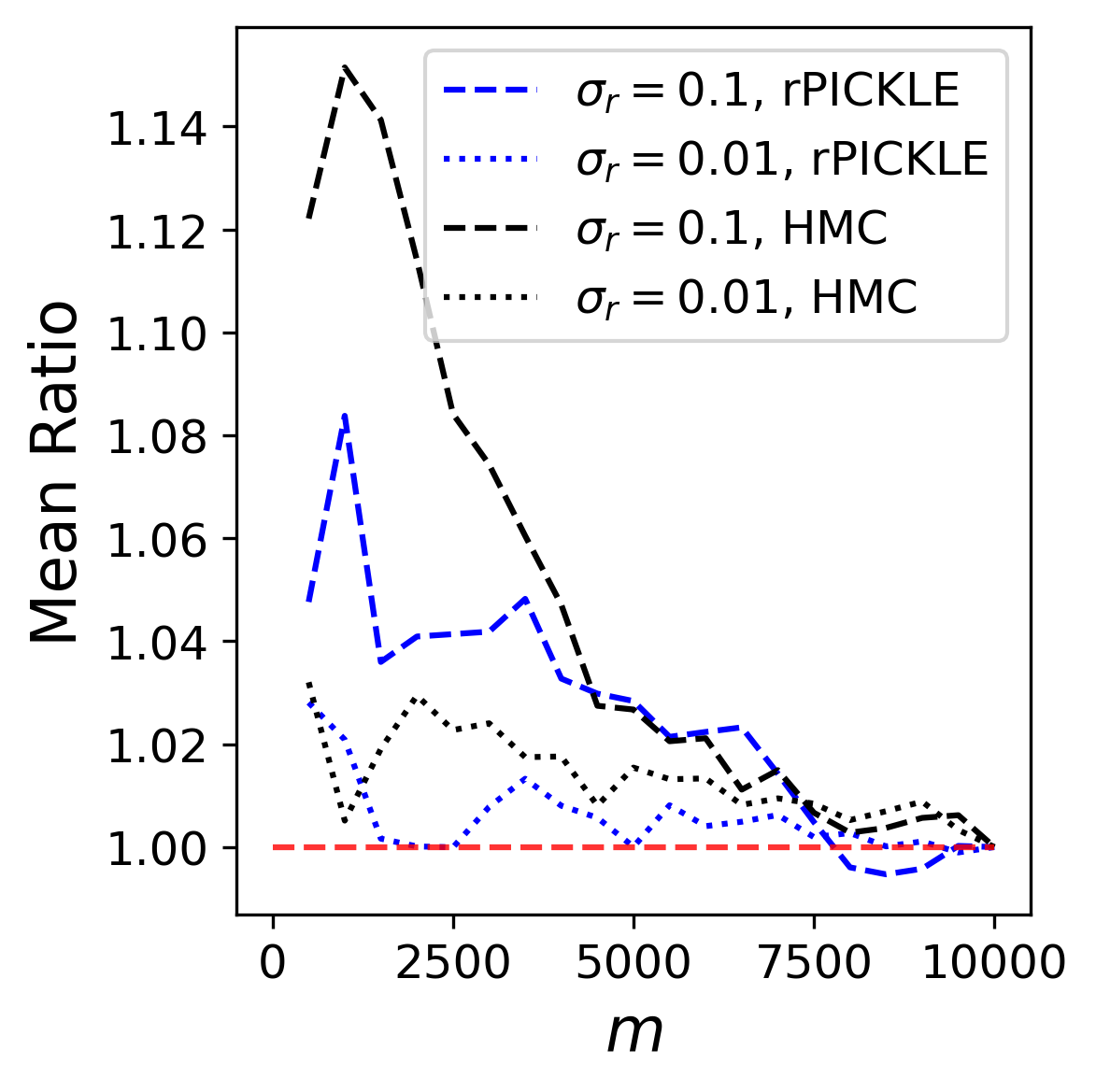}
        \includegraphics[width=0.48\textwidth]{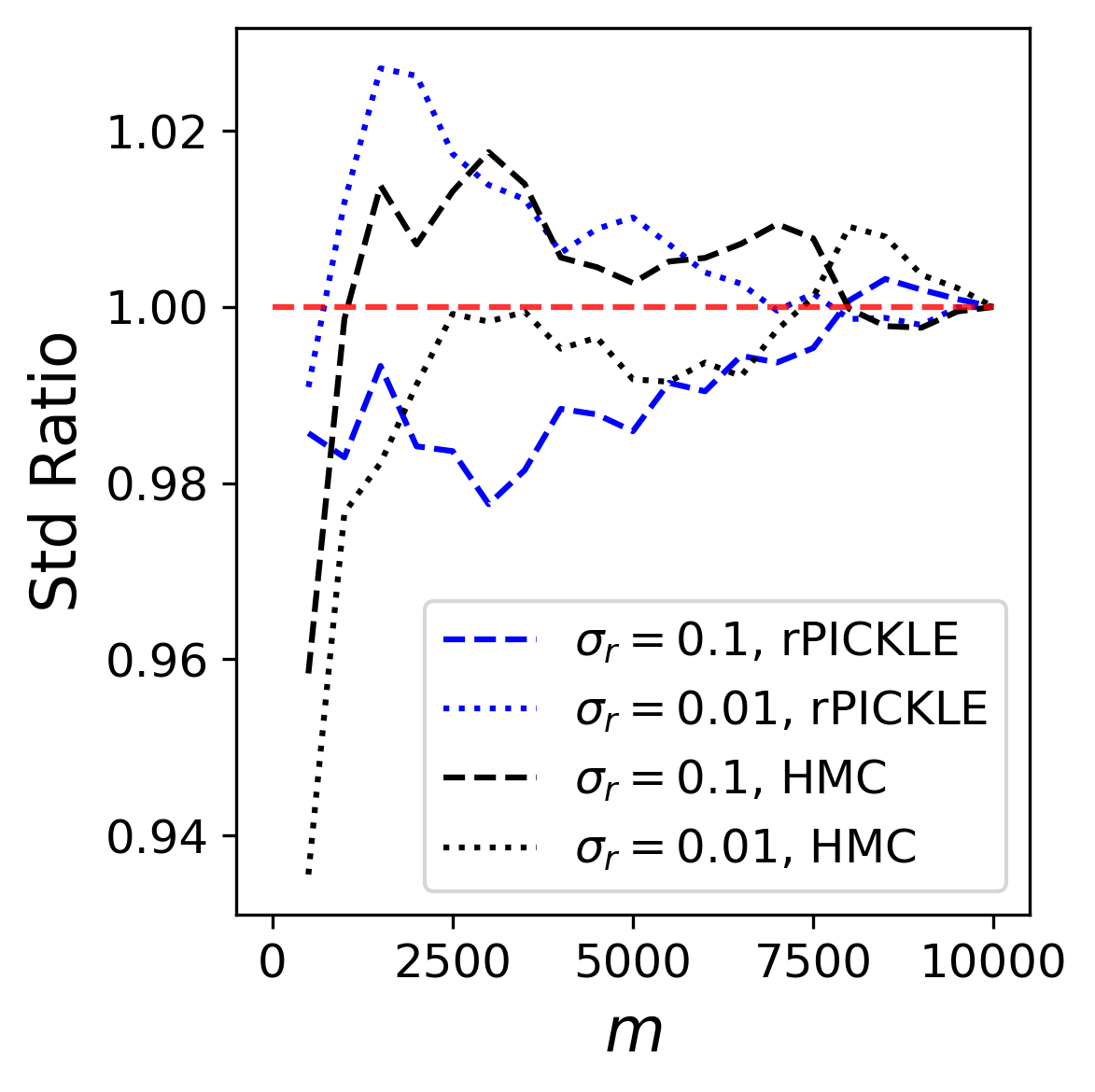}
        \caption{$y_{\mathrm{LD}}, \xi_1$}
    \end{subfigure}
    \hfill
    \begin{subfigure}{0.48\textwidth}
    \centering
        \includegraphics[width=0.48\textwidth]{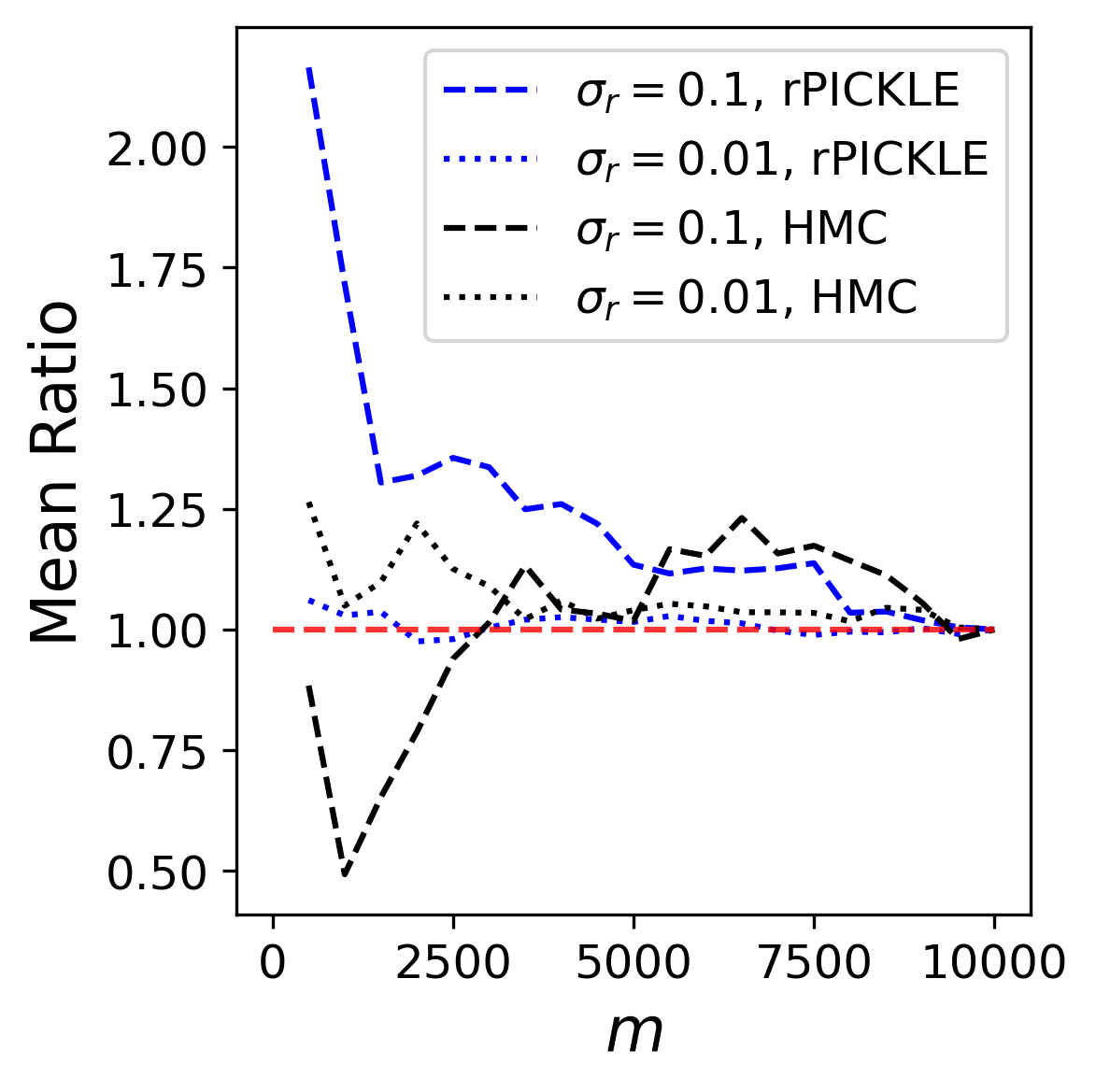}
        \includegraphics[width=0.48\textwidth]{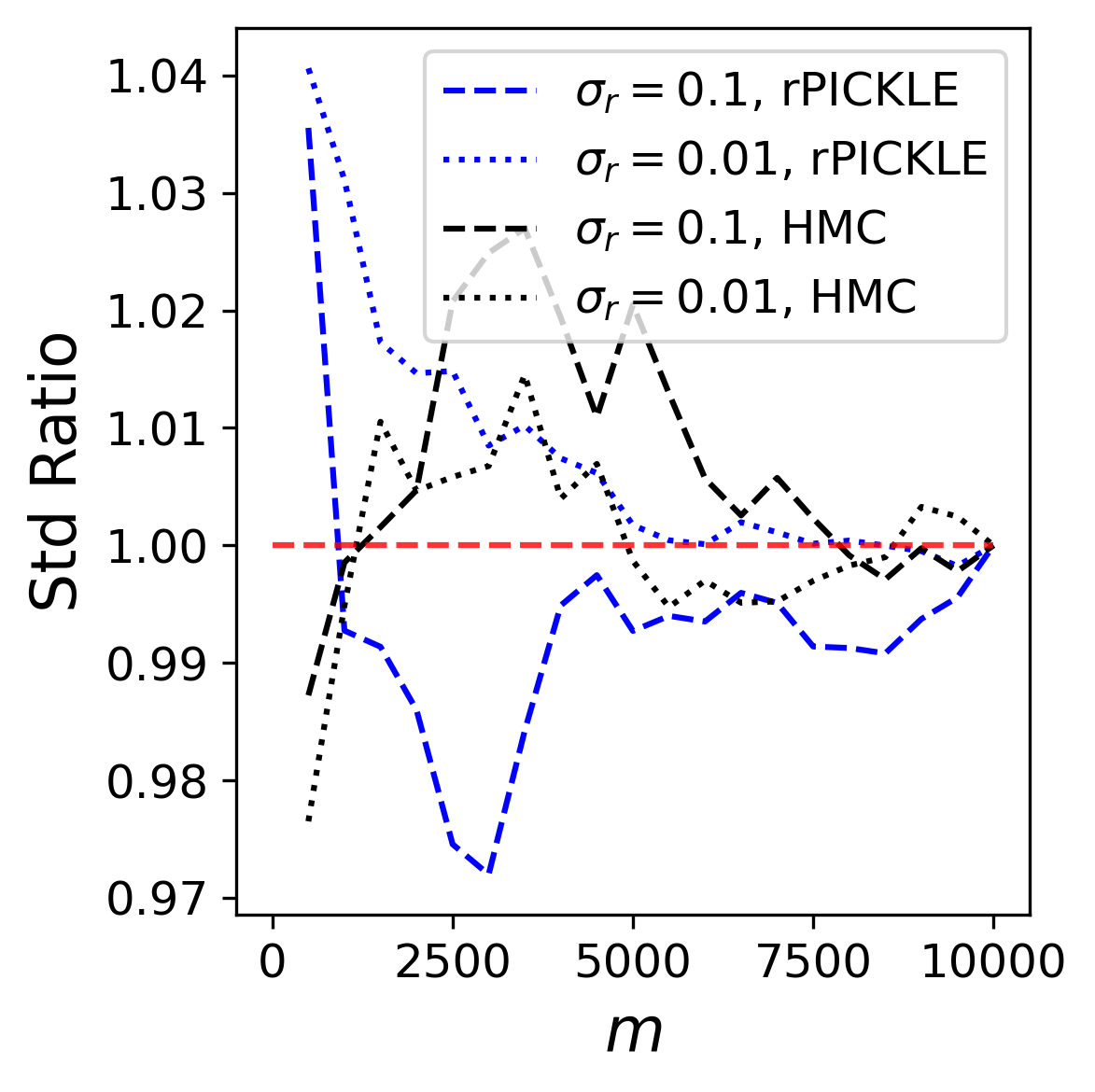}
        \caption{$y_{\mathrm{LD}}, \xi_{10}$}
    \end{subfigure}
    
    \vspace{10pt}

    \centering
    \begin{subfigure}{0.48\textwidth}
    \centering
        \includegraphics[width=0.48\textwidth]{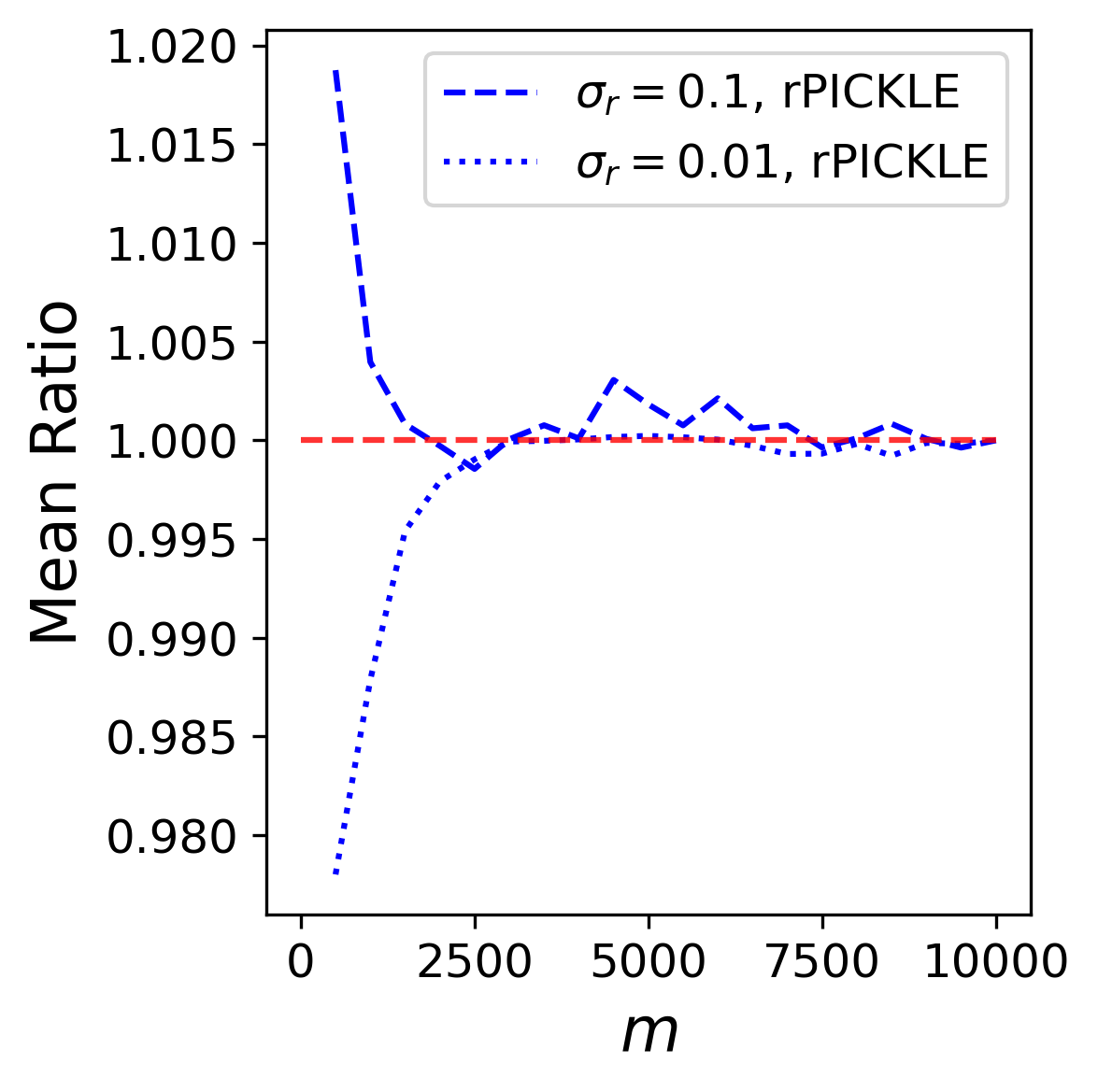}
        \includegraphics[width=0.48\textwidth]
        {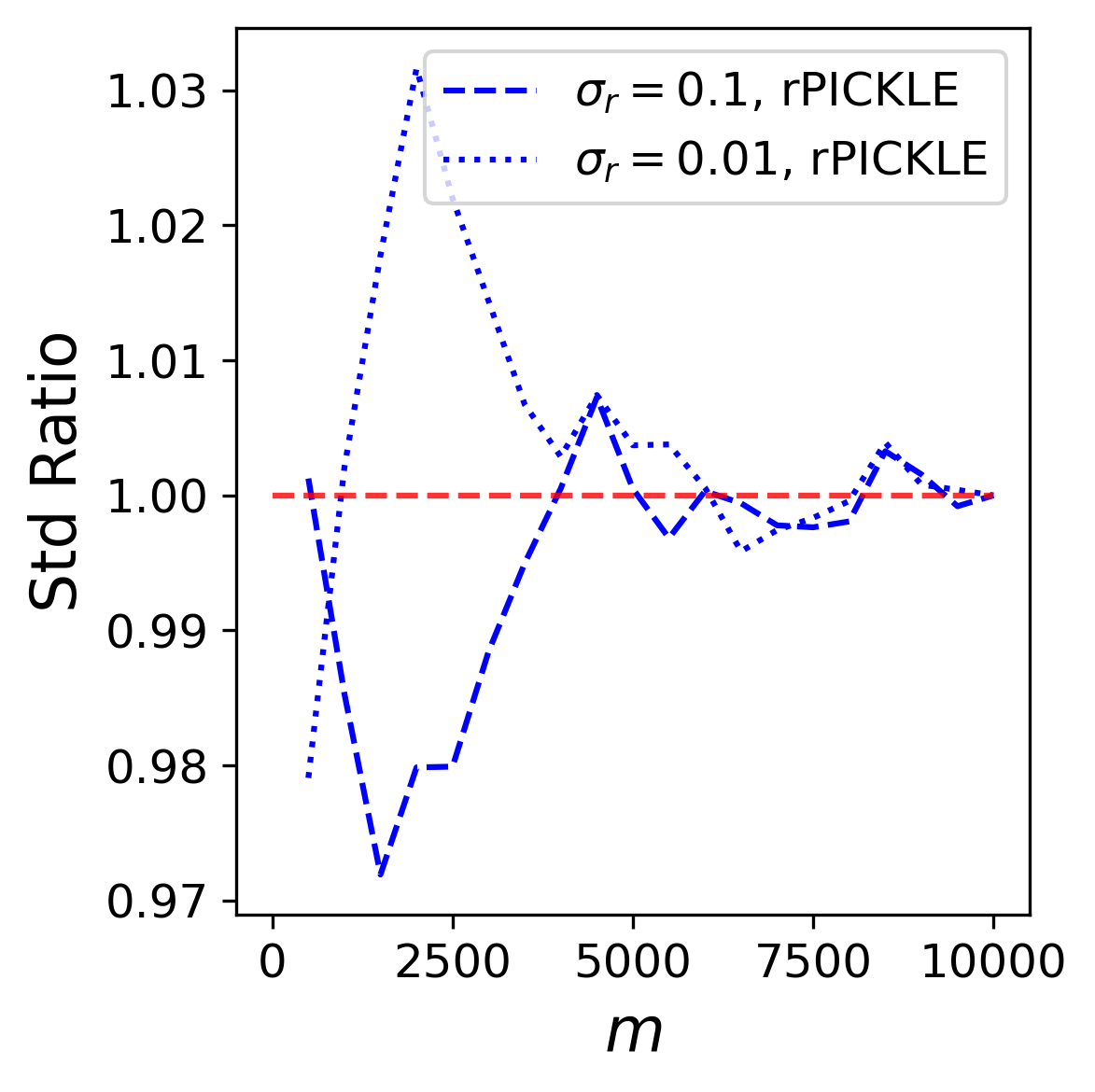}
        \caption{$y_{\mathrm{HD}}, \xi_1$}
    \end{subfigure}
    \hfill
    \begin{subfigure}{0.48\textwidth}
    \centering
        \includegraphics[width=0.48\textwidth]{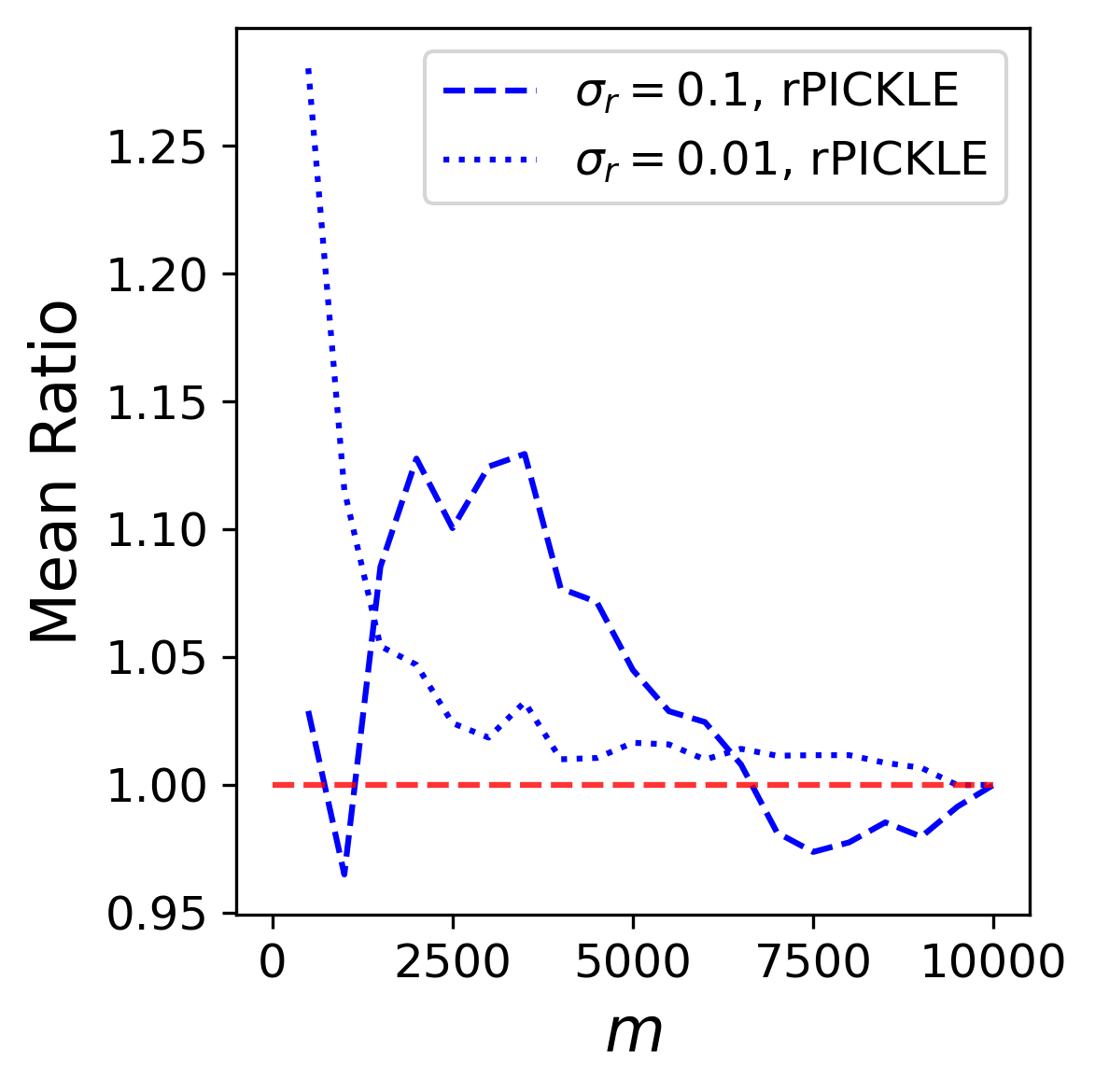}
        \includegraphics[width=0.48\textwidth]
        {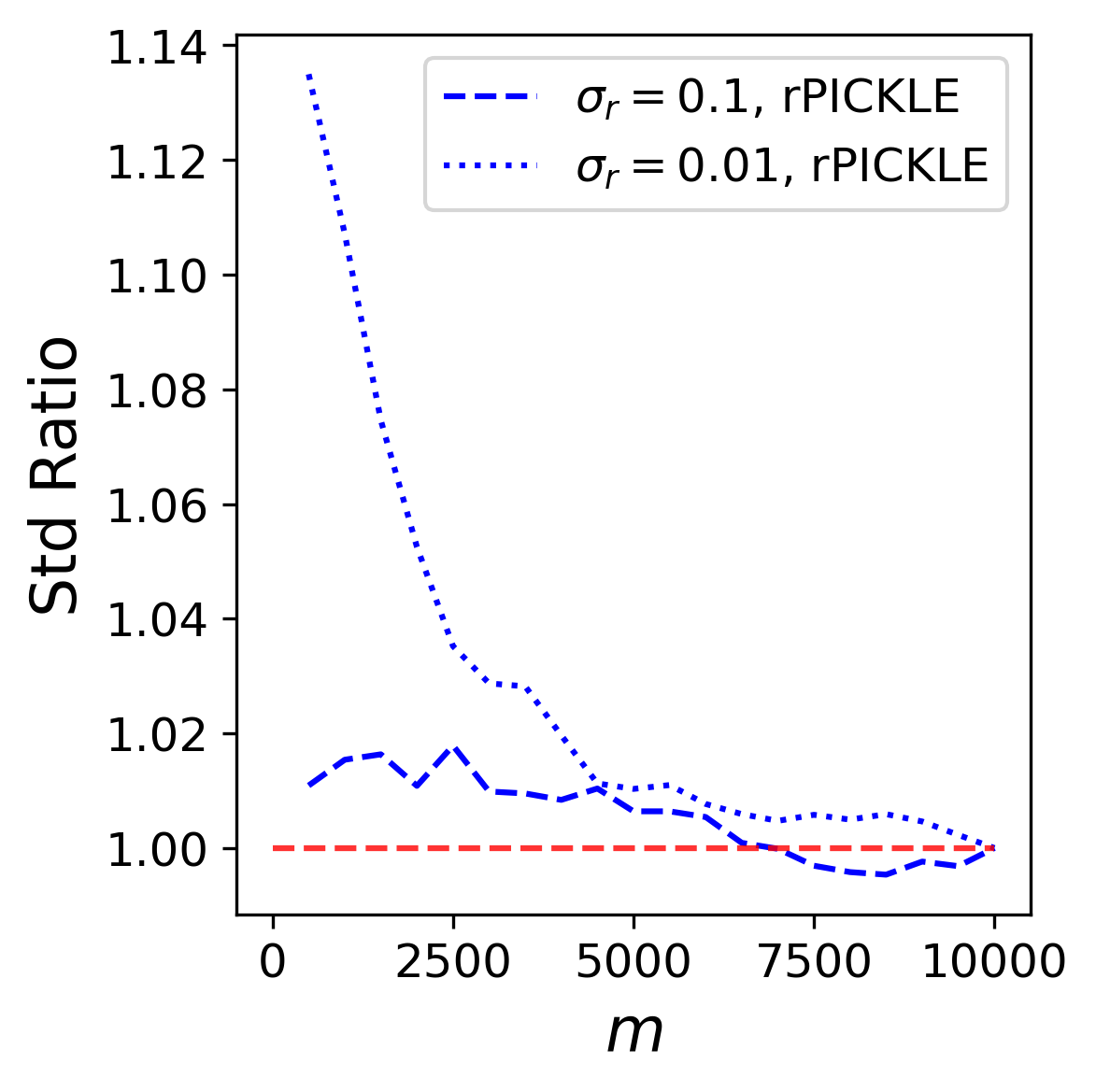}
        \caption{$y_{\mathrm{HD}}, \xi_{100}$}
    \end{subfigure}
    
    \caption{$ R_{\overline{\xi}_i^b}(m)$ and $R_{\sigma_{\xi_i}^b}(m)$ as functions of the ensemble size $m$ for the low- and high-dimensional cases and  $\sigma_r^2 = 10^{-4}$ and  $\sigma_r^2 = 10^{-2}$:   (a) $\xi_1$ of $y_{\mathrm{LD}}$, (b) $\xi_{10}$ of $y_{\mathrm{LD}}$, (c) $\xi_1$ of $y_{\mathrm{HD}}$, and (d) $\xi_{100}$ of $y_{\mathrm{LD}}$. HMC results are not available for the high-dimensional case.}
    \label{fig:convergence} 
\end{figure}

Next, we examine the convergence properties of rPICKLE and HMC  for low- and high-dimensional cases. 
For the $i$-th component of the $\boldsymbol\xi$ vector, we analyze the relative mean $\overline{\xi}^b_i (m)$ ($b = \mathrm{rPICKLE}$ or $\mathrm{HMC}$): 
\begin{eqnarray}
    R_{\overline{\xi}_i^b}(m) &=& \left| \frac{ \overline{\xi}_i^b(m)    }  {\overline{\xi}_i^{b}(N_{ens}) } \right| 
\end{eqnarray}
and relative standard deviation, $\sigma_{\xi_i}^b(m)$:
\begin{eqnarray}
    R_{\sigma_{\xi_i}^b}(m) &=& \left| \frac{\sigma_{\xi_i}^b(m)    }
    {\sigma_{\xi_i}^{b}(N_{ens}) } \right|.
\end{eqnarray}
as functions of the ensemble size $m$ ($N_{ens} = 10^4$ is the maximum ensemble size). 
Figure \ref{fig:convergence} shows the dependence of $R_{\overline{\xi}_i^b}(m)$ and  $ R_{\sigma_{\xi_i}^b}(m) $  on $m$ for the low-dimensional case ($i=1$ and 10) and the high-dimensional case ($i=1$ and 100). Here, we set $\sigma^2_r = 10^{-2}$ and $10^{-4}$. For the high-dimensional case, the number of $y$ observations for the high-dimensional case is set to $N_y^{\text{obs}} = 100$, and we only show the convergence of rPICKLE because of the prohibitively large computational time of HMC. 

Figure \ref{fig:convergence} shows 
that for the low-dimensional case, the convergence properties of HMC and rPICKLE are very similar. We also find that in both methods, the required number of samples for mean and variance to reach asymptotic values increases with $\sigma_r$.
Furthermore, we see that in rPICKLE, the required number of samples is not significantly affected by the dimensionality, which is to be expected because the rPICKLE samples are generated independently from one another.

 It should be noted that, as shown in Section \ref{sec:hmc_failure}, the condition number of the posterior covariance increases with decreasing $\sigma_r$, which decreases the time step in the HMC algorithm. As a result, we find that the computational time of HMC to get a set number of samples increases with decreasing  $\sigma_r$. On the other hand, the computational time of rPICKLE is not significantly affected by the value of  $\sigma_r$.

\subsection{HMC Performance for low- and High-dimensional Problems}\label{sec:hmc_failure}

In this section, we investigate the effect of $\sigma^2_r$ and OF problem dimensionality on the efficiency of HMC.

As mentioned earlier, for certain values of $\sigma^2_r$, the HMC does not reach the stopping criterion ($10^4$ samples) after more than 30 days of running the code. For comparison, rPICKLE generates $10^4$ samples in four to five days depending on the value of $\sigma^2_r$. For the low-dimensional case, rPICKLE takes approximately 4 minutes to generate the same number of samples for all considered $\sigma^2_r$, while the computational time of HMC varies from 3 to 6 hours depending on the $\sigma^2_r$ value.  

We find that the increase in HMC computational time is mainly due to smaller time steps in the integration of Hamiltonian dynamics equations required to maintain a desirable acceptance rate in the dual averaging algorithm.
It was shown in \cite{Girolami2011RiemannML, langmore2023hmcillcond} that the large condition number of the posterior covariance matrix of unknown parameters leads to a decrease in the HMC performance. 
Here, we demonstrate that the condition number increases with increasing problem dimensionality ($N_\xi + N_\eta$) and decreasing $\sigma_r^2$.

In theory, the posterior covariance can be computed directly from posterior samples obtained, for example, from rPICKLE. Here, we focus on a priori estimates of the posterior covariance that can be used as a criterion for using HMC. To obtain an a priori estimate of $\Sigma_{post}$, we employ the Laplace assumption and approximate the posterior covariance with the inverse of the Hessian of the log posterior.  

We start by approximating the log posterior \eqref{eq:Bayesian_posterior} using the Taylor expansion around the MAP point (which we assume is known from PICKLE) as
\begin{eqnarray}\label{eq:Laplace_approximation}
\log P(\boldsymbol\zeta | \mathcal{D}_{res}) & \approx & \log P(\boldsymbol\zeta^* | \mathcal{D}_{res}) \nonumber \\
&+&  \frac{1}{2}(\boldsymbol\zeta - \boldsymbol\zeta^*)^T(\nabla\nabla \log P(\boldsymbol\zeta | \mathcal{D}_{res})|_{\boldsymbol\zeta = \boldsymbol\zeta^*})(\boldsymbol\zeta - \boldsymbol\zeta^*), 
\end{eqnarray}
where $\boldsymbol\zeta^*$ is the MAP, and $\nabla\nabla \log P(\boldsymbol\zeta | \mathcal{D}_{res})|_{\boldsymbol\zeta = \boldsymbol\zeta^*}$ is the Hessian, which we compute by automatic differentiation, evaluated at the MAP. The first-order term vanishes because the gradient of $\log P$ at the MAP is zero.  
The right-hand side of Eq.~\eqref{eq:Laplace_approximation} is equivalent, up to a constant, to the log probability density of a Gaussian distribution. Under the Laplace assumption,  the posterior distribution can be approximated with a Gaussian distribution, the mean of which is given by the MAP. The covariance is found by the inverse of the Hessian evaluated at the MAP point $(\nabla\nabla \log P(\boldsymbol\zeta | \mathcal{D}_{res})|_{\boldsymbol\zeta = \boldsymbol\zeta^*})^{-1}$. 

\begin{figure}[!htb]
	\centering
	\begin{subfigure}[b]{0.4\textwidth}
        \includegraphics[width=\textwidth]{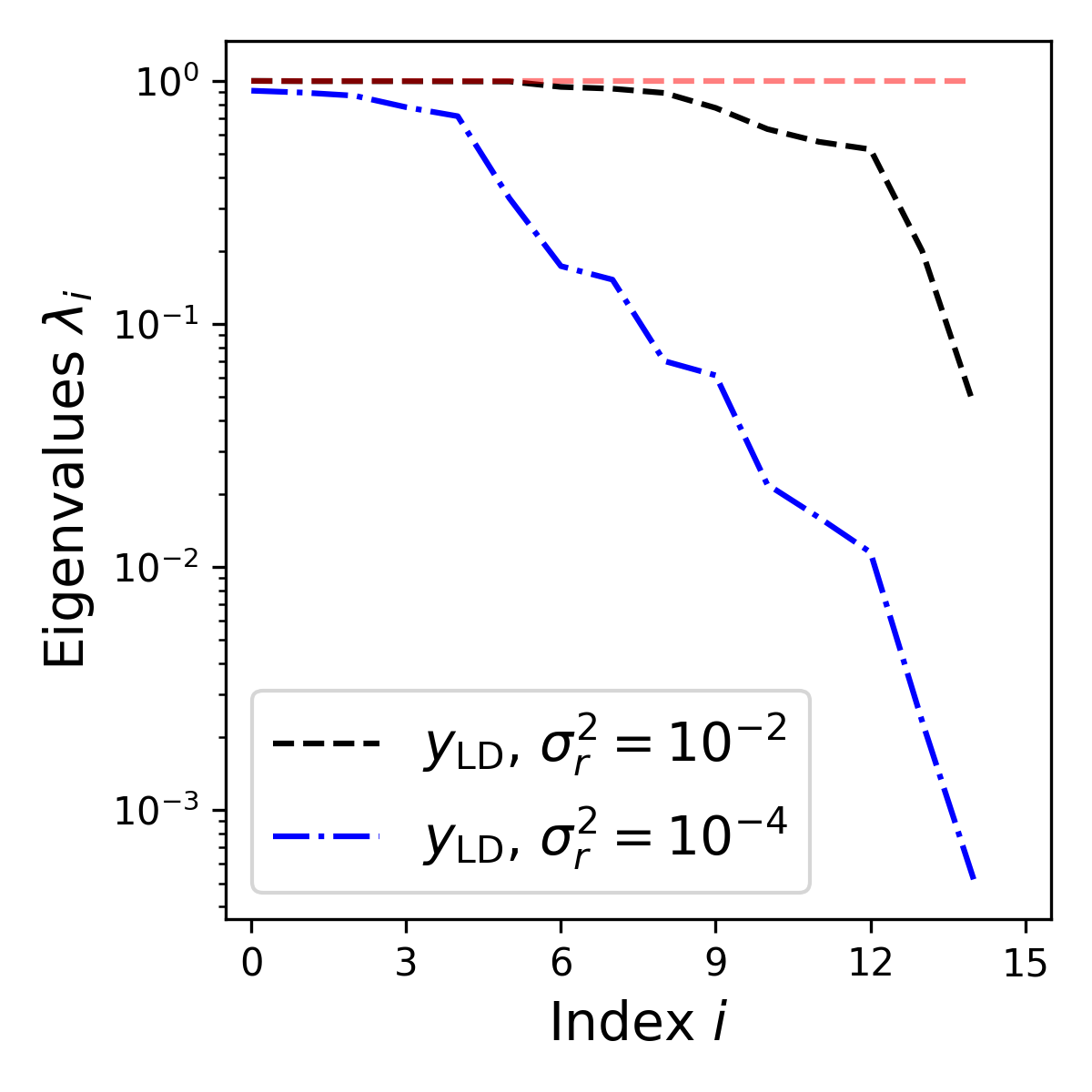}\caption{}
        \label{fig:Ysmooth_covariance_spectrum}
    \end{subfigure}
    \begin{subfigure}[b]{0.4\textwidth}
        \includegraphics[width=\textwidth]{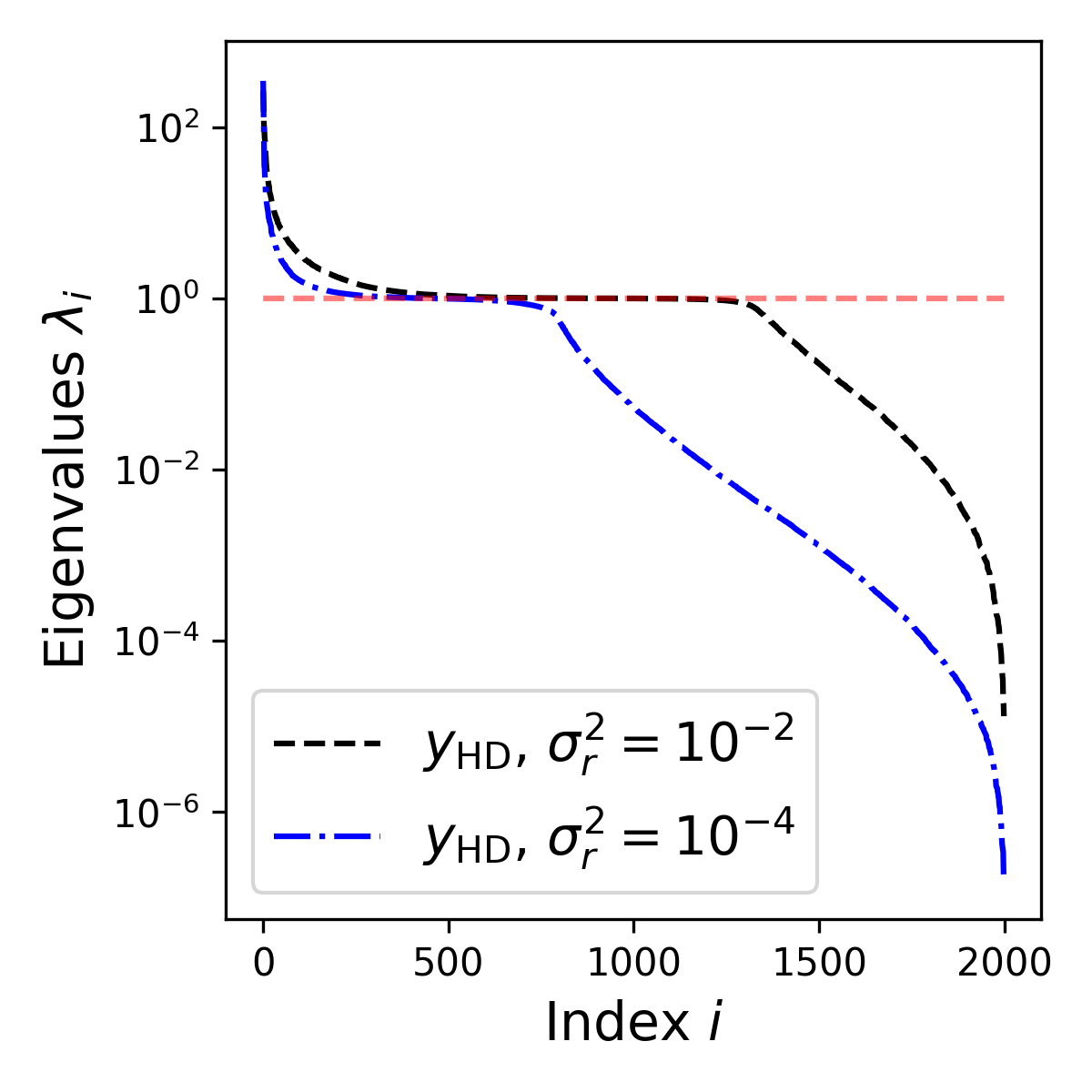}\caption{}
        \label{fig:Yref_covariance_spectrum}
    \end{subfigure}
	\caption{Covariance spectrum corresponding to the Laplace approximated posterior for (a) $y_{\mathrm{LD}}$ and (b) $y_{\mathrm{HD}}$ for $\sigma^2_r=10^{-4}$ and $10^{-2}$. The red dashed line represents the covariance structure informed by the normal prior. 
	}
	\label{fig:covariance_spectrum}
\end{figure}

Figure \ref{fig:covariance_spectrum} shows the eigenvalues (arranged in descending order) of the approximated posterior covariance for the low- and high-dimensional cases with $\sigma^2_r = 10^{-2}$ and $10^{-4}$. The red dashed lines indicate eigenvalues of the prior covariance (all eigenvalues are equal to one because of the diagonal form of the prior covariance and unit prior variances of the parameters). 
In the low-dimensional problem, the condition numbers are approximately 21 for $\sigma^2_r = 10^{-2}$ and 1760 for $\sigma^2_r = 10^{-4}$.
In the high-dimensional problem, the condition numbers are $\approx 2\times10^7$ for $\sigma^2_r = 10^{-2}$ and $2\times10^9$ for $\sigma^2_r = 10^{-4}$. 

Larger condition numbers indicate the presence of a stronger correlation between the components of the estimated $\boldsymbol{\xi}$.  Also, 
the eigenvalues $\lambda_i$ of $\Sigma_{post}$ are proportional to the variances $\sigma^2_i$ of the posterior marginal distributions of the $\xi_i$ components of $\boldsymbol\xi$. Therefore, a large condition number indicates a large range of $\sigma^2_i$ values, which gives rise to geometrically pathological features of the posterior parameter space (e.g., high curvature in the corner of posterior manifolds). The increase in correlation and the range of $\sigma^2_i$ with $\sigma_r^2$ and $N_\xi$ can be seen in Figures \ref{fig:Ysmooth_joint_posterior} and \ref{fig:Yref_joint_posterior}. These geometric complexities reduce the efficiency of HMC \cite{betancourt2017conceptual}.

\section{Discussions and Conclusion}\label{discussion}

We presented the rPICKLE method, an approximate Bayesian inference method for sampling high-dimensional posterior distributions of unknown parameters. The rPICKLE is derived by randomizing the PICKLE's objective function. 
 In the Bayesian framework, the posterior distribution of the estimated parameters depends on the choice of the prior parameter distribution and the likelihood function, which expresses the prior knowledge or assumptions about the modeled system. In the application of rPICKLE to the diffusion (Darcy) equation with an unknown space-dependent diffusion coefficient (transmissivity of an aquifer), it is assumed that the log-diffusion coefficient $y(x)$ has a Gaussian prior distribution with mean and covariance computed from $y$ measurements. This prior model imposes the form of eigenfunctions in the CKLE expansion of $y$ and also the prior distribution of the parameters $\boldsymbol{\xi}$ in this CKLE. The likelihood is defined by the residual least-square formulation of the governing PDE model in the form of the independent zero-mean Gaussian distribution of the residuals with the variance $\sigma^2_r$, which becomes a single free parameter controlling the posterior distribution of $\boldsymbol\xi$ (and $y(x)$).  
 
The goal of the Bayesian parameter estimation is to obtain a  posterior distribution providing a good statistical description of the observed system behavior. Here, we quantify the predictive capacity of the posterior in terms of the distance from the posterior mean or mode to the reference field (could be validation data if the reference field is not available), the percent of coverage, and the LPP. This metric can be used for selecting $\sigma^2_r$ to minimize the distance and maximize LPP. 

We demonstrated that PICKLE provides the mode of the posterior distribution if the PICKLE's regularization coefficient $\gamma$ is set to $\sigma^2_r$. 
Because the PICKLE mode estimation does not require sampling of the posterior, choosing $\sigma^2_r=\gamma$ to minimize the distance from the posterior mode to the validation data provides a computational advantage. In general, there is no guarantee that $\sigma^2_r$ selected according to this criterion would also maximize LPP and coverage. 
However, we found that for considered problems, a $\sigma^2_r$ that minimizes the distance between the mode of the posterior distribution also maximizes the LPP of the posterior distribution. We also found that LPP is more sensitive to $\sigma^2_r$ than the distance between the mode and data. Therefore, LPP should also be considered in selecting $\sigma^2_r$.

The robustness of rPICKLE was demonstrated by using it for estimating log-transmissivity $y$ in the high-dimensional Hanford site groundwater flow model with 2000 parameters in the CKLE  representations of the parameter and state variables. We found that rPICKLE produces posteriors for a wide range of $\sigma^2_r$ with the posterior mean of $y$ being close to the MAP of $y$ given by PICKLE. On the other hand, we found that HMC was not able to reach a stopping criterion ($10^4$ samples) after running for more than one month. For comparison, the rPICKLE code generated the same number of samples in four to five days depending on the value of $\sigma^2_r$.

To enable comparison between rPICKLE and HMC, we also considered a lower-dimensional problem where the parameters and states were represented with 15 CKLE parameters. 
For this problem, we found an excellent agreement between rPICKLE and HMC. We also found the $y$ predictions given by the posterior mean of rPICKLE and HMC to be in good agreement with the $y$ prediction given by the PICKLE mode. For this low-dimensional problem, rPICKLE generated $10^4$ samples in approximately 4 minutes regardless of the considered values of $\sigma^2_r$. The HMC time was found to increase from $\approx$ 3 hours for  $\sigma^2_r=1$ to $\approx$ 6 hours for  $\sigma^2_r=10^{-5}$. 
  
The efficiency of HMC is known to decrease with the increasing condition number of the posterior covariance matrix. We demonstrated that for the considered problem, the condition number increases with increasing dimensionality and decreasing $\sigma^2_r$, which also explains the observed trend in the HMC computational time. 

In summary, our results demonstrate the advantage of rPICKLE for high-dimensional problems with strict physics constraints (small values of $\sigma^2_r$).

\section{Acknowledgements}
This research was partially supported by the U.S. Department of Energy (DOE) Advanced Scientific
Computing program, 
the DOE project ``Science-Informed Machine Learning to Accelerate Real-time (SMART) Decisions in Subsurface Applications Phase 2 – Development and Field Validation,''
and the United States National Science Foundation. Pacific Northwest National Laboratory is
operated by Battelle for the DOE under Contract DE-AC05-76RL01830. 

\bibliographystyle{unsrtnat}
\bibliography{reference}

\begin{thebibliography}{39}
\providecommand{\natexlab}[1]{#1}
\providecommand{\url}[1]{\texttt{#1}}
\expandafter\ifx\csname urlstyle\endcsname\relax
  \providecommand{\doi}[1]{doi: #1}\else
  \providecommand{\doi}{doi: \begingroup \urlstyle{rm}\Url}\fi

\bibitem[Zhou et~al.(2014)Zhou, G{\'o}mez-Hern{\'a}ndez, and
  Li]{zhou2014inverse}
Haiyan Zhou, J~Jaime G{\'o}mez-Hern{\'a}ndez, and Liangping Li.
\newblock Inverse methods in hydrogeology: Evolution and recent trends.
\newblock \emph{Advances in Water Resources}, 63:\penalty0 22--37, 2014.

\bibitem[Linde et~al.(2017)Linde, Ginsbourger, Irving, Nobile, and
  Doucet]{linde2017uncertainty}
Niklas Linde, David Ginsbourger, James Irving, Fabio Nobile, and Arnaud Doucet.
\newblock On uncertainty quantification in hydrogeology and hydrogeophysics.
\newblock \emph{Advances in Water Resources}, 110:\penalty0 166--181, 2017.

\bibitem[Tartakovsky et~al.(2020)Tartakovsky, Marrero, Perdikaris, Tartakovsky,
  and Barajas-Solano]{tartakovsky2020pinn}
Alexandre~M Tartakovsky, C~Ortiz Marrero, Paris Perdikaris, Guzel~D
  Tartakovsky, and David Barajas-Solano.
\newblock Physics-informed deep neural networks for learning parameters and
  constitutive relationships in subsurface flow problems.
\newblock \emph{Water Resources Research}, 56\penalty0 (5):\penalty0
  e2019WR026731, 2020.

\bibitem[He and Tartakovsky(2021)]{He2021WRR}
QiZhi He and Alexandre~M. Tartakovsky.
\newblock Physics-informed neural network method for forward and backward
  advection-dispersion equations.
\newblock \emph{Water Resources Research}, 57\penalty0 (7):\penalty0
  e2020WR029479, 2021.

\bibitem[Zong et~al.(2023)Zong, He, and Tartakovsky]{Zong2023CMAME}
Yifei Zong, QiZhi He, and Alexandre~M. Tartakovsky.
\newblock Improved training of physics-informed neural networks for parabolic
  differential equations with sharply perturbed initial conditions.
\newblock \emph{Computer Methods in Applied Mechanics and Engineering},
  414:\penalty0 116125, 2023.

\bibitem[Yeung et~al.(2022)Yeung, Barajas-Solano, and
  Tartakovsky]{Yeung2021PICKLE}
Yu-Hong Yeung, David~A. Barajas-Solano, and Alexandre~M. Tartakovsky.
\newblock Physics-informed machine learning method for large-scale data
  assimilation problems.
\newblock \emph{Water Resources Research}, 58\penalty0 (5):\penalty0
  e2021WR031023, 2022.
\newblock \doi{https://doi.org/10.1029/2021WR031023}.
\newblock URL
  \url{https://agupubs.onlinelibrary.wiley.com/doi/abs/10.1029/2021WR031023}.
\newblock e2021WR031023 2021WR031023.

\bibitem[Anderson et~al.(2015)Anderson, Woessner, and
  Hunt]{anderson2015applied}
Mary~P Anderson, William~W Woessner, and Randall~J Hunt.
\newblock \emph{Applied groundwater modeling: simulation of flow and advective
  transport}.
\newblock Academic press, 2015.

\bibitem[RamaRao et~al.(1995)RamaRao, LaVenue, De~Marsily, and
  Marietta]{ramarao1995pilot}
Banda~S RamaRao, A~Marsh LaVenue, Ghislain De~Marsily, and Melvin~G Marietta.
\newblock Pilot point methodology for automated calibration of an ensemble of
  conditionally simulated transmissivity fields: 1. theory and computational
  experiments.
\newblock \emph{Water Resources Research}, 31\penalty0 (3):\penalty0 475--493,
  1995.

\bibitem[Doherty et~al.(2010)Doherty, Fienen, and Hunt]{doherty2010ppm}
John~E Doherty, Michael~N Fienen, and Randall~J Hunt.
\newblock \emph{Approaches to highly parameterized inversion: Pilot point
  theory, guidelines, and research directions}, volume 2010.
\newblock US Department of the Interior, US Geological Survey, 2010.

\bibitem[Tonkin and Doherty(2005)]{tonkin2005hybrid}
Matthew~James Tonkin and John Doherty.
\newblock A hybrid regularized inversion methodology for highly parameterized
  environmental models.
\newblock \emph{Water Resources Research}, 41\penalty0 (10), 2005.

\bibitem[Marzouk and Najm(2009)]{marzouk2009dimensionality}
Youssef~M Marzouk and Habib~N Najm.
\newblock Dimensionality reduction and polynomial chaos acceleration of
  bayesian inference in inverse problems.
\newblock \emph{Journal of Computational Physics}, 228\penalty0 (6):\penalty0
  1862--1902, 2009.

\bibitem[Li and Tartakovsky(2020)]{LI2020JCP}
Jing Li and Alexandre~M. Tartakovsky.
\newblock Gaussian process regression and conditional polynomial chaos for
  parameter estimation.
\newblock \emph{Journal of Computational Physics}, page 109520, 2020.

\bibitem[Tartakovsky et~al.(2021)Tartakovsky, Barajas-Solano, and
  He]{tartakovsky2021PICKLE}
Alexandre~M Tartakovsky, David~A Barajas-Solano, and Qizhi He.
\newblock Physics-informed machine learning with conditional karhunen-lo{\`e}ve
  expansions.
\newblock \emph{Journal of Computational Physics}, 426:\penalty0 109904, 2021.

\bibitem[Stuart(2010)]{stuart2010inverse}
Andrew~M Stuart.
\newblock Inverse problems: a bayesian perspective.
\newblock \emph{Acta numerica}, 19:\penalty0 451--559, 2010.

\bibitem[Herckenrath et~al.(2011)Herckenrath, Langevin, and
  Doherty]{Langevin2011NSMC}
Daan Herckenrath, Christian~D. Langevin, and John Doherty.
\newblock Predictive uncertainty analysis of a saltwater intrusion model using
  null-space monte carlo.
\newblock \emph{Water Resources Research}, 47\penalty0 (5), 2011.

\bibitem[Yoon et~al.(2013)Yoon, Hart, and McKenna]{Yoon2013nullspaceMC}
Hongkyu Yoon, David~B. Hart, and Sean~A. McKenna.
\newblock Parameter estimation and predictive uncertainty in stochastic inverse
  modeling of groundwater flow: Comparing null-space monte carlo and multiple
  starting point methods.
\newblock \emph{Water Resources Research}, 49\penalty0 (1):\penalty0 536--553,
  2013.

\bibitem[Brooks(1998)]{brooks1998markov}
Stephen Brooks.
\newblock Markov chain monte carlo method and its application.
\newblock \emph{Journal of the royal statistical society: series D (the
  Statistician)}, 47\penalty0 (1):\penalty0 69--100, 1998.

\bibitem[Abdar et~al.(2021)Abdar, Pourpanah, Hussain, Rezazadegan, Liu,
  Ghavamzadeh, Fieguth, Cao, Khosravi, Acharya, et~al.]{abdar2021UQDLreview}
Moloud Abdar, Farhad Pourpanah, Sadiq Hussain, Dana Rezazadegan, Li~Liu,
  Mohammad Ghavamzadeh, Paul Fieguth, Xiaochun Cao, Abbas Khosravi, U~Rajendra
  Acharya, et~al.
\newblock A review of uncertainty quantification in deep learning: Techniques,
  applications and challenges.
\newblock \emph{Information Fusion}, 76:\penalty0 243--297, 2021.

\bibitem[Sun and Wang(2020)]{sun2020physics}
Luning Sun and Jian-Xun Wang.
\newblock Physics-constrained bayesian neural network for fluid flow
  reconstruction with sparse and noisy data.
\newblock \emph{Theoretical and Applied Mechanics Letters}, 10\penalty0
  (3):\penalty0 161--169, 2020.

\bibitem[Gou et~al.(2022)Gou, Zhang, and Zhu]{gou2022bayesian}
Rongxi Gou, Yijie Zhang, and Xueyu Zhu.
\newblock Bayesian physics-informed neural networks for seismic tomography
  based on the eikonal equation.
\newblock \emph{arXiv preprint arXiv:2203.12351}, 2022.

\bibitem[Blei et~al.(2017)Blei, Kucukelbir, and McAuliffe]{blei2017vi}
David~M Blei, Alp Kucukelbir, and Jon~D McAuliffe.
\newblock Variational inference: A review for statisticians.
\newblock \emph{Journal of the American statistical Association}, 112\penalty0
  (518):\penalty0 859--877, 2017.

\bibitem[Yao et~al.(2019)Yao, Pan, Ghosh, and Doshi-Velez]{yao2019quality}
Jiayu Yao, Weiwei Pan, Soumya Ghosh, and Finale Doshi-Velez.
\newblock Quality of uncertainty quantification for bayesian neural network
  inference.
\newblock \emph{arXiv preprint arXiv:1906.09686}, 2019.

\bibitem[Neal(2011)]{neal2011mcmc}
Radford~M Neal.
\newblock Mcmc using hamiltonian dynamics.
\newblock \emph{Handbook of Markov Chain Monte Carlo}, 2\penalty0
  (11):\penalty0 2, 2011.

\bibitem[Fichtner et~al.(2019)Fichtner, Zunino, and
  Gebraad]{fichtner2019hmctopo}
Andreas Fichtner, Andrea Zunino, and Lars Gebraad.
\newblock Hamiltonian monte carlo solution of tomographic inverse problems.
\newblock \emph{Geophysical Journal International}, 216\penalty0 (2):\penalty0
  1344--1363, 2019.

\bibitem[Langmore et~al.(2023)Langmore, Dikovsky, Geraedts, Norgaard, and
  Von~Behren]{langmore2023hmcillcond}
Ian Langmore, Michael Dikovsky, Scott Geraedts, Peter Norgaard, and Rob
  Von~Behren.
\newblock Hamiltonian monte carlo in inverse problems. ill-conditioning and
  multimodality.
\newblock \emph{International Journal for Uncertainty Quantification},
  13\penalty0 (1), 2023.

\bibitem[Betancourt(2017)]{betancourt2017conceptual}
Michael Betancourt.
\newblock A conceptual introduction to hamiltonian monte carlo.
\newblock \emph{arXiv preprint arXiv:1701.02434}, 2017.

\bibitem[Wang et~al.(2018)Wang, Bui-Thanh, and Ghattas]{wang2018rmap}
Kainan Wang, Tan Bui-Thanh, and Omar Ghattas.
\newblock A randomized maximum a posteriori method for posterior sampling of
  high dimensional nonlinear bayesian inverse problems.
\newblock \emph{SIAM Journal on Scientific Computing}, 40\penalty0
  (1):\penalty0 A142--A171, 2018.

\bibitem[Chen and Oliver(2012)]{chen2012ensemble}
Yan Chen and Dean~S Oliver.
\newblock Ensemble randomized maximum likelihood method as an iterative
  ensemble smoother.
\newblock \emph{Mathematical Geosciences}, 44:\penalty0 1--26, 2012.

\bibitem[Bardsley et~al.(2014)Bardsley, Solonen, Haario, and
  Laine]{bardsley2014randomize}
Johnathan~M Bardsley, Antti Solonen, Heikki Haario, and Marko Laine.
\newblock Randomize-then-optimize: A method for sampling from posterior
  distributions in nonlinear inverse problems.
\newblock \emph{SIAM Journal on Scientific Computing}, 36\penalty0
  (4):\penalty0 A1895--A1910, 2014.

\bibitem[White(2018)]{white_ies_2018}
Jeremy~T. White.
\newblock A model-independent iterative ensemble smoother for efficient
  history-matching and uncertainty quantification in very high dimensions.
\newblock \emph{Environmental Modelling \& Software}, 109:\penalty0 191--201,
  2018.
\newblock ISSN 1364-8152.
\newblock \doi{https://doi.org/10.1016/j.envsoft.2018.06.009}.
\newblock URL
  \url{https://www.sciencedirect.com/science/article/pii/S1364815218302676}.

\bibitem[Rasmussen et~al.(2006)Rasmussen, Williams,
  et~al.]{rasmussen2006gaussian}
Carl~Edward Rasmussen, Christopher~KI Williams, et~al.
\newblock \emph{Gaussian processes for machine learning}, volume~1.
\newblock Springer, 2006.

\bibitem[Tierney(1994)]{tierney1994markov}
Luke Tierney.
\newblock Markov chains for exploring posterior distributions.
\newblock \emph{the Annals of Statistics}, pages 1701--1728, 1994.

\bibitem[Carrera and Neuman(1986)]{Carrera1986estimation}
Jesus Carrera and Shlomo~P. Neuman.
\newblock Estimation of aquifer parameters under transient and steady state
  conditions: 2. uniqueness, stability, and solution algorithms.
\newblock \emph{Water Resources Research}, 22\penalty0 (2):\penalty0 211--227,
  1986.
\newblock \doi{https://doi.org/10.1029/WR022i002p00211}.
\newblock URL
  \url{https://agupubs.onlinelibrary.wiley.com/doi/abs/10.1029/WR022i002p00211}.

\bibitem[Cole et~al.(2001)Cole, Bergeron, Wurstner, Thorne, Orr, and
  Mckinley]{cole2001transient}
Charles~R Cole, Marcel~P Bergeron, Signe~K Wurstner, Paul~D Thorne, Samuel Orr,
  and Mathew~I Mckinley.
\newblock Transient inverse calibration of {Hanford} site-wide groundwater
  model to {Hanford} operational impacts-1943 to 1996.
\newblock Technical report, Pacific Northwest National Laboratory (PNNL),
  Richland, Washington, United States, 2001.

\bibitem[Tartakovsky et~al.(2017)Tartakovsky, Panzeri, Tartakovsky, and
  Guadagnini]{Tart2017WRR}
A.~M. Tartakovsky, M.~Panzeri, G.~D. Tartakovsky, and A.~Guadagnini.
\newblock Uncertainty quantification in scale-dependent models of flow in
  porous media.
\newblock \emph{Water Resources Research}, 53:\penalty0 9392--9401, 2017.

\bibitem[Hoffman et~al.(2014)Hoffman, Gelman, et~al.]{hoffman2014NUTS}
Matthew~D Hoffman, Andrew Gelman, et~al.
\newblock The no-u-turn sampler: adaptively setting path lengths in hamiltonian
  monte carlo.
\newblock \emph{J. Mach. Learn. Res.}, 15\penalty0 (1):\penalty0 1593--1623,
  2014.

\bibitem[Betancourt et~al.(2015)Betancourt, Byrne, and
  Girolami]{betancourt2015optimizing}
M.~J. Betancourt, Simon Byrne, and Mark Girolami.
\newblock Optimizing the integrator step size for hamiltonian monte carlo,
  2015.

\bibitem[Oliver et~al.(1996)Oliver, He, and Reynolds]{oliver1996conditioning}
Dean~S Oliver, Nanqun He, and Albert~C Reynolds.
\newblock Conditioning permeability fields to pressure data.
\newblock In \emph{ECMOR V-5th European conference on the mathematics of oil
  recovery}, pages cp--101. European Association of Geoscientists \& Engineers,
  1996.

\bibitem[Girolami and Calderhead(2011)]{Girolami2011RiemannML}
Mark~A. Girolami and Ben Calderhead.
\newblock Riemann manifold langevin and hamiltonian monte carlo methods.
\newblock \emph{Journal of the Royal Statistical Society: Series B (Statistical
  Methodology)}, 73, 2011.
\newblock URL \url{https://api.semanticscholar.org/CorpusID:6630595}.

\end{thebibliography}

\appendix
\section{Gaussian Process Regression and the Conditional Covariance}\label{sec:gpr}

Given the prior covariance kernel \eqref{eq:cov_kernel} for $y(\mathbf{x}; \tilde{\omega})$ and the set of observations $\{y_i^{\text{obs}} \}_{i = 1}^{N_y^{\text{obs}}}$, we obtain the optimal hyperparameters of the prior covariance kernel by minimizing the negative marginal log-likelihood function \cite{rasmussen2006gaussian}. Then, we assume the conditional field  $y^c(\mathbf{x}, \psi) \sim \mathcal{GP}(\overline{\mathbf{y}^c}(\mathbf{x}), \mathbf{C}_y^c(\mathbf{x}, \mathbf{x}^\prime))$ and compute the conditional mean and covariance of $y$ using GPR equations:
\begin{eqnarray}\label{eq:gpr_k}
        \overline{y^c}(\mathbf{x}_*) &=& \mathbf{C}^{x,*}_y(\mathbf{C}^{x,x}_y)^{-1}\mathbf{y}^{\text{obs}} \\ \label{eq:gpr_cov}
        \mathbf{C}^c_y(\mathbf{x}_*, \mathbf{x}_*^\prime) &=& \mathbf{C}_y(\mathbf{x}_*, \mathbf{x}_*^\prime) - \mathbf{C}^{x,*}_y(\mathbf{C}^{x,x}_y)^{-1}\mathbf{C}^{*^\prime,x}_y
    \end{eqnarray}
where $\overline{y^c}(\mathbf{x}_*)$ is the conditional mean function evaluated at the test point $\mathbf{x}_*$, $\mathbf{C}^c_y(\mathbf{x}_*, \mathbf{x}_*^\prime)$ is the conditional covariance evaluated at test points $\mathbf{x}_*, \mathbf{x}_*^\prime$. $\mathbf{C}^{x,x}_y \in \mathbb{R}^{N_y^{\text{obs}} \times N_y^{\text{obs}}}$ is the observation covariance matrix with its element $\mathbf{C}^{x,x}_{y,(ij)} = C_y(\mathbf{x}^{\text{obs}}_i, \mathbf{x}^{\text{obs}}_j)$.  $\mathbf{C}^{x,*}_y \in \mathbb{R}^{1 \times N_y^{\text{obs}}}$ is the vector of covariance between observations and the test point $\mathbf{x}_*$ with the element $\mathbf{C}^{x,*}_{y,i} = C_y(\mathbf{x}_*, \mathbf{x}^{\text{obs}}_i)$, and $\mathbf{C}^{*^\prime,x}_y \in \mathbb{R}^{N_y^s \times 1}$ is the vector of covariance between observations and the test point $\mathbf{x}_*^\prime$ with the element $\mathbf{C}^{*^\prime,x}_{y,i} = C_y(\mathbf{x}^\prime_*, \mathbf{x}^{\text{obs}}_i)$. 

The eigenfunctions and eigenvalues of Eq.~\eqref{eq:CKLE_k} are obtained by solving the following eigenvalue problem:
\begin{eqnarray}\label{eq:eigenvalue_problem_Y}
    \int C_y^c(\mathbf{x}, \mathbf{x^\prime})\psi_i^y(\mathbf{x}^\prime)d\mathbf{x}^\prime = \lambda^y_i\psi^y_i(\mathbf{x}^\prime) \quad i = 1, ..., N_\xi
\end{eqnarray}

To solve the eigenvalue problem, numerical approximations of these mathematical objects need to be introduced, which is usually consistent with the numerical model to solve the governing PDE. In this work, we discretize the domain with a finite volume mesh. The eigenvalue problem reduces to the corresponding eigendecomposition in the finite-dimensional vector space. We note that when $N_\xi$  equals the number of elements $N$ in the numerical model, the numerically approximated random field is exactly recovered. However, to reduce the dimensionality of the problem, the number of truncated terms $N_\xi$ is generally less than $N$.  The criteria for selecting $N_\xi$ is based on retaining enough energy in the expansion and is discussed in detail in \cite{Yeung2021PICKLE}. 

For representing the random field $\mathbf{u}(\mathbf{x}; \tilde{\omega})$ conditioned on $u$ measurements, we start by employing Monte Carlo (MC) simulations for computing the prior statistics for $u$. The reason for not directly using another parameterized covariance kernel is because (1) the random field for $u$ is generally not stationary, and (2) the parameterized prior does not enforce the physical constraint. We randomly draw an ensemble of $N_{mc}$ CKLE coefficients $\{ \xi_i \}_{i = 1}^{N_{mc}}$ from $\mathcal{N}(\mathbf{0}, \mathbf{I})$, and plug them into Eq. (\ref{eq:CKLE_k}) to get $N_{mc}$ realizations of parameters. Then, we run numerical simulations to get an ensemble of $N_{mc}$ state predictions. The prior statistics for $u$ can be computed based on the empirical mean and covariance of the ensemble:
\begin{eqnarray}\label{eq:mean_u}
    \overline{\mathbf{u}}^c &=& \frac{1}{N_{mc}} \sum_{i=1}^{N_{mc}} \mathbf{u}_i  , \\ \label{eq:cov_u}
    \mathbf{C}_u^c &=& \frac{1}{N_{mc} - 1} \sum_{i=1}^{N_{mc}} [\mathbf{u}_i - \overline{\mathbf{u}}^c][\mathbf{u}_i - \overline{\mathbf{u}}^c]^T.
\end{eqnarray}
We use GPR to obtain the conditional mean and covariance of $\mathbf{u}^c(\mathbf{x}, \tilde\eta^c) \sim \mathcal{GP}(\overline{\mathbf{u}^c}(\mathbf{x}), \mathbf{C}_u^c(\mathbf{x}, \mathbf{x}^\prime))$:
\begin{eqnarray}\label{eq:gpr_u}
    \overline{\mathbf{u}^c}(\mathbf{x}_*) &=& \mathbf{C}^{x,*}_u(\mathbf{C}^{x,x}_u)^{-1}\mathbf{u}^s \\
    \mathbf{C}^c_u(\mathbf{x}_*, \mathbf{x}_*^\prime) &=& \mathbf{C}_u(\mathbf{x}_*, \mathbf{x}_*^\prime) - \mathbf{C}^{x,*}_u(\mathbf{C}^{x,x}_u)^{-1}\mathbf{C}^{*^\prime,x}_u
\end{eqnarray}
where all matrices are counterparts in Eq.~\eqref{eq:gpr_k}. We solve the eigenvalue problem for the conditional covariance kernel $\mathbf{C}_u^c$ to obtain the CKLE representation of the random $u$ field:
\begin{eqnarray}\label{eq:eigenvalue_problem_u}
    \int C_u^c(\mathbf{x}, \mathbf{x^\prime})\psi_i^u(\mathbf{x}^\prime)d\mathbf{x}^\prime = \lambda^u_i\psi^u_i(\mathbf{x}^\prime) \quad i = 1, ..., N_\eta
\end{eqnarray}

\end{document}